%% file: iclr2027_conference.tex
\documentclass{article} % For LaTeX2e
\usepackage{iclr2027_conference,times}
\usepackage{graphicx}%
\usepackage{multirow}%
\usepackage{amsmath,amssymb,amsfonts}%
\usepackage{mathtools}      % math tools
\usepackage{amsthm}%
\usepackage{mathrsfs}%
\usepackage[title]{appendix}%
\usepackage{xcolor}%
\usepackage{textcomp}%
\usepackage{manyfoot}%
\usepackage{booktabs}%
\usepackage{algorithm}%
\usepackage{algorithmicx}%
\usepackage{afterpage}
\usepackage{algpseudocode}%
\usepackage{listings}%
\usepackage{tabularx}
\usepackage{bookmark}
\usepackage{longtable}
\usepackage{anyfontsize}
\usepackage{xcolor}
\usepackage{siunitx}
\usepackage{subcaption}
\usepackage{colortbl}
\usepackage{float}
\usepackage{wrapfig}
\usepackage{footnote}
\usepackage{caption} 
\usepackage{diagbox}
\makesavenoteenv{tabular}
\newcolumntype{K}[1]{>{\centering\arraybackslash}p{#1}}
\definecolor{mygray}{gray}{0.8}
\definecolor{mygray2}{gray}{0.9}
\definecolor{bestbg}{HTML}{D9EAD3}
\definecolor{worstbg}{HTML}{F4CCCC}
\newcommand{\best}[1]{\cellcolor{bestbg}#1}
\newcommand{\worst}[1]{\cellcolor{worstbg}#1}
\usepackage{array}   % for \newcolumntype
\newcolumntype{?}{@{\hspace{0.9em}}}   %

\input{math_commands.tex}

\usepackage{hyperref}
\usepackage{url}

\title{Cross-Domain Pretraining for Steady-State Neural CFD Surrogates}

\newcommand{\aff}[1]{\textsuperscript{#1}}

\author{Anthony Zhou$^{\,1,2,4}$,
Amir Barati Farimani$^{\,1}$,
Shirley Ho$^{\,2,3,4,5}$,
Rudy Morel\thanks{Correspondence: \texttt{rmorel@flatironinstitute.org}}$\;^{\,,4,6,7}$
\vspace{5pt}
   \\
  \textnormal{\aff{1}Carnegie Mellon University \quad
              \aff{2}New York University \quad
              \aff{3}Princeton University \quad
              \aff{4}Polymathic AI} \\
  \textnormal{\aff{5}Flatiron Institute, Center for Computational Astrophysics} \\
  \textnormal{\aff{6}Flatiron Institute, Center for Computational Mathematics} \\
  \textnormal{\aff{7}Flatiron Institute, Scientific Computing Core}}

\iclrfinalcopy % Uncomment for camera-ready version, but NOT for submission.
\begin{document}

\maketitle

\begin{abstract}

Neural surrogates for computational fluid dynamics (CFD) have the potential to greatly enhance engineering innovation through accelerating simulation. 
However, the primary limitation for neural surrogates is the lack of generalization to geometries and applications beyond the training set, which is significant given the diversity of engineering scenarios. 
Currently, this is addressed by generating a new dataset for a specific application; however, this requires running costly numerical solvers. In this work, we take a step toward addressing this by studying neural surrogates trained across different geometries, boundary conditions, and fidelities. We find that cross-domain pretraining improves zero- and few-shot performance on held-out datasets relative to both training from scratch and transferring from domain-specific experts. In particular, finetuning a pretrained, cross-domain model can achieve 2-3x lower errors at the same sample size and use 8x fewer samples to achieve the same error, compared to training from scratch. 
This benefit is architecture agnostic and improves with model size and pretraining dataset diversity. Furthermore, we study how and why cross-domain pretraining works in CFD surrogates, and find that simply pooling steady-state datasets is both sufficient and effective. Given the high cost of generating CFD data, leveraging existing datasets through cross-domain pretraining will likely be a valuable strategy as future surrogates expand to tackle new problems and use cases. 

%On held-out datasets, we find that finetuning a pretrained, cross-domain model is competitive with a domain-specific expert, or training a model from scratch, in predicting a steady-state solution.

%conventional pretraining insights from uniform grid, time-dependent PDEs do not usually transfer to the steady-state, mesh-based CFD setting. 
% mention zero shot and few shot to be more clear about what generalization is ... somehow be more precise about generalization

% mention why generalization is key; there is a ton of diversity in engineering, differnet industries, companies, softwares, etc. 

\end{abstract}

\section{Introduction}

Computational fluid dynamics (CFD) is an essential tool in modern engineering, producing simulations that inform design decisions across the automotive, aerospace, and consumer products industries. Although CFD now supports design and analysis throughout the engineering industry, the simulations themselves remain slow and expensive due to the difficulty of accurately resolving turbulent flows. Recent work in neural PDE surrogates has looked to address this through approximating solver outputs in a data-driven manner. These neural surrogates have been well-studied on uniform-grid, time-dependent PDEs \citep{ohana2025welllargescalecollectiondiverse, McGreivy_2024} and are now showing promise on 3D, mesh-based CFD problems. 

Given sufficient training data, advances in mesh-based transformer models now allow surrogates to accurately predict
steady-state surface and volume fields for a geometry and set of
simulation inputs \citep{luo2025transolveraccurateneuralsolver,
alkin2025abuptscalingneuralcfd, hagnberger2026smartscalablemeshfreeaerodynamic}. These models also reproduce aggregate quantities
such as \(C_d\) and \(C_l\) and support design optimization on practically
relevant datasets \citep{alkin2025abuptautomotiveaerospaceapplications,
paischer2026goingspeedsoundpushing, thumiger2026fasterdesigninteractiveaerodynamics}.
However, these capabilities are largely limited to the application 
% and operating envelope spanned 
covered
by the training set, whereas engineering practice spans a far wider space of geometries, boundary conditions, and solver fidelities.

Therefore, when deploying a surrogate model to a new application, we currently need to generate a new dataset. This can be very costly and requires a large investment in running the numerical solver the surrogate is meant to replace \citep{ashton2025fluidintelligenceforwardlook}. Furthermore, neural surrogates unlock the most benefit when approximating high-fidelity simulations, yet these are the most difficult to generate. A recent aerospace dataset from \citet{ashton2026hiliftaeromlhighfidelitycomputationalfluid} required 8 GB200 GPU nodes per sample, putting the estimated cost of its \(\sim\!10^3\) samples at millions of dollars of compute. Therefore, while we have methods to train CFD surrogates, the limiting factor is their lack of generalizability to new geometries and use cases and the high cost of generating data for a new application. 

In this work, we propose mitigating the need for application-specific data by leveraging existing CFD datasets through cross-domain pretraining. We aggregate a diverse set of steady-state, mesh-based CFD datasets, pretrain a set of models, and examine the downstream performance on held-out datasets. 
We present the first study of cross-domain pretraining for 3D, mesh-based CFD surrogates and make the following contributions:
\begin{enumerate}
    % \item We pretrain joint models across CFD domains and show that this improves few-shot and zero-shot errors to unseen datasets, compared to training an expert on any of the pretraining datasets alone. 
    \item We find that a joint model pretrained across CFD domains outperforms single-domain experts in zero- and few-shot transfer on a variety of held-out datasets.

    \item Finetuning the joint model reaches a target error on a held-out dataset with 8–16× fewer training samples compared to training from scratch.

    \item We show that finetuning performance on unseen datasets improves with both pretraining dataset diversity and model size.
    
    \item We release a unified collection of 8 publicly available CFD datasets, converted to a common format, along with dataloaders and training code for cross-domain experiments.

    % include this? 
    %\item The benefits of joint training are agnostic to the architecture. 
\end{enumerate}

Additionally, we release the code, datasets, and pretrained model checkpoints for this work, which can be found on \href{https://github.com/anthonyzhou-1/cross_domain_cfd}{Github} and \href{https://huggingface.co/collections/ayz2/cross-domain-cfd}{Huggingface}. 

\begin{figure}
    \centering
    \includegraphics[width=\linewidth]{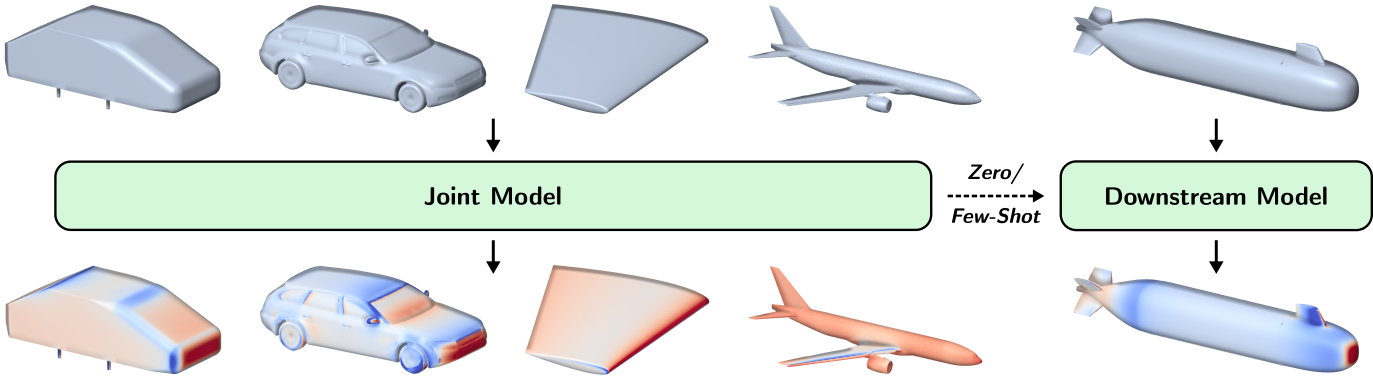}
    \caption{After pretraining a joint model across $6$ CFD datasets, we evaluate its zero/few-shot performance on $4$ held out datasets (see Table~\ref{tab:datasets}).}
    \label{fig:main}
\end{figure}

\section{Related Works}
\paragraph{Cross-Domain Training for PDEs.} 
Several works have considered training large-scale neural surrogates across uniform-grid, time-dependent PDEs. This is largely driven by robust data infrastructure and benchmarks \citep{ohana2025welllargescalecollectiondiverse,takamoto2024pdebenchextensivebenchmarkscientific,subramanian2023foundationmodelsscientificmachine}, as well as the speed of uniform-grid spectral methods used for data generation \citep{Koehler_2024, liu2026tadpoleautoencodersfoundationmodels}. These foundation models \citep{mccabe2024multiplephysicspretrainingphysical, hao2024dpotautoregressivedenoisingoperator, herde2024poseidonefficientfoundationmodels, sun2025foundationmodelpartialdifferential,morel2025discolearningdiscoverevolution, ye2026pdeformer2versatilefoundationmodel, rautela2026morphpdefoundationmodels,mccabe2026walruscrossdomainfoundationmodel} demonstrate good performance within their pretraining set, improved finetuning performance on held-out datasets, and can generalize to challenging, experimental applications \citep{mukhopadhyay2026emergenttransferphysicsfoundation}. 

In addition to leveraging scale, previous works have developed specific methods for cross-domain learning in PDEs, such as meta-learning \citep{kirchmeyer2022generalizingnewphysicalsystems, blanke2024interpretablemetalearningphysicalsystems, koupaï2024gepsboostinggeneralizationparametric}, in-context learning \citep{Yang_2023, serrano2025zebraincontextgenerativepretraining, lu_oneshot}, or scaling test-time compute \citep{serrano2026test, mansingh2026reasoningpdefoundationmodels}. Prior work has also explored self-supervised approaches, such as JEPA \citep{assran2023selfsupervisedlearningimagesjointembedding} and masked reconstruction \citep{he2021maskedautoencodersscalablevision}, for learning representations across PDE datasets \citep{zhou2024maskedautoencoderspdelearners, chen2025dataefficientoperatorlearningunsupervised, zhou2024strategiespretrainingneuraloperators, qu2026representationlearningspatiotemporalphysical}.

% New references

\paragraph{Mesh-based CFD Surrogates.} While uniform grids are a good test bench and are used in certain applications (such as weather \citep{rasp2024weatherbench2benchmarkgeneration}), the majority of engineering applications solve fluid dynamics on an unstructured mesh. Therefore, several prior works have developed surrogate models that can operate on meshes, such as using GNNs \citep{pfaff2021learningmeshbasedsimulationgraph, lino2025learningdistributionscomplexfluid, lam2023graphcastlearningskillfulmediumrange}, transformers \citep{li2023transformerpartialdifferentialequations, hao2023gnotgeneralneuraloperator, wu2024transolverfasttransformersolver}, continuous convolutions \citep{li2023geometryinformedneuraloperatorlargescale, hagnberger2025calmpdecontinuousadaptiveconvolutions, ranade2025dominodecomposablemultiscaleiterative, koupaï2025efficientgenerativetransformeroperators}, or latent spaces \citep{geo_fno, wang2024latentneuraloperatorsolving, zhou2025text2pdelatentdiffusionmodels,   li2025geometricoperatorlearningoptimal, li2025generativelatentneuralpde}. 

As these methods have matured, mesh-based surrogates are increasingly being applied to harder problems, where the main challenge is the sheer number of mesh points. Engineering CFD simulations are typically 3D and finely discretized, owing to the high Reynolds numbers ($10^5\!-\!10^7$) and complex geometries of industrial flows. As a result, these meshes can exceed 100 million cells, and only recently have methods emerged that can predict solution fields at this scale. Current state-of-the-art surrogates generally rely on some form of efficient attention, either through sub-sampling \citep{alkin2025abuptscalingneuralcfd, hagnberger2026smartscalablemeshfreeaerodynamic, giral2026aerojepalearningsemanticlatent}, local attention \citep{zhdanov2025erwintreebasedhierarchicaltransformer}, or sparse attention \citep{curvo2026msptefficientlargescalephysical, zhdanov2026sparseattentiondetailspreserving, zhou2026transolver3scalingtransformersolvers}. It is also important to be able to predict high-resolution fields sequentially, such that the memory can scale sub-linearly with the number of mesh points. Most related to our current work is a transfer learning study from \citet{keum2026adaptingautomotiveaerodynamicssurrogates}, which examines transferring CFD surrogates within a single domain and family of geometries (SUVs). In contrast, we pretrain across multiple domains (e.g., automobiles and aircraft), show that cross-domain pretraining outperforms single-domain experts, and study transfer and scaling behavior on domains unseen during pretraining.

\section{Methods}

\paragraph{Background.}
We consider the task of steady-state field prediction. Given a set of query points
$x = \{x_i\}_{i=1}^{N}$ with $x_i \in \mathbb{R}^3$, the goal is to predict the
corresponding field values $f(x) \in \mathbb{R}^{N\times d}$. These values depend on the geometry
$G = \{g_j\}_{j=1}^{M}$ of an object immersed in the flow, represented as a
point cloud with $g_j \in \mathbb{R}^3$, and on a vector of simulation parameters
$\theta \in \mathbb{R}^c$ (e.g., inlet velocity or angle of attack). We approximate this
mapping with a surrogate model $f_\phi$ with parameters $\phi$:
\begin{equation*}
    f_\phi : \mathbb{R}^{N\times 3} \times \mathbb{R}^{M \times 3} \times \mathbb{R}^c \to \mathbb{R}^{N \times d},
    \qquad
    f_\phi(x;\, G, \theta) \approx f(x;\, G, \theta), \quad i = 1, \ldots, N,
\end{equation*}
where each prediction depends globally on the geometry $G$, the simulation parameters
$\theta$, and potentially on the other queries in $x$. In general, \(N\) can be large (up to \(10^8\) or more), while \(M\) can be controlled by sampling the raw geometry. It is desirable that the surrogate model \(f_\phi(x; G, \theta)\) does not depend on the discretization of the geometry or the queries, and that the solution field can be predicted sequentially in batches of size \(n < N\) when the full resolution is needed. 

The field values are divided into surface and volume fields, depending on where the query lies. In particular, we consider $\mathcal{S} \subset \mathbb{R}^3$ as the surface of the geometry, from which the point cloud $G$ is sampled, and $\Omega \subset \mathbb{R}^3$ as the surrounding fluid volume. This results in the standard surface and volume fields from the Navier-Stokes equations: surface pressure $p$ and wall-shear stress $\vec{\tau}_w$ on $\mathcal{S}$, and pressure $p$ and velocity $\vec{u}$ in $\Omega$. The target field is then
\begin{equation}
    f(x;\, G, \theta) =
    \begin{cases}
        f_s(x;\, G, \theta), & x \in \mathcal{S}, \\
        f_v(x;\, G, \theta), & x \in \Omega. 
    \end{cases}, \quad     f_s = (p, \vec{\tau}_w) \in \mathbb{R}^4,
    \quad
    f_v = (p, \vec{u}) \in \mathbb{R}^4
\end{equation}
so that $d = 4$ for both fields. These fields are both important for CFD. Surface fields can be integrated to obtain the aerodynamic forces on the object and derive drag and lift, while volume fields can show how geometric features affect the surrounding flow and indicate regions for improvement.

\begin{table}[]
    \centering
    \small
    \addtolength{\tabcolsep}{-0.425em}
    \begin{tabular}{lllllllll}
    \toprule
        Name & Geometry & Closure & Solver & \# Samples & \# Cells & Re \# & Inlet (m/s) & AoA (\(^\circ\))\\
        \midrule
        DrivAerNet++ & DrivAer Car & RANS \(\kappa\text{-}\omega\) & OpenFOAM & 8000 & 24M & 8e6& 30 & - \\
        DrivAerML & DrivAer Car & SA-DDES & OpenFOAM & 500 & 160M & 7.2e6 & 38.9 & - \\
        WindsorML & Windsor Body & WMLES & Volcano & 355 & 280M & 2.9e6 & 40 & - \\
        AhmedML\(^*\) & Ahmed Body & SA-DDES & OpenFOAM & 500 & 20M & 7.7e5 & 1 & - \\
        SHIFT-Sub\(^*\) & Submarine & RANS & Luminary & 100 & 32M & 3.2e7 & 5 & - \\
        \midrule
        Emmi Wing & Tapered Wings & RANS SA & OpenFOAM & 30,000 & 3.3M & 5-20e6 & [150, 300] & [-10, 10]\\
        SuperWing & Kinked Wings & RANS SA & ADflow & 28,856 & 3.1M & 2e7 & [260, 313] & [2, 12]\\
        Double-Delta & Delta Wings & RANS SA & SU2 & 2448 & 7M & 8e7 & 101.5 & [11, 19]\\
        HiLiftAeroML\(^*\) & Passenger Plane & WMLES & charLES & 1800 & 500M & 1.6e6 & 68 & [4, 22]\\
        SHIFT-CCA\(^*\) & Combat Aircraft & RANS & Luminary & 100 & 32M & 3.1e7 & 212.5  & 2 \\
         \bottomrule
    \end{tabular}
    % \caption{An overview of the datasets considered in this work, roughly split into automotive and aerospace applications. The fidelity, physical regime, and use case of each dataset can vary widely.}
    \caption{Datasets considered in this work, roughly split into automotive (top) and aerospace (bottom) applications, spanning a variety of geometries, fidelities and physical regimes. Datasets with an asterisk\(^*\) are held-out for downstream evaluation. }
    \label{tab:datasets}
\end{table}

\paragraph{Datasets.}

To study cross-domain learning for CFD surrogates, we combine $K$ datasets
into a single training set $\mathcal{D} = \mathcal{D}_1 \cup \cdots \cup \mathcal{D}_K$
and sample from it, optionally weighting each dataset $\mathcal{D}_k$ by a sampling
probability $p_k$, with $\sum_{k=1}^{K} p_k = 1$. We consider 10 steady-state CFD datasets, split into 6 pretraining sets and 4 fine-tuning sets. 

The pretraining sets are: DrivAerNet++ \citep{elrefaie2025drivaernetlargescalemultimodalcar}, DrivAerML \citep{ashton2025drivaermlhighfidelitycomputationalfluid}, WindsorML \citep{ashton2025windsormlhighfidelitycomputationalfluid}, Emmi Wing \citep{paischer2026goingspeedsoundpushing}, SuperWing \citep{yang2026superwingcomprehensivetransonicwing}, and Double-Delta \citep{Shen_2026}. These cover common automotive and aerospace applications and span a diverse set of physical and numerical regimes. For example, the solutions are computed with both steady and unsteady solvers, using scale-resolving and time-averaged (RANS) approaches, and for incompressible and compressible flows. The main details of each dataset are shown in Table \ref{tab:datasets} to visualize the overall diversity. 

The finetuning sets are chosen to have meaningfully different geometries and simulation setups, and are AhmedML \citep{ashton2024ahmedmlhighfidelitycomputationalfluid}, HiLiftAeroML \citep{ashton2026hiliftaeromlhighfidelitycomputationalfluid}, SHIFT-Submarine \citep{shift-sub}, and SHIFT-CCA \citep{shift-cca}. We ensure that new, unseen physical regimes are present during finetuning, such as high-resolution aerodynamics (HiLiftAeroML) and hydrodynamics (SHIFT-Submarine). Lastly, the datasets differ in sign conventions, storage formats, and metadata, so we post-process each one into a common format optimized for fast dataloading. Additional details and visualizations for each dataset can be found in Appendix \ref{app:dataset}. 

Examining the datasets (Figures \ref{fig:all_surf} and \ref{fig:all_vol}) suggests why cross-domain training may be beneficial: many physical phenomena are shared across CFD simulations regardless of the specific geometry or flow configuration. For example, flow stagnates at the front of each geometry, producing high pressure and low wall-shear stress. Volumetric pressure and velocity follow a roughly inverse relationship outside of the boundary layer and wake, where regions of high pressure correspond to low velocity and vice versa. A low-velocity wake typically also forms downstream of the trailing edge.

%As the flow accelerates around the leading edge under a favorable pressure gradient, velocity and wall-shear stress increase, and pressure decreases. Sharp geometric features or high angles of attack (AoA) induce flow separation, which is accompanied by a drop in wall-shear stress. A low-velocity wake typically forms downstream of the trailing edge. Finally, in datasets where the AoA varies, the pressure distribution is consistent across geometries: positive angles produce a high-pressure lower surface and a low-pressure upper surface, and negative angles reverse this pattern.

%
\paragraph{Normalization.} Since datasets can span many physical regimes, the raw field values can span many orders of magnitude. We address this by predicting the non-dimensional quantities (coefficient of pressure \(C_p\), skin friction coefficient \(C_f\), and non-dimensional velocity \(u^*\)):
\begin{equation*}
    C_p = (p-p_\infty)/q_\infty, \quad C_f = \vec{\tau}_w / q_\infty, \quad u^* = \vec{u}/U_\infty, \quad q_\infty = \frac{1}{2}\rho_\infty U_\infty^2
\end{equation*}
where $\rho_\infty$, $p_\infty$, $U_\infty$, and $q_\infty$ are the freestream density, pressure, velocity magnitude, and dynamic pressure. After non-dimensionalization, the values are normalized by dataset-specific means and standard deviations. The points in the input geometry \(G\) are normalized by subtracting their centroid and scaling by a dataset-specific factor, computed such that the average geometry in a dataset is bounded by a $2\,$x$\,2\,$x$\,2$ box. The queries \(x\) are also oriented to the same frame by subtracting the centroid and scaling by the same factor. Geometries and output fields are aligned such that the inlet is at the $-x$ direction, and the $+z$ direction points upwards. Simulation parameters \(\theta\) are defined as the Mach number and angle of attack ([Ma, AoA]), and are normalized based on global min/max values across all datasets; in automotive datasets, these values are not used and are set to $0$. 
% \(\theta=[0\:\text{Ma}, 0^\circ]\). 

\paragraph{Models.}
The primary architecture we use is SMART \citep{hagnberger2026smartscalablemeshfreeaerodynamic}, due to its state-of-the-art performance on steady-state CFD tasks. However, we also evaluate architectures (AB-UPT \citep{alkin2025abuptscalingneuralcfd} and Transolver++ \citep{luo2025transolveraccurateneuralsolver}) to verify that the results are model-agnostic. Given the input geometry \(G\), simulation parameters \(\theta\) and queries \(x\), each model \(f_\phi\) is trained to predict volume and surface fields \(f(x)\) using a mean-squared error \(||f_\phi(x; G, \theta) - f(x; G,\theta) ||_2^2\). During training, we use a batch size of 1 and sample 32,768 surface and volume query points per batch. To ensure a fair comparison, all models are pretrained with identical settings unless otherwise noted: 32M parameters, 100k gradient steps, and the same learning rate and optimizer. For cross-domain pretraining, each batch is sampled from one of the six pretraining datasets without any extra information such as the dataset index. Note that the joint model therefore sees each dataset around 6 times less often than its corresponding experts. Further details on the architectures, training, and hyperparameters can be found in Appendix \ref{app:models}. 

\paragraph{Evaluation.}

We use the Normalized RMSE (NRMSE) to evaluate our models. Each model predicts four surface fields and four volume fields at a fixed set of 32,768 randomly sampled surface and volume points, and the error is computed separately for each quantity: \(C_p^{\text{surf}}, C_f, C_p^{\text{vol}}, u^*\). For brevity, these are shown in Appendix \ref{sec:var_specifics}, and in the main results the error is reported as the average over the 4 variable-level errors. Given a predicted field \(\hat{\mathbf{u}}\) with \(n\) points and \(c\) channels (\(\hat{\mathbf{u}} \in \mathbb{R}^{n \times c}\)): 
\begin{align*}
    \mathcal{L}_\text{NRMSE}(\hat{\mathbf{u}}, \mathbf{u})
    &= \frac{\lVert \hat{\mathbf{u}} - \mathbf{u} \rVert_2}{\lVert \mathbf{u} \rVert_2}, \quad 
    \mathcal{L}_\text{avg}
    = \frac{1}{\lvert \mathcal{U} \rvert} \sum_{\mathbf{u} \in \mathcal{U}}
       \mathcal{L}_\text{NRMSE}(\hat{\mathbf{u}}, \mathbf{u}),
    \quad
    \mathcal{U} = \{ C_p^\text{surf},\, C_f,\, C_p^\text{vol},\, u^* \}.
\end{align*}
All models are evaluated on validation sets that contain geometries unseen during training. Lastly, all reported errors are computed in the denormalized, physical space (raw \(C_p, C_f, u^*\)). 

\section{Results}

\subsection{Generalization to Unseen Datasets}
\label{sec:generalization}

\paragraph{Zero/Few-Shot Performance.}
The primary goal of this work is to leverage prior knowledge from existing CFD datasets to reduce the need for new data when deploying a surrogate to a new engineering task. Therefore, we consider pretraining a joint model across CFD domains, as well as domain experts on each individual dataset. After pretraining, we finetune each model to a restricted number of samples \(S\in\{2,4,8\}\) with a budget of 400 gradient steps, as well as evaluate zero-shot performance \((S=0\)). We consider an additional baseline of a randomly initialized model, which is trained from scratch on the finetuning samples.

\begin{table}[t]
    \centering
    \small
    \addtolength{\tabcolsep}{-0.375em}
    \begin{tabular}{@{}l?cccc?cccc?cccc?cccc@{}}
        \toprule
        %& \multicolumn{16}{c}{Held-out target dataset} \\
        %\cmidrule(lr){2-17}
        \rlap{Held-out Datasets:} & \multicolumn{4}{c}{AhmedML}
        & \multicolumn{4}{c}{SHIFT-Sub}
        & \multicolumn{4}{c}{SHIFT-CCA}
        & \multicolumn{4}{c}{HiLiftAeroML} \\
        \cmidrule(lr){2-5}
        \cmidrule(lr){6-9}
        \cmidrule(lr){10-13}
        \cmidrule(lr){14-17}
        \rlap{\(\#\) FT Samples:} 
        & 0 & 2 & 4 & 8
        & 0 & 2 & 4 & 8
        & 0 & 2 & 4 & 8
        & 0 & 2 & 4 & 8 \\
        \midrule

        \multicolumn{17}{@{}l}{\textit{No Pretraining}} \\[-0.15em]
        Random init.
        & 0.68 & 0.54 & 0.49 & 0.34
        & 0.64 & 0.39 & 0.33 & 0.33
        & 0.68 & 0.23 & 0.23 & 0.23
        & \textbf{0.81} & 0.63 & 0.64 & 0.63 \\

        \addlinespace[0.25em]
        \multicolumn{17}{@{}l}{
            \textit{Domain Experts}
        } \\[-0.15em]

        DrivAerNet++
        & 0.53 & 0.18 & 0.15 & 0.12
        & 0.61 & 0.22 & 0.13 & 0.10
        & 0.73 & 0.15 & 0.11 & 0.11
        & 1.10 & 0.53 & 0.47 & 0.46 \\

        Emmi-Wing
        & 0.73 & 0.21 & 0.21 & \worst{0.19}
        & 0.66 & \worst{0.24} & 0.19 & \worst{0.16}
        & \best{0.61} & 0.14 & 0.12 & 0.12
        & 1.10 & 0.52 & 0.48 & 0.49 \\

        SuperWing
        & \worst{0.82} & \worst{0.24} & \worst{0.22} & 0.19
        & \worst{0.78} & 0.23 & 0.16 & 0.14
        & 0.74 & 0.13 & 0.12 & 0.12
        & \worst{1.26} & 0.50 & 0.46 & 0.45 \\

        DrivAerML
        & 0.51 & 0.18 & 0.18 & 0.16
        & 0.61 & \best{0.20} & \best{0.12} & \best{\textbf{0.09}}
        & 0.71 & 0.14 & 0.11 & \best{\textbf{0.10}}
        & 0.99 & 0.55 & 0.46 & 0.46 \\

        WindsorML
        & \best{0.38} & \best{0.17} & \best{0.13} & \best{0.11}
        & \best{0.59} & 0.21 & 0.15 & 0.11
        & \worst{0.75} & \worst{0.17} & \worst{0.14} & \worst{0.14}
        & 1.04 & \worst{0.57} & \worst{0.51} & \worst{0.50} \\

        Double-Delta
        & 0.71 & 0.19 & 0.19 & 0.17
        & 0.65 & 0.24 & \worst{0.20} & 0.14
        & 0.69 & \best{\textbf{0.11}} & \best{0.11} & 0.11
        & \best{0.86} & \best{0.48} & \best{0.37} & \best{0.36} \\

        \addlinespace[0.25em]
        \multicolumn{17}{@{}l}{\textit{Cross-Domain}} \\[-0.15em]

        Joint Model
        & \textbf{0.31} & \textbf{0.12} & \textbf{0.10} & \textbf{0.09}
        & \textbf{0.52} & \textbf{0.14} & \textbf{0.10} & \textbf{0.09}
        & \textbf{0.52} & \textbf{0.11} & \textbf{0.10} & \textbf{0.10}
        & 0.90 & \textbf{0.46} & \textbf{0.34} & \textbf{0.33} \\
        \bottomrule
    \end{tabular}

    \caption{Validation error on four held-out target datasets.
    Rows indicate the model initialization; single-domain experts are
    identified by their pretraining dataset. Columns give the number of
    fine-tuning samples, with 0 denoting zero-shot evaluation. Green and
    red indicate the best and worst single-domain experts, respectively,
    and bold indicates the lowest overall error.}
    \label{tab:generalization}
\end{table}

\begin{figure}[t]
    \centering
    \includegraphics[width=\linewidth]{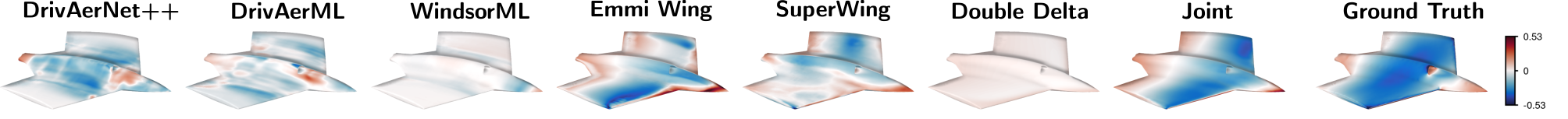}
    \caption{
    Zero-shot predictions (\(C_p\)) of pretrained models on a SHIFT-CCA geometry. Expert models can overfit to their pretraining sets, while the Joint model better predicts new geometries.
    }
    \label{fig:aml_zeroshot}
\end{figure}

Zero-shot and few-shot errors are given in Table \ref{tab:generalization}. We find that cross-domain training achieves lower errors than other models during zero-shot and few-shot evaluation, even under an equal pretraining budget. A potential explanation for this is that training across domains prevents overfitting to a specific geometry/simulation setup and learns more general representations that can be used when finetuning to new data. We observe this in Figure \ref{fig:aml_zeroshot}, where zero-shot predictions a SHIFT-CCA geometry are plotted. When queried on unseen geometries, expert models tend to inherit biases from their pretraining data, for example predicting car-like features from DrivAerNet/ML or flow separation from Emmi Wing/Superwing that are not present in the target geometry. Appendix \ref{app:viz} provides additional visualizations, which further show the advantage of the joint model. Beyond improving generalization, cross-domain pretraining offers a practical benefit: existing datasets can be pooled to train a single model that serves all downstream tasks, eliminating the need to maintain a library of experts and select among them at deployment.

Another observation is that experts close to the finetuning set tend to perform better and farther ones tend to perform worse. For example, adapting WindsorML to AhmedML performs well as both are simple automotive bodies, but adapting WindsorML to HiLiftAeroML performs poorly due to transferring from automotive to aerospace datasets (Table \ref{tab:generalization}). In addition, the worst expert is always better than training from scratch in the few-shot setting, suggesting that any form of pretraining is beneficial in limited-data scenarios. However, in the zero-shot setting, the worst expert can be worse than a random initialization, so expert models can be unreliable when no finetuning data is available.

\begin{figure}
    \centering
    \includegraphics[width=\linewidth]{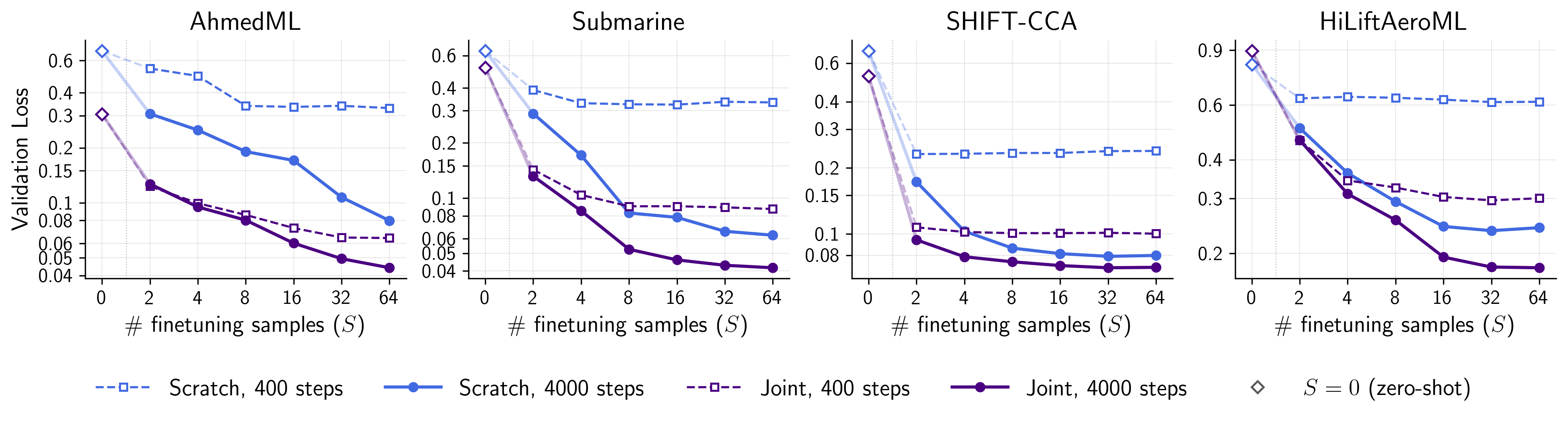}
    \caption{Validation loss vs. number of finetuning samples, plotted for the pretrained joint model and a model initialized from scratch. Two finetuning budgets are used (400/4000 steps). Note both axes are log scale.}
    \label{fig:efficiency}
\end{figure}

\paragraph{Sample Efficiency.} To quantify sample efficiency, we compare the joint model to a fixed reference, chosen as the randomly initialized model. On the four held-out sets, we plot the validation loss as a function of the finetuning sample size \(S\) in Figure \ref{fig:efficiency}. At the original finetuning budget (400 steps), no randomly initialized model is able to outperform a joint model with \(S=2\), even as we increase \(S\) to the full held-out dataset size (\(S>64\)). In this compute-limited setting, a joint model with only two samples outperforms a randomly initialized model that is provided the entire dataset. 

This is important for fast adaptation of a pretrained model, however, we can often afford a larger finetuning budget, which we increase by 10 times (4000 steps). Given the error of the scratch model at a fixed sample size (\(S=64\)) we compute what sample size the joint model needs to achieve the same error. Across the four finetuning sets, the joint model achieves the same error at \(S\approx4\text{-}8\) samples, which is 8-16 times fewer samples. Furthermore, a pretrained model needs very little data to adapt: at any finetuning budget, just two training samples can cut the zero-shot error in half or more. With two finetuning samples, the NRMSE can also be competitive (\(\sim\!0.1\)), even though validation is performed on more than ten times as many samples (\(20\text{-}100\)), each with a new geometry. Given these performance gains, future CFD surrogates are likely to benefit substantially from a pretrained initialization.  This benefit comes from using a pretrained model to accelerate convergence and achieve lower error on a new dataset, but also from reducing the number of samples that must be generated for a new application.

\paragraph{Different Architectures.}
To validate that the findings are agnostic to the underlying architecture, we replicate the primary pretraining and finetuning results for AB-UPT \citep{alkin2025abuptscalingneuralcfd} and Transolver++ \citep{luo2025transolveraccurateneuralsolver}. Expert and joint models are pretrained and evaluated on the held-out datasets, and the results are shown in Appendix \ref{app:gen_results}. We find that the same trends hold for other architectures, where the joint model consistently outperforms any expert model during zero-shot and few-shot evaluation. This suggests that cross-domain pretraining is likely a good strategy for future architectures as well and complements ongoing work in CFD surrogates. We also verify that each architecture (SMART/AB-UPT/Transolver++) has similar pretraining errors (Appendix \ref{app:gen_results}). 

\begin{figure}[t]
    \centering
    \includegraphics[width=\linewidth]{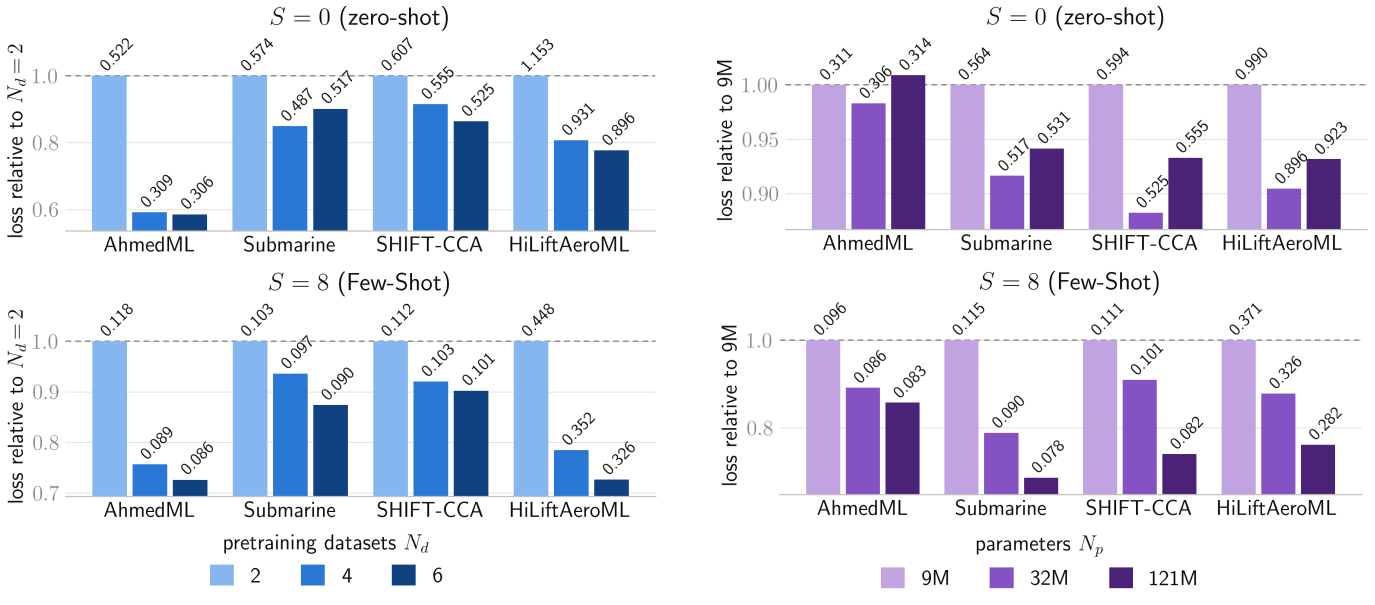}
    \caption{Error of joint models on held-out datasets, after using \(S=0\) or \(S=8\) finetuning samples. \textit{Left:} Joint models are pretrained with different numbers of pretraining datasets (2/4/6), with the model size held constant (\(N_p = \text{32M}\)). \textit{Right: } Joint models are pretrained at varying model sizes (9/32/121M), with the size of the pretraining dataset held constant (\(N_d = \text{6}\)).}
    \label{fig:ds_scaling}
\end{figure}

\subsection{Scaling Model and Dataset Size}

If cross-dataset pretraining improves downstream performance, a natural question is how that performance scales with model size and dataset diversity. To test this, we train variants of the joint model on different numbers of pretraining datasets \(N_d\) as well as different model sizes \(N_p\). We consider training on 2 datasets (DrivAerNet/Emmi Wing), 4 datasets (DrivAerNet/Emmi Wing/WindsorML/Double Delta), as well as all 6 pretraining datasets. Furthermore, we consider 3 model sizes, with parameter counts of \(N_p = 9/32/121\text{M}\). The zero-shot and few-shot performance of joint models on held-out datasets are shown in Figure \ref{fig:ds_scaling}. In general, we find that increasing the number of datasets seen during pretraining or the model size improves downstream performance. In the zero-shot setting there are some exceptions, where it seems that the largest model may overfit to the pretraining dataset. However, both large models and models trained on a diverse pretraining set are more sample efficient during finetuning, reaching lower errors at a fixed gradient and data budget (400 steps, \(S=8\)). 

The results suggest that the best way to achieve zero-shot generalization is by pretraining on a more diverse collection of data, rather than just increasing the model size. If data samples are available for the downstream application, then a larger model size can use these samples more efficiently. Therefore, generating a large and diverse pretraining dataset is the most effective lever for improving a cross-domain CFD surrogate, but it is also the main obstacle since simulation is likely more expensive than the model training itself. If this isn't feasible, a large model combined with a handful of samples (\(S<10\)) per downstream application is also a viable strategy.

%All joint models are pretrained to the same number of gradient steps (\(100k\)) and finetuned for \(400\) steps. 

\subsection{Pretraining Performance}

The primary focus of this work examines model performance on new, unseen datasets. This has led to useful results and scaling trends; however, there are also some interesting findings from examining the pretraining performance. These may be relevant when models are deployed to standard, preexisting applications or just to understand what the tradeoffs are for cross-domain pretraining. We examine the performance of expert and joint models on the pretraining datasets in Table \ref{tab:full}. For joint models, we also consider varying the number of pretraining datasets (\(N_d = 2, 4, 6\)) to include progressively more datasets. 
%All models are the same size (\(32\text{M}\) parameters) and are trained for the same number of gradient steps (\(100k\)).

For a fixed model size and compute budget, the best-performing model on each pretraining dataset is its corresponding expert, since that model sees only one dataset during training and can devote all of its parameters and gradient steps to fitting it. Training an expert is therefore the preferred strategy when a task has abundant data, though it sacrifices performance on the remaining datasets. We also find that experts generalize better to datasets similar to their own: the DrivAerNet model achieves lower errors on WindsorML and DrivAerML, whereas Double-Delta is the hardest target to generalize to, being the only full-body aircraft dataset.

\begin{table}[t]
    \centering
    \small 
    \addtolength{\tabcolsep}{-0.15em}
        \begin{tabular}{p{2.25cm}cccccc}
        \toprule
        \textit{Model} & DrivAerNet & Emmi-Wing & WindsorML & Double-Delta & DrivAerML & SuperWing \\
        \midrule
        DrivAerNet & \cellcolor[HTML]{BBE2B9} 0.143 & \cellcolor[HTML]{FBC5AE} 0.855 & \cellcolor[HTML]{F9FCD9} 0.470 & \cellcolor[HTML]{F2ACA2} 0.973 & \cellcolor[HTML]{EFF8CD} 0.401 & \cellcolor[HTML]{FCCCB1} 0.830 \\
        Emmi-Wing & \cellcolor[HTML]{FDD1B3} 0.809 & \cellcolor[HTML]{9DD3B1} 0.030 & \cellcolor[HTML]{FFF4CF} 0.617 & \cellcolor[HTML]{D78C9D} 1.239 & \cellcolor[HTML]{FCCCB1} 0.830 & \cellcolor[HTML]{FEDFBB} 0.742 \\
        WindsorML & \cellcolor[HTML]{FEEAC5} 0.678 & \cellcolor[HTML]{FEDEBA} 0.747 & \cellcolor[HTML]{A7D8B4} 0.067 & \cellcolor[HTML]{F3AFA4} 0.956 & \cellcolor[HTML]{FEE6C2} 0.701 & \cellcolor[HTML]{FEE0BD} 0.736 \\
        Double-Delta & \cellcolor[HTML]{FEDAB8} 0.766 & \cellcolor[HTML]{FEEEC9} 0.657 & \cellcolor[HTML]{FFF7D5} 0.587 & \cellcolor[HTML]{A5D7B3} 0.057 & \cellcolor[HTML]{FED7B6} 0.781 & \cellcolor[HTML]{FEECC6} 0.668 \\
        DrivAerML & \cellcolor[HTML]{F3FAD1} 0.421 & \cellcolor[HTML]{FDD5B6} 0.786 & \cellcolor[HTML]{F7FCD8} 0.458 & \cellcolor[HTML]{F2ADA3} 0.966 & \cellcolor[HTML]{A2D6B3} 0.048 & \cellcolor[HTML]{FDD2B4} 0.803 \\
        SuperWing & \cellcolor[HTML]{F2ACA2} 0.973 & \cellcolor[HTML]{FFEFC9} 0.648 & \cellcolor[HTML]{FEE6C1} 0.705 & \cellcolor[HTML]{D78C9D} 1.255 & \cellcolor[HTML]{EEA59F} 1.006 & \cellcolor[HTML]{A2D6B3} 0.047 \\
        \midrule
        Joint (\(N_d=2\)) & \cellcolor[HTML]{BEE3B9} 0.151 & \cellcolor[HTML]{9ED4B2} 0.035 & \cellcolor[HTML]{F7FCD8} 0.458 & \cellcolor[HTML]{E3989E} 1.075 & \cellcolor[HTML]{F1F9CE} 0.408 & \cellcolor[HTML]{FEE5C0} 0.708 \\
        Joint (\(N_d=4\)) & \cellcolor[HTML]{C0E4B9} 0.164 & \cellcolor[HTML]{9FD4B2} 0.041 & \cellcolor[HTML]{A9D9B4} 0.072 & \cellcolor[HTML]{ABDAB5} 0.083 & \cellcolor[HTML]{EFF8CC} 0.394 & \cellcolor[HTML]{FFEFC9} 0.652 \\
        Joint (\(N_d=6\)) & \cellcolor[HTML]{C2E5BA} 0.174 & \cellcolor[HTML]{A1D5B2} 0.043 & \cellcolor[HTML]{A9D9B4} 0.075 & \cellcolor[HTML]{AEDBB6} 0.091 & \cellcolor[HTML]{B9E1B8} 0.134 & \cellcolor[HTML]{ABDAB5} 0.082 \\
        \bottomrule
        \end{tabular}
    \caption{Validation error of expert and joint models across the pretraining datasets. Joint models are trained at different pretraining dataset sizes \(N_d=2,4,6\).
    Cells are colored by error magnitude.
    }
    \label{tab:full}
\end{table}

Joint training trades low error on any single dataset for consistent performance across all of them. The trade-off is gradual: as the pretraining pool grows, per-dataset error rises slightly but average error improves, with no sign of catastrophic forgetting. Under a fixed parameter and gradient budget this is unavoidable since each added dataset reduces the steps and capacity available per task. Joint training is also a harder learning problem, requiring the model to accommodate multiple geometries and output fields, but we hypothesize that this difficulty is what yields better generalization. Scaling model size or pretraining compute largely recovers pretraining performance (Appendix \ref{app:scaling}), but this can be misleading. Lower pretraining loss does not imply better transfer, as these scaled models can have higher zero-shot error on held-out domains, suggesting that overfitting can occur.

\iffalse 
If the joint task is more difficult, then an increased model capacity may be able to address this. We consider training a larger joint model for the same gradient budget, as well as increasing the gradient steps at the same model size. At six times the number of gradient steps, the model has trained on each dataset an equal amount compared to its associated expert model. In general, we find that increasing the model size or compute budget recovers most of the performance compared to an expert model. However, this can be misleading as a lower pretraining loss does not guarantee lower zero-/few-shot errors on unseen datasets. Indeed, we find that both the 121M and 6xSteps models have higher zero-shot errors compared to the base joint model, which suggests that overfitting occurs. We further discuss this in Appendix \ref{}. 
\fi 

\subsection{Examining Ambiguity in Cross-Domain Pretraining}

\paragraph{Overview.}

Predicting a steady-state solution from geometry alone can be ill-defined, as the same geometry can map to different solutions depending on variables we cannot always observe. For example, different closure models, mesh resolutions, or modeling choices such as whether to use rotating wheels, can result in different solutions for the same geometry \citep{ashton2025fluidintelligenceforwardlook}. Fully enumerating these in the simulation parameters \(\theta\) is usually not feasible, however a domain-specific model can resolve this ambiguity by training on data with fixed simulation choices. However, in the cross-domain setting these hidden variables can vary across datasets, which we study in this section. 

In the conventional, time-dependent PDE setting, this ambiguity is usually addressed by using context frames from the same rollout as a model input \citep{mccabe2024multiplephysicspretrainingphysical}. In this case, the context frames are guaranteed to have the same physics as the current state, which allows the model to infer the dynamics and uniquely forward the current state in time. In steady-state problems we do not have access to context frames; nevertheless, joint models are still able to resolve this ambiguity and distinguish different dynamics from the geometry \(G\) and simulation parameters \(\theta\) alone. This is true even when the geometries originate from the same base geometry, such as in DrivAerNet/DrivAerML. Therefore, to study this behavior, we create toy problems where the geometry alone cannot determine the dynamics, as well as train joint models on the combined DrivAerNet/ML datasets.

\begin{figure}[t]
    \centering
    \begin{minipage}{0.45\textwidth}
        \centering
        \includegraphics[width=\linewidth]{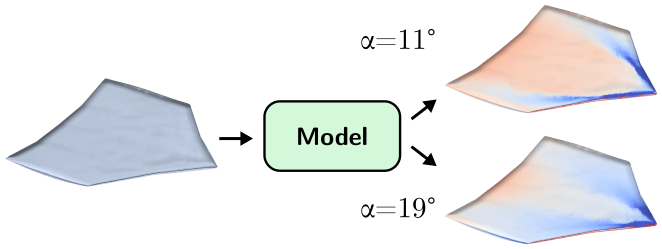}
        \captionof{figure}{A case where the same geometry can have different physics.}
        \label{fig:alpha}
    \end{minipage}\hfill
    \begin{minipage}{0.52\textwidth}
        \centering
        \addtolength{\tabcolsep}{-0.3 em}
        \begin{tabular}{lccc}
    \toprule
        Model Input: & \(G\) & \(G\) + context & \(G\) + \(\theta\) \\
        \midrule
        Double-Delta-\(\alpha\) & 0.093 & 0.076 & 0.068 \\
        DrivAerNet-\(\nu\) & 0.450 & 0.164 & 0.150 \\
        DrivAerNet/ML & 0.109 & 0.111 & 0.109 \\
     \bottomrule
    \end{tabular}
    \captionof{table}{Validation error on the toy problems, for models receiving the geometry \(G\) only (Vanilla), \(G\) plus a paired simulation
    sharing the same \(\theta\) (In-Context), or \(G\) plus \(\theta\) directly (Oracle).}
    \label{tab:toyprobs}
    \end{minipage}
\end{figure}

\paragraph{Toy Problems.} To study the ill-defined setting, we construct three toy problems in which
the simulation parameters $\theta$ are withheld from the model. In the first
(Double-Delta-$\alpha$), a fixed set of geometries is solved across a range of angles
of attack ($\alpha \in [11, 19]^\circ$), so that a single geometry maps to several
distinct solutions (Figure~\ref{fig:alpha}). In the second (DrivAerNet-$\nu$),
DrivAerNet geometries are solved at two combinations of viscosity and inlet velocity,
$(\nu, U) \in \{(1.0 \times 10^{-5},\, 20\,\mathrm{m/s}),\,
(2.3 \times 10^{-5},\, 45\,\mathrm{m/s})\}$. The third mixes samples from DrivAerNet
and DrivAerML that were selected for geometric similarity, with detailed underbodies
and open wheels in both. For each problem we compare three variants that differ in what the model receives as input alongside the geometry: nothing (Vanilla), a paired simulation sharing the same
$\theta$ (In-context), or $\theta$ itself (Oracle). In the DrivAerNet/ML case, the
in-context sample is drawn from the matching dataset and the oracle signal is the dataset label. All models are the same size (32M) and are trained to the
same budget (50k steps).

Results are reported in Table~\ref{tab:toyprobs}. For Double-Delta-\(\alpha\), we see that the vanilla model struggles without the parameter information, while the in-context model has a small penalty for needing to infer \(\theta\). This trend is also seen in DrivAerNet-\(\nu\), although it is a more difficult setting. Surprisingly, in the DrivAerNet/ML case, the vanilla model recovers performance identical to that of the oracle. While the DrivAerNet/ML geometries are similar, it seems that differences in discretization or parametric morphing are enough for neural surrogates to distinguish the two. In this case, there is no benefit to providing context to the model during training and it can even be detrimental.

\paragraph{Takeaways.}

Our experiments show that cross-domain learning for CFD surrogates can succeed in practice by simply pooling datasets, without any context about the source domain or knowledge of hidden variables. In the DrivAerNet/ML case and when pretraining across six datasets, supplying an extra dataset label or context simulation does not help, and can even reduce performance slightly. We further discuss this in Appendix \ref{app:cross}. 

This departure from context-based PDE foundation models reflects several differences in the steady-state CFD setting. First, CFD surrogates take a dense, unstructured point cloud as input, and unlike a fixed uniform grid, the discretization itself carries variation that the model can use to distinguish domains. Second, PDE benchmarks are typically built around well-defined splits over PDE terms or coefficients that can be inferred from context or supplied as conditioning \citep{pmlr-v202-takamoto23a, gupta2022multispatiotemporalscalegeneralizedpdemodeling}. We construct two such splits as toy problems, but practical CFD settings are rarely so well-defined. For example, when training the oracle in DrivAerNet/ML, using the dataset index as conditioning is no longer physically meaningful, unlike for \(\alpha, \nu, U\). Third, steady-state CFD datasets are constructed so that the output is determined by the geometry and simulation parameters alone, and therefore context or additional information is likely redundant. Additional information is needed for parameters that vary across domains (closure, solver, discretization), however these rarely vary within a geometry, since there is little incentive to solve the same geometry twice across separate datasets. Lastly, during finetuning, we assume a fixed simulation setup, so any remaining hidden variables are constant and can be absorbed few-shot. Therefore, in the current setting, cross-domain methods designed around PDE terms and coefficients appear unnecessary for CFD surrogates, unless a split has been deliberately constructed.

\section{Conclusion}
Overall, this work shows that cross-domain pretraining improves the zero- and few-shot performance of CFD surrogates on unseen datasets. A pretrained model yields lower error and reduces the number of samples that must be generated to deploy a surrogate in a new application. These gains also grow with both the diversity of the pretraining data and the size of the model. 

While these results are encouraging, there a few limitations of this work. Firstly, the number of datasets considered is still small compared to cross-domain works in PDEs, and current CFD datasets likely have less diversity than time-dependent PDE benchmarks. Second, the main results assume ground truth normalization statistics for the finetuning sets, which may not always be available. However, this can be mostly addressed through non-dimensionalization, which we discuss in Appendix \ref{app:no_norm}. Lastly, we do not test the discretization invariance of the models, nor do we compute any errors on the full-resolution field (millions of cells). We rely on previous work that does this, where the full resolution generally has a slight increase in the error (for AB-UPT/SMART). 

Despite these limitations, cross-domain pretraining shows clear promise as neural CFD surrogates expand into a wider range of engineering applications. We therefore advocate for continued work to generate steady-state CFD datasets spanning as many geometries and applications as possible. Pooled together, these datasets could support increasingly large, cross-domain models that serve as general-purpose CFD surrogates.

\subsection*{AI use statement}

In this work, we used generative AI tools for implementing methods, cleaning and reformatting datasets, and for supporting qualitative and thematic data analysis. We have not used generative AI tools to propose or refine hypotheses, design or provide feedback on research methodology or experiments, or interpret results. The rest of the tasks (generating synthetic data sets, developing theoretical models or conceptual frameworks, formulating mathematical claims, providing critical ingredients for proving mathematical claims, or assisting in the writing of proofs) are not applicable to this work. We have reviewed all AI-assisted work. For coding agents, we have manually reviewed all implementations and verified that they work as intended. For data visualization and writing, we have taken raw outputs and modified them to our specific needs. We take responsibility for the final content of this work, including text, claims or artifacts produced with the aid of generative AI.

\subsection*{Ethics statement}

There are no ethics considerations to disclose. 

\subsection*{Reproducibility statement}
All source code and pretrained model checkpoints will be released to replicate results. Furthermore, all datasets will also be released in the processed, formatted versions, with the exception of datasets generated by private companies. 

\iffalse 
\subsubsection*{Author Contributions}
If you'd like to, you may include  a section for author contributions as is done
in many journals. This is optional and at the discretion of the authors.
\fi 

\subsubsection*{Acknowledgments}
We would like to thank the Scientific Computing Core at the Flatiron Institute, a division of the Simons Foundation, for providing computational resources and support. We also thank Jan Hagnberger, Fabian Paischer, and Sacha Lewin for insightful discussions and support. We also thank Michael Emory and Luminary Cloud for providing sample datasets for public use. We would like to acknowledge the support of the Simons Foundation and Schmidt Sciences. This work was supported in part by the AI2050 program at Schmidt Sciences (Grant G-25-70028). 

\bibliography{bob}
\bibliographystyle{iclr2027_conference}

\newpage 
\appendix
\section{Dataset Details}
\label{app:dataset}

\begin{figure}
    \centering
    \includegraphics[width=\linewidth]{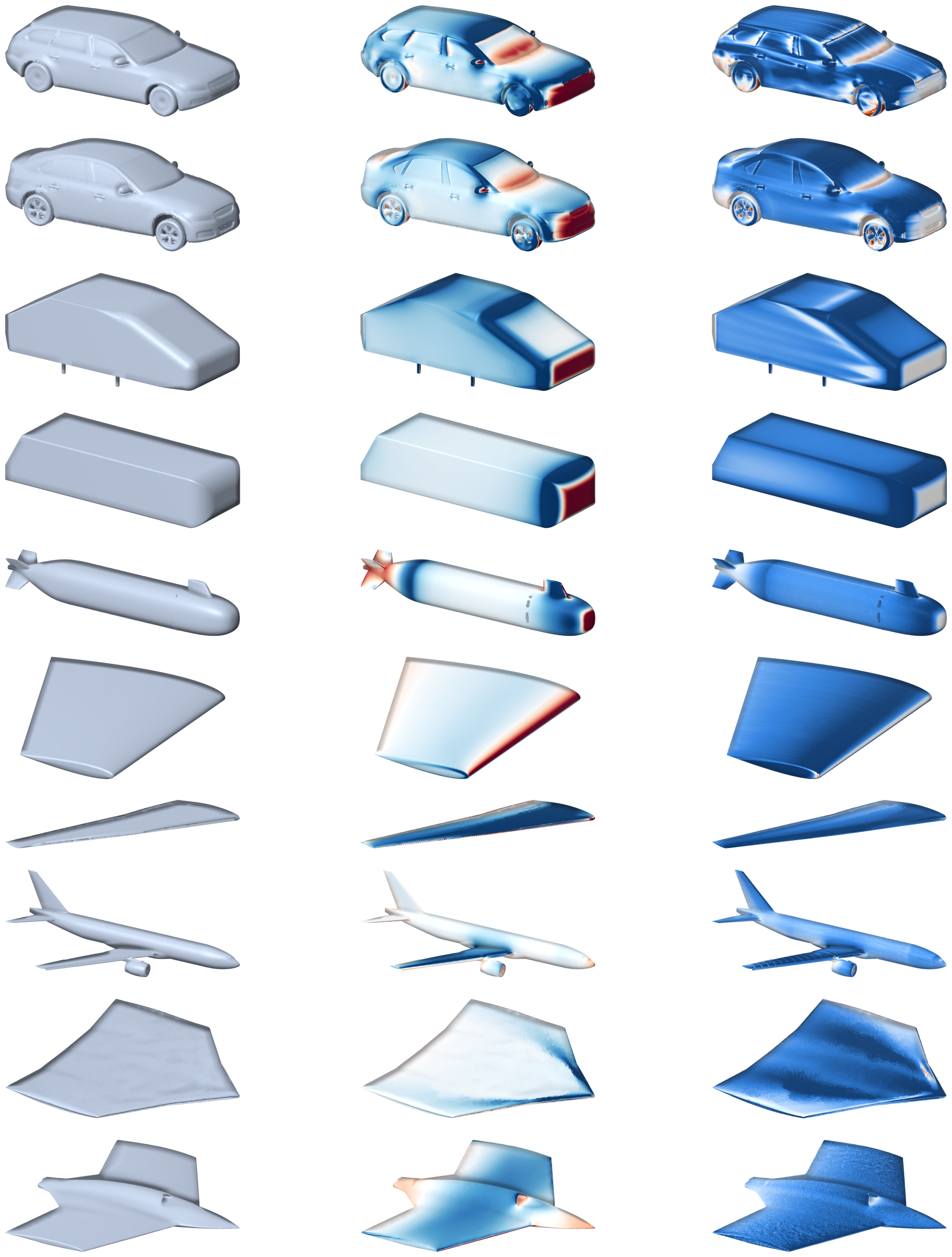}
    \caption{The geometry (left), coefficient of pressure \(C_p\) (center), and x-component of the skin friction coefficient \(C_{f_x}\) (right) visualized for a single sample in all datasets. From top to bottom: DrivAerNet, DrivAerML, WindsorML, AhmedML, SHIFT-Submarine, Emmi Wing, SuperWing, HiLiftAeroML, Double Delta, SHIFT-CCA.}
    \label{fig:all_surf}
\end{figure}

\begin{figure}
    \centering
    \includegraphics[width=\linewidth]{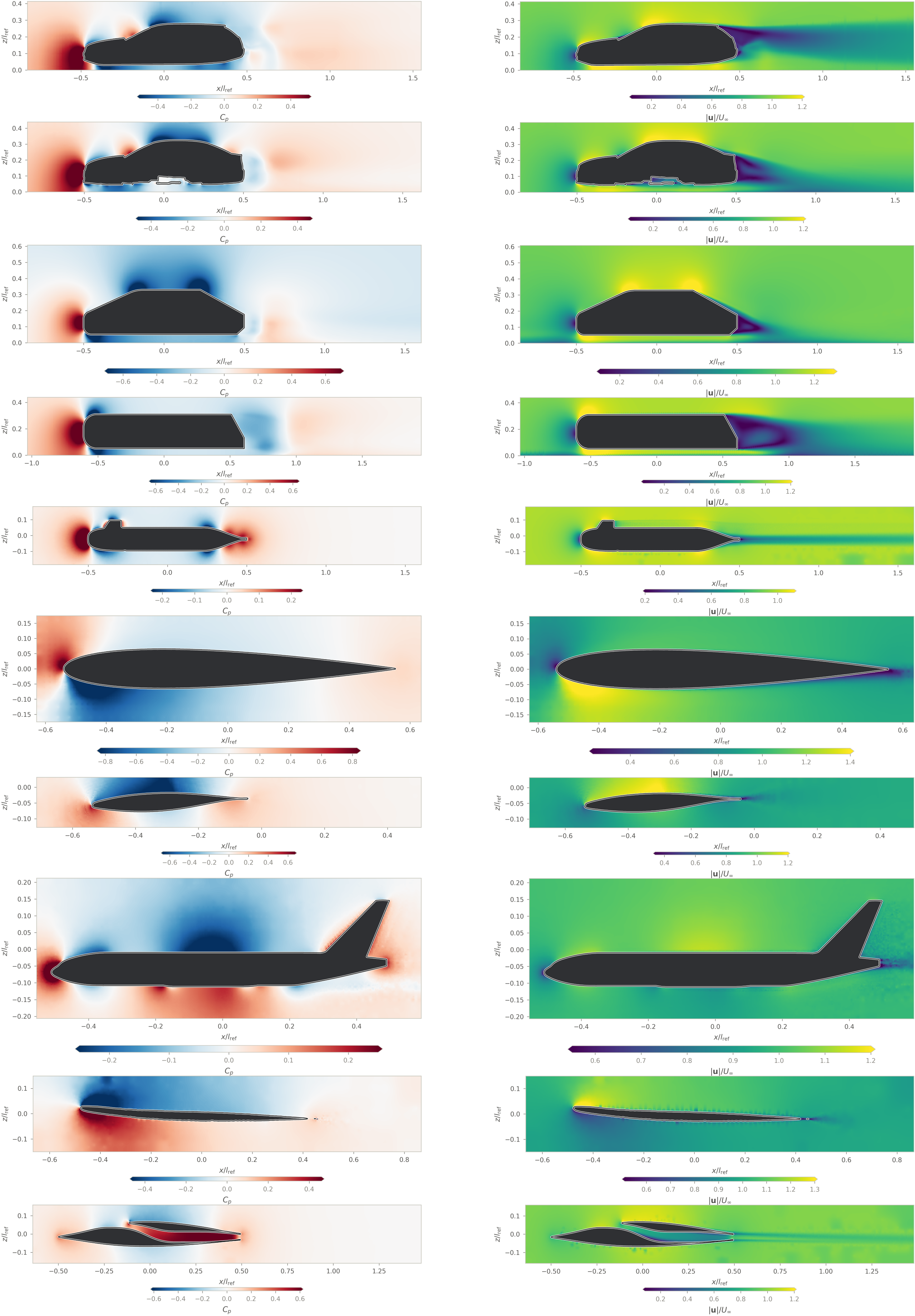}
    \caption{The volumetric pressure (shown as the coefficient of pressure \(C_p\)) and non-dimensional volumetric velocity magnitude \(|\vec{u}|/U_\infty\), plotted for a single sample in all datasets. From top to bottom: DrivAerNet, DrivAerML, WindsorML, AhmedML, SHIFT-Submarine, Emmi Wing, SuperWing, HiLiftAeroML, Double Delta, SHIFT-CCA.}
    \label{fig:all_vol}
\end{figure}

\paragraph{Overview.} We give additional details on each dataset, such as how the original data was generated, the relevant parameters, and any post-processing done for this work. Additionally, to visualize each dataset we plot the geometry, coefficient of pressure \(C_p\), and x-component of the skin friction coefficient \(C_{f_x}\) for a single sample in Figure \ref{fig:all_surf}. Furthermore, we plot the corresponding volumetric fields (coefficient of pressure \(C_p\) and non-dimensional velocity magnitude \(|u^*| = |\vec{u}|/U_\infty\)) for these samples in Figure \ref{fig:all_vol}. 

For all datasets, we crop the volume domain to exclude far-field regions where the velocity and pressure fields are effectively constant. During preprocessing, each sample is loaded from its raw format (e.g., .stl, .vtp, .vtu) and converted into a NumPy memory-mapped array (.npy). Before writing to disk, we randomly permute the points within each sample so that any contiguous slice of the memory map yields a random subset of points and their associated field values. This lets us draw random point subsets by reading contiguous slices, rather than loading entire samples (which can contain millions of points) into memory, and enables fast data loading.

Finally, the datasets can vary widely in their spatial discretization. High-resolution, scale-resolving simulations typically use local mesh refinement in regions of strong gradients, and some datasets fully resolve the boundary layer, which concentrates a large fraction of the volume mesh near the geometry. As a result, randomly sampling a batch of volume points can under-represent features in the freestream. To mitigate this and make the effective discretization more consistent across datasets, we adopt a sampling scheme that produces a more spatially uniform point distribution. This is a design choice that we ablate in Appendix \ref{app:discretization}. To sample \(N\) points, we first draw an oversampled set of \(kN\) points from storage, then select \(N\) points from this set with probability inversely proportional to the occupancy of their cell on a voxel grid. A final note is that normalization statistics are computed under this measure. 

\subsection{DrivAerNet++    ~\citep{elrefaie2025drivaernetlargescalemultimodalcar}}

\paragraph{Overview.} The dataset consists of 8000 parametrically morphed DrivAer passenger-car bodies. These consist of different base geometries (fastback, estateback and notchback), with open/closed and smooth/detailed wheels, and detailed or smooth underbodies, and are morphed with 26 geometric design parameters.

\paragraph{Data Generation.} Samples are solved with a steady, incompressible RANS solver with a
\(\kappa-\omega\) SST closure, using OpenFOAM v11, SIMPLE coupling, and a wall-function
(nutUSpaldingWallFunction) for the boundary layer. There are around 24 M cells per case
(500--750 k on the car surface). 7000 iterations are solved per case, with forces averaged over
the last 1000. 

\paragraph{Parameters.} \(U_\infty\) = 30 m/s, Re 8.37e6–1.01e7 (varying with car length), air at
\(\nu\) = 1.56e-5 \(m^2/s\), \(\rho\) = 1.184 \(kg/m^3\). 

\paragraph{Processing.} Certain samples assume a symmetric boundary condition and therefore only have +y values. To be consistent with DrivAerML, we mirror the geometry and fields to resemble a full car. Furthermore, we negate OpenFOAM's shear stress convention, such that positive wall-shear stresses point in the direction of flow. The volume domain is cropped based on the geometry bounding box; given a bounding box \([(x_{min}, y_{min}, z_{min}), (x_{max}, y_{max}, z_{max})]\) of size \(L = (L_x, L_y, L_z)\), the volume domain is cropped to \([x_{min} - 0.5L_x, x_{max} + 2.0L_x]\) (wake is +x), \([y_{min} - 0.75L_y, y_{max} + 0.75L_y]\) (symmetric) and  \([z_{min}, z_{max} + 0.5L_z]\) (no downward extension due to the floor). 

\subsection{DrivAerML ~\citep{ashton2025drivaermlhighfidelitycomputationalfluid}}

\paragraph{Overview.} The dataset consists of 500 morphs of the DrivAer notchback body, released
as time-averaged hybrid RANS/LES statistics rather than as a single converged solution. 

\paragraph{Data Generation.} Samples are solved with an SA-based \(\sigma\)-DDES: a
Spalart-Allmaras RANS branch with the Nicoud \(\sigma\)-model as the subgrid-scale operator, plus
Enhanced Protection shielding to keep LES content out of attached boundary layers, and Spalding
all-\(y^+\) wall functions. The solver is a custom \verb|pimpleFoam| derivative (OpenFOAM v2212)
with a modified Rhie-Chow interpolation and second-order implicit Euler time integration. Meshes
carry around 160 M cells, with a first boundary-layer height of 0.75 mm and seven layers to
12 mm. Statistics are accumulated until the drag estimate settles to \(\pm 1.5\) drag counts,
generally 40--60 convective time units.

\paragraph{Parameters.} \(U_\infty\) = 38.889 m/s, Re = 7.2e6 based on the 2.786 m wheelbase
(1.2e7 if length-based), air at \(\nu\) = 1.507e-5 \(m^2/s\), \(\rho\) = 1.2041 \(kg/m^3\). 

\paragraph{Processing.} The ground plane sits
at \(z \approx -0.318\) rather than \(z = 0\), but  subtracting the geometry centroid during normalization accounts for this. The volume field is cropped identically to DrivAerNet samples. OpenFOAM's shear stress sign convention is also flipped such that positive wall-shear stresses point in the direction of flow. 

\subsection{AhmedML ~\citep{ashton2024ahmedmlhighfidelitycomputationalfluid}}

\paragraph{Overview.} The dataset consists of 500 variants of the Ahmed body, the canonical
simplified automotive bluff body, spanning slant angle, body length, width and height, and ground
clearance.

\paragraph{Data Generation.} Samples are solved with a plain SA-DDES using the standard \(f_d\)
shielding function, similar to DrivAerML, but
without the \(\sigma\)-model and Enhanced Protection. Meshes carry 15--21 M prismatic and hex-dominant cells and are coarse at the wall; there are three prism layers yielding \(y^+ \approx 50\) and the boundary layer is modeled with Spalding wall
functions. Each case runs for 40 convective transit times with time-averaging beginning after the
first 10. 

\paragraph{Parameters.} \(U_\infty\) = 1 m/s, with Re = 7.68e5 based on the body height (2.78e6 if
length-based). The density and viscosity are set at \(\rho\) = 1 \(kg/m^3\) and \(\nu\) = 3.75e-7 \(m^2/s\).

\paragraph{Processing.} The tunnel floor is at
\(z = 0\), but the Ahmed body is offset on stilts so the lowest geometry point sits at
\(z \approx 0.05\); the volume crop therefore clamps its lower bound down to the ground plane
rather than to the surface bounding box. Otherwise the volume crop is identical to that of DrivAerNet/DrivAerML. OpenFOAM's shear stress sign convention is flipped as well. 

\subsection{WindsorML ~\citep{ashton2025windsormlhighfidelitycomputationalfluid}}

\paragraph{Overview.} The dataset consists of 355 variants of the Windsor body, a squareback
automotive reference geometry, released as time-averaged wall-modelled LES. It is the dataset with no RANS anywhere in the closure.

\paragraph{Data Generation.} Samples are solved with wall-modelled LES using the constant-coefficient Vreman subgrid-scale model and an equilibrium wall model imposed as a
shear stress constraint. The
solver (Volcano ScaLES) is compressible and explicit, nominally fourth-order in space with SSP-RK3 time stepping, on Cartesian octree grids with an immersed-boundary geometry representation. The baseline grid carries 275 M cells at 0.75 mm minimum
spacing, and fields are time-averaged. 

\paragraph{Parameters.} \(U_\infty\) = 40 m/s, Re = 2.9e6 based on the 1.044 m body length. The density and viscosity are set to \(\rho\) = 1.2712 \(kg/m^3\) and \(\nu\) = 1.44e-5 \(m^2/s\).

\paragraph{Processing.} The native dataset axes differ from other datasets: \(x\) is streamwise but \(y\) is vertical, with the floor at \(y \approx 0\),
and \(z\) is lateral. Therefore we apply a signed permutation \((x, y, z) \mapsto (x, -z, y)\) to
positions, the surface shear stress and to the volume velocity, which brings the dataset
into the shared frame. An identical volume crop to the prior automotive datasets is applied. The dataset did not ship with a freestream pressure, so we compute one at \(p_\infty\) = 37340.5 Pa to compute non-dimensional values. 

\subsection{SHIFT-Submarine ~\citep{shift-sub}}

\paragraph{Overview.} The dataset consists of 101 parameterized Myring-type submarine hulls, varied
over more than twenty shape parameters covering cross-section, fore- and aft-body profiles,
sail and fin placement, and length-to-diameter ratio. It is the only dataset that is in water.

\paragraph{Data Generation.} Specific details are not disclosed, but we assume that samples are solved as steady RANS with a Spalart-Allmaras closure on the Luminary finite-volume solver. Meshes carry around 5.6 M volume points and 32 M tetrahedra, with about 0.8 M surface points per case.

\paragraph{Parameters.} \(U_\infty\) = 5 m/s in water, \(\rho\) = 998 \(kg/m^3\),
\(\nu \approx\) 1.0e-6 \(m^2/s\), hull length 6.455 m, so Re \(\approx\) 3.2e7. 

\paragraph{Processing.} The raw wall-shear stress seems to have numerical artifacts at the strake and sail edges, with \(\max|C_f| \approx 0.10\) against a 99.9th percentile of
0.013, so we drop points beyond \(\pm 3\sigma\) on any shear component, computed per sample. The volume crop is isotropic and symmetric in the y-z
plane, with no floor treatment, since the body is fully submerged. Given a bounding box \([(x_{min}, y_{min}, z_{min}), (x_{max}, y_{max}, z_{max})]\) of size \(L = (L_x, L_y, L_z)\), the volume domain is cropped to \([x_{min} - 0.5L_x, x_{max} + 2.0L_x]\), \([y_{min} - 0.75L_y, y_{max} + 0.75L_y]\), and  \([z_{min}-0.75L_z, z_{max} + 0.75L_z]\). 

\subsection{Emmi-Wing ~\citep{paischer2026goingspeedsoundpushing}}

\paragraph{Overview.} The dataset consists of 29{,}609 steady compressible solutions over
parameterized tapered swept wings, described by four geometric parameters (root chord, span, taper
ratio and sweep) and two inflow parameters.

\paragraph{Data Generation.} Samples are solved with OpenFOAM v2506 \verb|rhoSimpleFoam| assuming a steady, compressible, perfect gas, with a Spalart-Allmaras closure and second-order schemes. Body-fitted
\verb|snappyHexMesh| grids carry prismatic boundary layers at \(y^+ \in [50, 200]\) and the boundary layer is modeled rather than resolved.  Meshes hold around 3.3 M volume points and 46--114 k surface points. 

\paragraph{Parameters.} The freestream state is computed to be:
\(p_\infty\) = 100000 Pa, \(\rho_\infty\) = 1.16639 \(kg/m^3\) and \(T_\infty\) = 298 K with
\(R\) = 287.7 \(J/(kg\,K)\). \(U_\infty\) varies, over 150--300 m/s, giving Mach 0.437--0.875 and Re 5--20e6 at angles of attack of \(-10^\circ\) to
\(10^\circ\). Angle-of-attack is modeled from boundary conditions, such that the inlet velocity is \(U_\infty[\cos\alpha, 0, \sin\alpha]\).

\paragraph{Processing.} The original authors publish 62 erroneous cases that were due to numerical instability, so we exclude these. We also negate OpenFOAM's shear stress convention. Given a bounding box \([(x_{min}, y_{min}, z_{min}), (x_{max}, y_{max}, z_{max})]\) of size \(L = (L_x, L_y, L_z)\), the volume domain is cropped to \([x_{min} - 0.5L_x, x_{max} + 1.5L_x]\) (streamwise), \([y_{min} - 0.15L_y, y_{max} + 0.35L_y]\) (spanwise) and  \([z_{min}-0.35L_z, z_{max} + 0.35L_z]\) (vertical).

\subsection{SuperWing ~\citep{yang2026superwingcomprehensivetransonicwing}}

\paragraph{Overview.} The dataset consists of 28{,}856 transonic solutions over 4239 kinked,
double-swept wing planforms, each simulated at up to eight operating conditions, under 56 design
variables per wing. 

\paragraph{Data Generation.} Samples are solved with ADflow as steady compressible RANS with a
Spalart-Allmaras closure. The mesh is a
structured grid that is extruded from the surface, with wall-normal layers at \(y^+ \approx 1\), leading to the boundary layer being fully resolved. The solution uses a steady-state multigrid solver, with around 44k surface points and 3.1M volume cells per sample.  

\paragraph{Parameters.} Mach 0.75--0.90 at around fixed Re = 2.0e7 and \(T_\infty\) = 300 K, giving
\(U_\infty \approx\) 260--313 m/s, at angles of attack of \(2^\circ\) to \(12^\circ\). 

\paragraph{Processing.} The raw data frame has the chord along \(+x\), span along \(z\), and vertical along \(+y\), with \(+y\) being the suction side of the wing (low pressure); we therefore apply a signed permutation
\((x, y, z) \mapsto (x, -z, y)\), which places negative \(C_p\) on the upper surface of
a lifting wing. The volume mesh is highly dense due to being a wall-resolved simulation and around half of all volume points lie within 0.0085 chord of the wall and layer 0 reproduces the surface field
exactly, however, we do not modify this. We prune wall-shear stress at \(3\sigma\) per
sample on all three components to remove outliers. Given a bounding box \([(x_{min}, y_{min}, z_{min}), (x_{max}, y_{max}, z_{max})]\) of size \(L = (L_x, L_y, L_z)\), the volume domain is cropped to \([x_{min} - 0.25L_x, x_{max} + 0.75L_x]\) (streamwise), \([y_{min} - 0.20L_y, y_{max} + 0.02L_y]\) (spanwise, +y is base of wing) and \([z_{min}-0.20L_z, z_{max} + 0.20L_z]\) (vertical).

\subsection{Double-Delta ~\citep{Shen_2026}}

\paragraph{Overview.} The dataset consists of 2448 solutions over 272 parametric cranked-delta
wings, each at nine angles of attack in \(1^\circ\) steps. It models leading-edge vortices and their breakdown at large angles of attack, and has large pressure/wall-shear deviations. 

\paragraph{Data Generation.} Samples are solved with SU2 as steady compressible RANS using SA-R, the
Spalart-Allmaras model with a rotation and curvature correction. The convective scheme is JST with Green-Gauss reconstruction, integrated with implicit Euler at adaptive CFL; 57 runs required an AUSM scheme instead to converge and are
flagged. Meshes are wall-resolved at a mean \(y^+\) of 1.66, with around \(1.18\)e5 surface elements
and \(6.96\)e6 volume cells of mixed tetrahedra, prisms and pyramids. 

\paragraph{Parameters.} Mach 0.3 with \(p_\infty\) = 71833.4 Pa, \(\rho_\infty\) = 0.878035
\(kg/m^3\), \(|U_\infty|\) = 101.53 m/s, \(q_\infty\) = 4525.54 Pa and \(T_\infty\) = 285 K, giving
Re = 8.04e7 on a 16 m mean aerodynamic chord, at angles of attack of \(11^\circ\) to \(19^\circ\).

\paragraph{Processing.} There are a few duplicate runs that are excluded. Due to being a wall-resolved simulation, the majority (\(>90\%\)) of mesh points lie in a thin slab around the wing, but we do not modify this. Surface
pressure is pruned based on the stagnation bound \(C_p \le C_{p,0} = 1.0227\) at
Mach 0.3 and a two-sided \(3\sigma\) range, which removes around 2\% of surface points. This clips the surface \(C_p\) minimum from \(-15.9\) to about \(-5\). The volume, which reaches
\(C_p = -9.7\), is not pruned. Splits are the dataset's native, geometry-disjoint ones: the 256
training-set geometries (2304 samples) against the 16 holdout geometries (144 samples). Given a bounding box \([(x_{min}, y_{min}, z_{min}), (x_{max}, y_{max}, z_{max})]\) of size \(L = (L_x, L_y, L_z)\), the volume domain is cropped to \([x_{min} - 0.25L_x, x_{max} + 0.75L_x]\) (streamwise), \([y_{min} - 0.35L_y, y_{max} + 0.35L_y]\) (spanwise) and  \([z_{min}-0.2L_z, z_{max} + 0.2L_z]\) (vertical).

\subsection{HiLiftAeroML ~\citep{ashton2026hiliftaeromlhighfidelitycomputationalfluid}}

\paragraph{Overview.} The dataset consists of 1787 wall-modelled LES solutions over the NASA CRM-HL
high-lift configuration. There are 180 geometry variants, described by eight design variables covering slat and flap deflections and gap multipliers inboard and outboard, each at ten angles of attack in
\(2^\circ\) steps. It models high-lift aerodynamics through flow separation and is the
largest dataset we use in terms of size and points per sample.

\paragraph{Data Generation.} The
geometry is a half-domain with a symmetry plane at \(y = 0\) and positions in inches. Samples are solved with Fidelity Charles, an explicit unstructured
finite-volume solver, second-order in space and third-order in time, using a dynamic Smagorinsky
subgrid-scale model, whose coefficient is computed dynamically in space and time, and an equilibrium
wall model solving simplified boundary-layer equations on an embedded near-wall grid. Voronoi meshes
carry 300--500 M cells, reduced to 150--300 M on export. Statistics are gathered over 15 convective
time units following a 10 unit transient below \(12^\circ\), and over 30--200 units at the separated
high-AoA conditions. Meshes are adapted per angle of attack to improve accuracy, so a geometry does not reuse one grid. 

\paragraph{Parameters.} Mach 0.2 at Re = 1.6e6, with a mean aerodynamic chord of
275.80 in (7.005 m) and a half-span of 1156.75 in.

\paragraph{Processing.} Given the large size of the dataset, we downsample the downloaded fields and meshes by a factor of two. Training and validation splits are grouped by geometry (90/10). Given a bounding box \([(x_{min}, y_{min}, z_{min}), (x_{max}, y_{max}, z_{max})]\) of size \(L = (L_x, L_y, L_z)\), the volume domain is cropped to \([x_{min} - 0.15L_x, x_{max} + 0.5L_x]\) (+x is wake), \([y_{min} - 0.05L_y, y_{max} + 0.15L_y]\) (-y is symmetry plane) and  \([z_{min}-L_z, z_{max} + L_z]\) (vertical)

\subsection{SHIFT-CCA ~\citep{shift-cca}}

\paragraph{Overview.} The dataset consists of 99 variants of a Group-5 UAV reference configuration
representing collaborative combat aircraft concepts, morphed in root chord, panel break position,
leading- and trailing-edge sweep and wing-tip geometry. All 99 geometries are distinct and share a single physical regime, and it models transonic flight with a shock on the wing.

\paragraph{Data Generation.} Specific details are not disclosed, but we assume that samples are solved as steady RANS with a Spalart-Allmaras closure on the Luminary finite-volume solver. Meshes carry around 5.55 M volume points and 32.1 M
tetrahedra, with about 1.5 M surface cells per case.

\paragraph{Parameters.} Mach 0.72 at \(p_\infty\) = 18754 Pa, \(T_\infty\) = 216.65 K,
\(\rho_\infty\) = 0.30155 \(kg/m^3\) and \(U_\infty\) = 212.452 m/s, Re \(\approx\) 3.1e7 and the
angle of attack is \(2^\circ\).

\paragraph{Processing.} We
prune surface shear at \(4\sigma\) per sample on all three components, removing about 2\% of points. Given a bounding box \([(x_{min}, y_{min}, z_{min}), (x_{max}, y_{max}, z_{max})]\) of size \(L = (L_x, L_y, L_z)\), the volume domain is cropped to \([x_{min} - 0.25L_x, x_{max} + L_x]\) (+x is wake), \([y_{min} - 0.35L_y, y_{max} + 0.35L_y]\) and  \([z_{min}-0.2L_z, z_{max} + 0.2L_z]\) (vertical). 

\section{Model Details}
\label{app:models}

\begin{table}[t]
    \centering
    \small
    \addtolength{\tabcolsep}{-0.4em}
    \begin{tabular}{llllllll}
        \toprule
        \multicolumn{2}{c}{SMART} & \multicolumn{2}{c}{SMART-IC} & \multicolumn{2}{c}{AB-UPT} & \multicolumn{2}{c}{Transolver++} \\
        \cmidrule(lr){1-2} \cmidrule(lr){3-4} \cmidrule(lr){5-6} \cmidrule(lr){7-8}
        Parameter & Value & Parameter & Value & Parameter & Value & Parameter & Value \\
        \midrule
        Model size        & 32.0M & Model size        & 32.4M & Model size          & 36.1M & Model size     & 29.1M \\
        Hidden dim        & 384   & Hidden dim        & 384   & Hidden dim          & 384   & Hidden dim     & 384   \\
        Heads             & 8     & Heads             & 8     & Heads               & 8     & Heads          & 8     \\
        Enc./dec. blocks  & 12    & Enc./dec. blocks  & 8     & Shared blocks       & \texttt{ppscscs} & Layers & 12 \\
        Num latents       & 4096  & Num latents       & 4096  & Supernodes          & 1024  & Slices         & 32    \\
        Geom.\ pts / block & 4096 & Geom.\ pts / block & 4096 & Geometry blocks     & 2     & MLP ratio      & 4     \\
        --                & --    & Context pts / block & 8192 & Surf.\ / vol.\ blocks & 3 / 3 & --  & --  \\
        --                & --    & --    & --   & Surf.\ / vol.\ anchors & 4096 / 4096 & -- & --\\
        \bottomrule
    \end{tabular}
    \caption{Model hyperparameters. All models use a hidden width of 384 and are matched at roughly 30M parameters. The optimizer is also constant across all models (Adam ($\beta_1{=}0.9$, $\beta_2{=}0.999$), learning rate $10^{-4}$, StepLR $\gamma{=}0.99$ every 1k steps) }
    \label{tab:hyperparameters}
\end{table}

\begin{figure}[t]
    \centering
    \includegraphics[width=\linewidth]{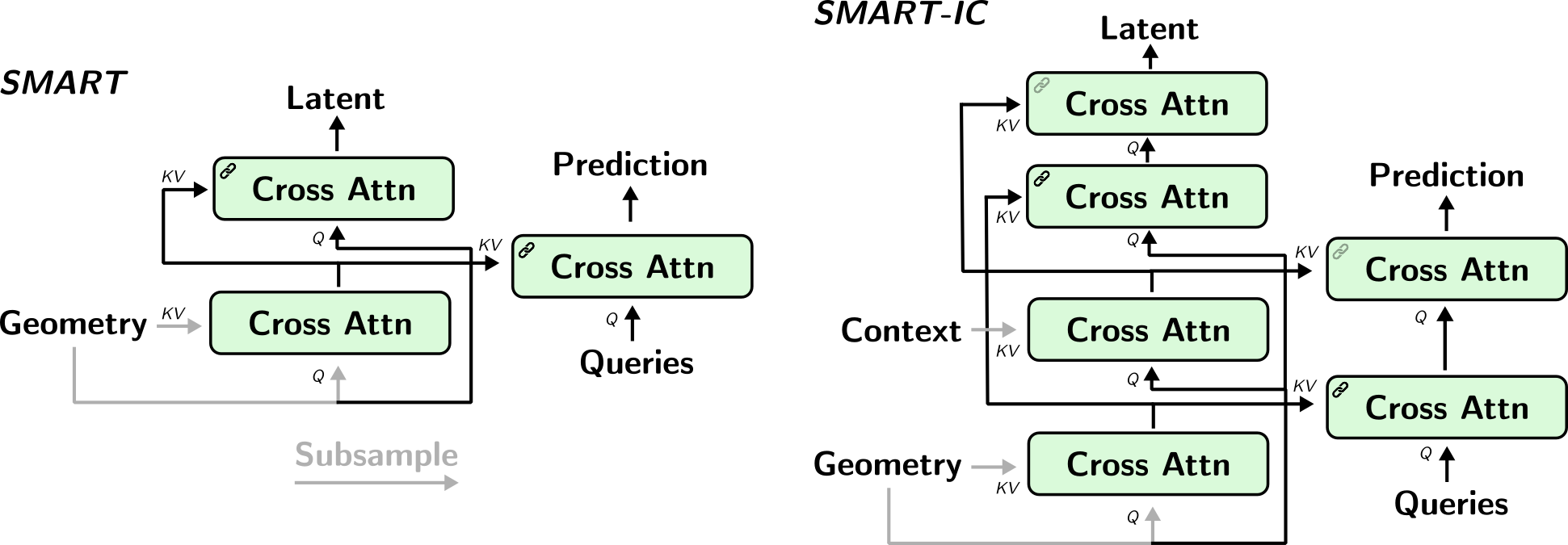}
    \caption{Schematic of the SMART model variants. Cross-attention is routed so that the encoder and decoder branches can share weights (link symbols). Gray arrows indicate subsampling. Simulation parameters \(\theta\) enter through Feature-wise Linear Modulation (FiLM) layers \citep{perez2017filmvisualreasoninggeneral}}
    \label{fig:smart_models}
\end{figure}

\begin{figure}[t]
    \centering
    \includegraphics[width=\linewidth]{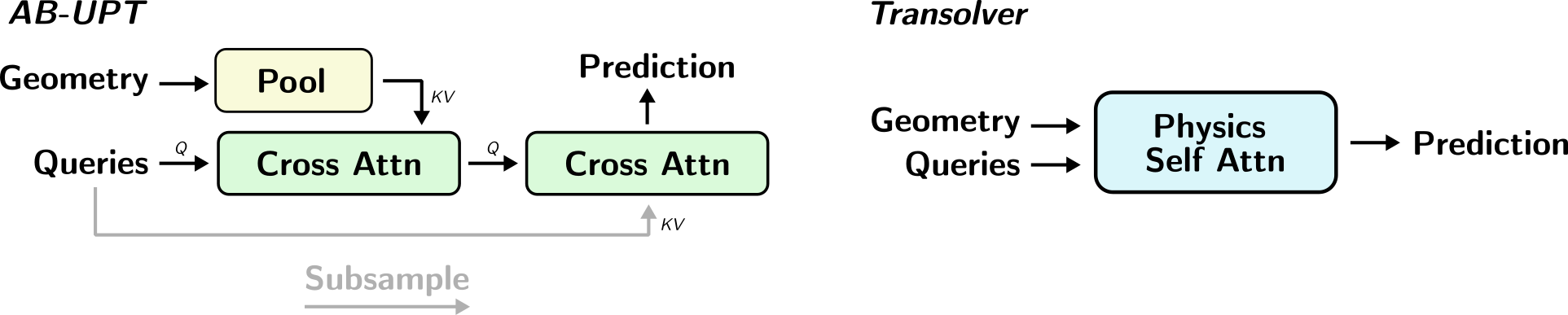}
    \caption{Schematics of AB-UPT and Transolver. Gray arrows indicate subsampling. For clarity, branches and weight sharing in AB-UPT, and slicing/unslicing in Transolver, are not shown. Simulation parameters \(\theta\) enter through Adaptive LayerNorm (AdaLN) layers.}
    \label{fig:other_models}
\end{figure}

An overall figure comparing the main aspects of each architecture is shown in Figures \ref{fig:smart_models} and \ref{fig:other_models}. Additionally, we list the main hyperparameters of each model in Table \ref{tab:hyperparameters}. We briefly describe the main features of each architecture as well as any modifications we implement; however, for specific details we encourage readers to look at the original papers. 

\paragraph{SMART.} SMART \citep{hagnberger2026smartscalablemeshfreeaerodynamic} is a transformer-based surrogate model designed to approximate solutions on dense, unstructured meshes. Its central mechanism is to ensure that query point \(x\) attends to latent representations of the geometry \(G\) rather than to other queries, using cross attention. Avoiding interaction among queries has two benefits. First, it keeps computational cost under control: because queries attend only to the geometry encoding, whose discretization and sampling we choose, the cost no longer scales with the large number of query points. Second, it preserves discretization invariance, since the prediction at a given point does not change with the number of queries. We make a few slight improvements to the original SMART architecture. We replace MLP layers with SwiGLU \citep{shazeer2020gluvariantsimprovetransformer}, as well as use separate output heads for the volume and surface fields. 

\paragraph{SMART-IC.}
We build on the base SMART model to allow it to use an additional context simulation as input. We treat the context simulation as a surface and volume field, and draw samples from it to update the latent state in SMART. This latent state is additionally used in the decoder branch to propagate information to the queries. We assume that the surface field in the context contains information about the context geometry, so it is not additionally provided. However, both the context simulation parameters and the query simulation parameters \(\theta_c, \theta_q\) are provided. 

\paragraph{AB-UPT.}
Anchored Branch Universal Physics Transformers (AB-UPT) \citep{alkin2025abuptautomotiveaerospaceapplications} is another transformer-based surrogate model for approximating solutions on dense, unstructured meshes. Its main mechanism for reducing computational cost is to designate a subsampled set of query points as anchors; the remaining queries then obtain information from these anchors through cross-attention. Because the anchors are sampled randomly, each query depends only on a random subset of points rather than on all other queries. This makes predictions independent of the full query set and bounds the computational cost. In addition, global geometric information is incorporated through a pooling step followed by a cross-attention layer. 

We make several modifications to improve model performance. First, we replace the radius pooling in the geometry branch with a pooling mechanism based on ball-tree decomposition \citep{zhdanov2025erwintreebasedhierarchicaltransformer}. Rather than selecting a random set of supernodes and pooling all points within a fixed radius of each, we partition the geometry point cloud into balls and pool the information within each ball to its centroid. Second, we replace the MLP layers with SwiGLU feed-forward layers. 

\paragraph{Transolver++.}
Transolver++ \citep{luo2025transolveraccurateneuralsolver} builds on Transolver \citep{wu2024transolverfasttransformersolver}, which introduced an efficient physics-attention mechanism for handling large input meshes. In this mechanism, a learned projection aggregates the input points onto a fixed set of slices, and self-attention is then applied among these slices. While this reduces computational cost, the resulting representation remains tied to the discretization of the input.

In our implementation, we assume that the geometry information is contained in the surface queries, which are combined with the volume queries to form the model input. We additionally introduce adaptive layer normalization (AdaLN) to incorporate conditioning information into the model. Finally, as with the other models, we replace the MLP layers with SwiGLU feed-forward layers and use separate output heads for the surface and volume predictions.

\iffalse 
\begin{figure}
    \centering
    \includegraphics[width=\linewidth]{figures/comparison.png}
    \caption{\textit{Left, A}: The closest geometry from DrivaerML/Net overlaid, with a chamfer distance of 37.5 mm (random samples within DrivaerML/Net have average chamfer distances of 49.8/41.6mm) \textit{Middle, B}: Surface \(c_p\) along the centerline (green plane), averaged over the 25 closest pairs. \textit{Right, C}: Average volumetric velocity at the wake (purple plane), averaged across the 25 closest pairs.}
    \label{fig:comparison}
\end{figure}
\fi 

\section{Additional Experiments}
\label{app:additional_results}

\subsection{Different Architectures}
\label{app:gen_results}

\begin{table}[t]
    \centering
    \small
    \addtolength{\tabcolsep}{-0.375em}
    \begin{tabular}{@{}l?cccc?cccc?cccc?cccc@{}}
        \toprule
        %& \multicolumn{16}{c}{Held-out target dataset} \\
        %\cmidrule(lr){2-17}
        \rlap{Held-out Datasets:} & \multicolumn{4}{c}{AhmedML}
        & \multicolumn{4}{c}{SHIFT-Sub}
        & \multicolumn{4}{c}{SHIFT-CCA}
        & \multicolumn{4}{c}{HiLiftAeroML} \\
        \cmidrule(lr){2-5}
        \cmidrule(lr){6-9}
        \cmidrule(lr){10-13}
        \cmidrule(lr){14-17}
        \rlap{\(\#\) FT Samples:} 
        & 0 & 2 & 4 & 8
        & 0 & 2 & 4 & 8
        & 0 & 2 & 4 & 8
        & 0 & 2 & 4 & 8 \\
        \midrule

        \multicolumn{17}{@{}l}{\textit{No Pretraining}} \\[-0.15em]
        Random init.
        & 0.68 & 0.35 &  0.33    & 0.25 & 0.65 & 0.32 & 0.28 & 0.30 & 0.69 & 0.28 & 0.27 & 0.26 & 0.82 & 0.63 & 0.63 & 0.61 \\

        \addlinespace[0.25em]
        \multicolumn{17}{@{}l}{
            \textit{Domain Experts}
        } \\[-0.15em]

        DrivAerNet++    & 0.61 & 0.24 & 0.21 & 0.16 & 0.62 & 0.28 & 0.23 & 0.17 & 0.71 & 0.19 & 0.16 & 0.16 & 0.90 & 0.53 & 0.48 & 0.48 \\
        
        Emmi-Wing     & 0.74 & 0.24 & 0.22 & 0.19 & 0.68 & 0.29 & 0.24 & \worst{0.22} & 0.70 & \worst{0.20} & \worst{0.18} & \worst{0.17} & 1.02 & \worst{0.60} & \worst{0.54} & \worst{0.53} \\
        
        SuperWing     & \worst{1.61} & 0.23 & 0.22 & 0.18 & \worst{1.35} & \worst{0.30} & \worst{0.24} & 0.22 & 0.69 & 0.18 & 0.16 & 0.15 & \worst{2.76} & 0.57 & 0.54 & 0.53 \\
        
        DrivAerML     & 0.64 & \worst{0.26} & 0.21 & 0.17 & 0.62 & 0.27 & 0.22 & 0.17 & \worst{0.79} & 0.20 & 0.16 & 0.15 & 0.95 & \best{0.52} & 0.45 & 0.45 \\
        
        WindsorML     & \best{0.37} & \best{0.19} & \best{0.15} & \best{0.11} & \best{0.58} & \best{0.27} & \best{0.20} & \best{0.14} & \best{0.69} & 0.20 & 0.16 & 0.16 & 0.90 & 0.56 & 0.50 & 0.51 \\
        
        Double-Delta  & 0.73 & 0.24 & \worst{0.24} & \worst{0.19} & 0.68 & 0.27 & 0.23 & 0.19 & 0.76 & \best{0.14} & \best{0.12} & \best{0.12} & \best{0.84} & 0.54 & \best{0.44} & \best{0.42} \\

        \addlinespace[0.25em]
        \multicolumn{17}{@{}l}{\textit{Cross-Domain}} \\[-0.15em]

        Joint Model    & \textbf{0.31} & \textbf{0.14} & \textbf{0.11} & \textbf{0.10} & \textbf{0.53} & \textbf{0.23} & \textbf{0.18} & \textbf{0.13} & \textbf{0.61} & \textbf{0.13} & \textbf{0.11} & \textbf{0.11} & \textbf{0.82} & \textbf{0.48} & \textbf{0.42} & \textbf{0.39} \\
        \bottomrule
    \end{tabular}

    \caption{Validation error for \textit{AB-UPT} on four held-out target datasets.
    Rows indicate the model initialization; single-domain experts are
    identified by their pretraining dataset. Columns give the number of
    fine-tuning samples, with 0 denoting zero-shot evaluation. Green and
    red indicate the best and worst single-domain experts, respectively,
    and bold indicates the lowest overall error.}
    \label{tab:abupt_generalization}
\end{table}

\begin{table}[t]
    \centering
    \small
    \addtolength{\tabcolsep}{-0.375em}
    \begin{tabular}{@{}l?cccc?cccc?cccc?cccc@{}}
        \toprule
        %& \multicolumn{16}{c}{Held-out target dataset} \\
        %\cmidrule(lr){2-17}
        \rlap{Held-out Datasets:} & \multicolumn{4}{c}{AhmedML}
        & \multicolumn{4}{c}{SHIFT-Sub}
        & \multicolumn{4}{c}{SHIFT-CCA}
        & \multicolumn{4}{c}{HiLiftAeroML} \\
        \cmidrule(lr){2-5}
        \cmidrule(lr){6-9}
        \cmidrule(lr){10-13}
        \cmidrule(lr){14-17}
        \rlap{\(\#\) FT Samples:} 
        & 0 & 2 & 4 & 8
        & 0 & 2 & 4 & 8
        & 0 & 2 & 4 & 8
        & 0 & 2 & 4 & 8 \\
        \midrule

        \multicolumn{17}{@{}l}{\textit{No Pretraining}} \\[-0.15em]
        Random Init.  & 0.68 & 0.36 & 0.34 & 0.30 & 0.64 & 0.31 & 0.29 & 0.27 & 0.68 & 0.22 & 0.21 & 0.20 & 0.81 & 0.61 & 0.59 & 0.58 \\

        \addlinespace[0.25em]
        \multicolumn{17}{@{}l}{
            \textit{Domain Experts}
        } \\[-0.15em]

        DrivAerNet++    & 0.59 & 0.27 & 0.27 & 0.19 & \best{0.57} & 0.28 & 0.24 & 0.22 & 0.79 & 0.21 & 0.19 & 0.18 & 0.97 & 0.56 & 0.46 & 0.51 \\
        Emmi-Wing     & 0.70 & 0.27 & 0.24 & 0.21 & 0.66 & \worst{0.29} & 0.24 & 0.22 & \best{0.65} & 0.21 & 0.18 & 0.18 & 1.01 & 0.58 & 0.50 & 0.49 \\
        SuperWing     & \worst{0.73} & 0.36 & 0.31 & 0.23 & \worst{0.71} & 0.29 & 0.25 & \worst{0.25} & 0.78 & 0.19 & 0.16 & 0.16 & 0.96 & \best{\textbf{0.50}} & \best{0.43} & \best{0.41} \\
        DrivAerML     & 0.62 & 0.28 & 0.25 & 0.19 & 0.63 & \best{0.27} & 0.24 & \best{0.20} & \worst{0.87} & 0.20 & 0.18 & 0.17 & 1.01 & 0.55 & 0.47 & 0.47 \\
        WindsorML     & \best{0.47} & \best{0.26} & \best{0.19} & \best{0.16} & 0.60 & 0.28 & \best{\textbf{0.23}} & 0.21 & 0.77 & \worst{0.22} & \worst{0.20} & \worst{0.20} & \worst{1.06} & \worst{0.58} & \worst{0.54} & \worst{0.56} \\
        Double-Delta  & 0.70 & \worst{0.37} & \worst{0.34} & \worst{0.24} & 0.66 & 0.29 & \worst{0.25} & 0.23 & 0.76 & \best{0.19} & \best{0.15} & \best{0.15} & \best{\textbf{0.88}} & 0.58 & 0.48 & 0.47 \\

        \addlinespace[0.25em]
        \multicolumn{17}{@{}l}{\textit{Cross-Domain}} \\[-0.15em]

        Joint Model       & \textbf{0.40} & \textbf{0.15} & \textbf{0.12} & \textbf{0.11} & \textbf{0.55} & \textbf{0.26} & \textbf{0.23} & \textbf{0.18} & \textbf{0.62} & \textbf{0.17} & \textbf{0.14} & \textbf{0.13} & 0.91 & 0.53 & \textbf{0.42} & \textbf{0.39} \\
        \bottomrule
    \end{tabular}

    \caption{Validation error for \textit{Transolver++} on four held-out target datasets.
    Rows indicate the model initialization; single-domain experts are
    identified by their pretraining dataset. Columns give the number of
    fine-tuning samples, with 0 denoting zero-shot evaluation. Green and
    red indicate the best and worst single-domain experts, respectively,
    and bold indicates the lowest overall error.}
    \label{tab:transolver_generalization}
\end{table}

\paragraph{Generalization to Unseen Datasets.}
We repeat the main experiments in Section \ref{sec:generalization} for AB-UPT and Transolver++ models. AB-UPT and Transolver++ models are instantiated to around the same size (\(\sim\!\)30M parameters) and trained for the same budget (100k steps). After pretraining the expert and joint models, they are finetuned to the held-out datasets using \(S\in \{0, 2, 4, 8\}\) finetuning samples. The resulting errors are shown in Table \ref{tab:abupt_generalization} for AB-UPT and Table \ref{tab:transolver_generalization} for Transolver++. Similar to the main text, we observe that cross-domain pretraining improves downstream performance compared to any of the expert models or training from scratch. The overall trends and performance gains are also generally consistent with the main text. 

\begin{table}[t]
    \centering
    \small
    \addtolength{\tabcolsep}{-0.4em}
    \begin{tabular}{@{}llccccccc}
        \toprule
            & Model & DrivAerNet & DrivAerML & Emmi-Wing & WindsorML & SuperWing & Double-Delta & Mean \\
        \midrule
        \multirow{3}{*}{Expert}
            & SMART        & \textbf{0.142} & \textbf{0.047} &          0.030  & \textbf{0.065} & \textbf{0.047} & \textbf{0.055} & \textbf{0.064} \\
            & AB-UPT       &          0.148 &          0.057 &          0.030  &          0.068 &          0.058 &          0.064 &          0.071 \\
            & Transolver++ &          0.150 &          0.067 & \textbf{0.029}  &          0.074 &          0.063 &          0.071 &          0.076 \\
        \midrule
        \multirow{3}{*}{Joint}
            & SMART        & \textbf{0.173} & \textbf{0.134} &          0.041  & \textbf{0.074} & \textbf{0.082} & \textbf{0.090} & \textbf{0.099} \\
            & AB-UPT       &          0.177 &          0.148 &          0.041  &          0.080 &          0.089 &          0.097 &          0.105 \\
            & Transolver++ &          0.184 &          0.166 &          0.044  &          0.097 &          0.107 &          0.124 &          0.120 \\
        \bottomrule
    \end{tabular}
    \caption{SMART, AB-UPT, and Transolver++ models, evaluated on the pretraining datasets. Note that each cell in the Expert block is a different model (for example, the DrivAerNet SMART expert), while each row in the Joint block is a single model evaluated across all datasets.}
    \label{tab:pretraining}
\end{table}

\paragraph{Pretraining Performance.} 
We additionally report the performance of all expert and joint models on the pretraining validation sets, in Table \ref{tab:pretraining}. In general, SMART seems to be the best model, but every model performs quite well and the differences between architectures are marginal. This also corroborates prior studies \citep{alkin2025abuptscalingneuralcfd, hagnberger2026smartscalablemeshfreeaerodynamic}, where SMART/AB-UPT models are shown to outperform Transolver. However, the main benefit is that these cross attention-based architectures have query independence, such that they can be queried sequentially on a much denser mesh during inference without a large increase in error or memory. 

\subsection{Global Normalization}
\label{app:no_norm}

\paragraph{Overview.} A frequently overlooked problem in finetuning is that accurate normalization statistics cannot be assumed in practice. Large existing datasets contain enough samples to compute reliable means and variances, but finetuning aims to use only a handful of samples, where these statistics might not be well-estimated. However, we can leverage physical priors to help address this. Most CFD datasets solve the Navier–Stokes equations, which admit well-known non-dimensional quantities. These quantities are anchored to reference quantities set by the simulation, most commonly the freestream velocity, pressure, and density. Because these reference values are known at deployment, we use them to compute non-dimensional coefficients and variables, such as the coefficient of pressure \(C_p\), skin friction coefficient \(C_f\), and non-dimensional velocity \(u^*\):
\begin{equation*}
    C_p = \frac{p-p_\infty}{\frac{1}{2}\rho_\infty U_\infty^2}, \quad C_f = \frac{\vec{\tau}_w}{\frac{1}{2}\rho_\infty U_\infty^2}, \quad u^* = \frac{\vec{u}}{U_\infty}
\end{equation*}
where \(\rho_\infty\) is the freestream density, \(p_\infty\) is the freestream pressure, and \(U_\infty\) is the freestream velocity magnitude. These quantities greatly help to align different ranges across datasets. For example, AhmedML has a maximum velocity magnitude \(|\mathbf{u}|\) of around 1m/s, while Emmi Wing has a maximum velocity magnitude \(|\mathbf{u}|\) of around 300 m/s. After non-dimensionalization, the velocity magnitude across all datasets collapses to an approximate range \(|\mathbf{u}^*| \in [0, 1.5]\), which can be seen in Figure \ref{fig:all_vol}. Similarly, the raw volumetric pressure can vary widely across datasets, but after non-dimensionalization, the coefficient of pressure collapses to an approximate range \(C_p \in [-1, 1]\) (can be much more negative). These ranges are sometimes even enforced by conservation laws; for example, the coefficient of pressure is bounded in incompressible flows (\(C_p < 1\)); values above this limit would require negative velocity magnitudes \(|\mathbf{u}| < 0\). 

We therefore perform an additional set of studies where we train models using global normalization statistics, in which each variable is scaled by the same constant across all datasets. Different constants for each variable are still used to bring different quantities into a
comparable range; the skin friction coefficient $C_f$, for example, is typically
orders of magnitude smaller than the pressure coefficient or velocity. After
non-dimensionalization, the velocity is additionally expressed as a deficit relative to
the freestream (\(u' = (\vec{u} - \vec{U}_\infty)/U_\infty\)) 
and then each output field is divided by its scaling factor: $f_{C_p} = 0.33$,
$f_{C_f} = 0.002$, and $f_{u'} = 0.2$. For vector quantities, the same factor is
applied to every component. These factors were computed on the pretraining set, excluding the finetuning data, so that the resulting non-dimensional fields have approximately unit scale. 

\begin{table}[t]
    \centering
    \small 
    \begin{tabular}{@{}lll@{}}
        \toprule
        Quantity & Dataset-specific & Global \\
        \midrule
        Pressure coefficient
            & \((C_p - \mu_d^{(C_p)}) / \sigma_d^{(C_p)}\)
            & \(C_p / f_{C_p}\), \(f_{C_p} = 0.33\) \\
        Skin friction coefficient
            & \((C_f - \mu_d^{(C_f)}) / \sigma_d^{(C_f)}\)
            & \(C_f / f_{C_f}\), \(f_{C_f} = 0.002\) \\
        Velocity
            & \((u^* - \mu_d^{(u^*)}) / \sigma_d^{(u^*)}\), \(u^* = \vec{u} / U_\infty\)
            & \(u' / f_{u'}\), \(u' = (\vec{u} - \vec{U}_\infty) / U_\infty\), \(f_{u'} = 0.2\) \\
        Simulation parameters
            & \((\theta - \theta_{\min}) / (\theta_{\max} - \theta_{\min})\) 
            & \((\theta - \theta_{\min}) / (\theta_{\max} - \theta_{\min})\) \\
        Geometry and queries
            & \((G - \bar{c}_G) / s_d\), \((x - \bar{c}_G) / s_d\)
            & \((G - \bar{c}_G) / s_d\), \((x - \bar{c}_G) / s_d\) \\
        \midrule
        Estimated from simulation
            & \(\mu_d, \sigma_d\) for every field
            & None \\
        \bottomrule
    \end{tabular}
    \caption{Comparison of dataset-specific and global normalization. The subscript \(d\) denotes a quantity computed separately for each dataset.}
    \label{tab:norm_comparison}
\end{table}

\begin{table}[h!]
    \centering
    \small
    \addtolength{\tabcolsep}{-0.375em}
    \begin{tabular}{@{}l?cccc?cccc?cccc?cccc@{}}
        \toprule
        %& \multicolumn{16}{c}{Held-out target dataset} \\
        %\cmidrule(lr){2-17}
        \rlap{Held-out Datasets:} & \multicolumn{4}{c}{AhmedML}
        & \multicolumn{4}{c}{SHIFT-Sub}
        & \multicolumn{4}{c}{SHIFT-CCA}
        & \multicolumn{4}{c}{HiLiftAeroML} \\
        \cmidrule(lr){2-5}
        \cmidrule(lr){6-9}
        \cmidrule(lr){10-13}
        \cmidrule(lr){14-17}
        \rlap{\(\#\) FT Samples:} 
        & 0 & 2 & 4 & 8
        & 0 & 2 & 4 & 8
        & 0 & 2 & 4 & 8
        & 0 & 2 & 4 & 8 \\
        \midrule

        \multicolumn{17}{@{}l}{\textit{No Pretraining}} \\[-0.15em]
        Random Init.  & 0.82 & 0.42 & 0.40 & 0.33 & 0.82 & 0.38 & 0.31 & 0.30 & 0.85 & 0.22 & 0.21 & 0.21 & 0.90 & 0.65 & 0.65 & 0.64 \\

        \addlinespace[0.25em]
        \multicolumn{17}{@{}l}{
            \textit{Domain Experts}
        } \\[-0.15em]

        DrivAerNet++    & 0.53 & \best{0.15} & 0.13 & 0.12 & 0.89 & 0.19 & 0.12 & 0.09 & 0.89 & 0.13 & 0.11 & 0.10 & \worst{0.94} & 0.55 & 0.47 & 0.46 \\
        Emmi-Wing     & 0.71 & 0.20 & 0.20 & 0.19 & \best{0.67} & \worst{0.24} & 0.17 & \worst{0.15} & 0.68 & 0.14 & 0.12 & 0.12 & \best{0.82} & 0.54 & 0.46 & 0.46 \\
        SuperWing     & 0.76 & \worst{0.23} & \worst{0.24} & \worst{0.19} & 0.72 & 0.22 & 0.16 & 0.13 & \best{0.65} & 0.12 & 0.11 & 0.11 & 0.83 & \best{\textbf{0.46}} & 0.37 & 0.38 \\
        DrivAerML     & \best{0.51} & 0.18 & 0.16 & 0.14 & 0.83 & \best{0.17} & \best{\textbf{0.10}} & \best{\textbf{0.08}} & 0.88 & 0.13 & \best{\textbf{0.10}} & \best{0.10} & 0.94 & 0.51 & 0.44 & 0.45 \\
        WindsorML     & 0.57 & 0.18 & \best{0.13} & \best{0.11} & \worst{1.43} & 0.23 & 0.14 & 0.12 & \worst{1.13} & \worst{0.15} & \worst{0.13} & \worst{0.13} & 0.88 & \worst{0.57} & \worst{0.52} & \worst{0.51} \\
        Double-Delta  & \worst{0.82} & 0.20 & 0.20 & 0.17 & 0.83 & 0.22 & \worst{0.18} & 0.13 & 0.68 & \best{0.11} & 0.11 & 0.10 & 0.86 & 0.49 & \best{0.37} & \best{0.36} \\

        \addlinespace[0.25em]
        \multicolumn{17}{@{}l}{\textit{Cross-Domain}} \\[-0.15em]

        Joint Model       & \textbf{0.50} & \textbf{0.13} & \textbf{0.11} & \textbf{0.09} & \textbf{0.64} & \textbf{0.14} & \textbf{0.10} & 0.09 & \textbf{0.55} & \textbf{0.10} & \textbf{0.10} & \textbf{0.09} & \textbf{0.75} & \textbf{0.46} & \textbf{0.36} & \textbf{0.34} \\
        \bottomrule
    \end{tabular}

    \caption{Validation error for \textit{SMART (with global normalization)} on four held-out target datasets.
    Rows indicate the model initialization; single-domain experts are
    identified by their pretraining dataset. Columns give the number of
    fine-tuning samples, with 0 denoting zero-shot evaluation. Green and
    red indicate the best and worst single-domain experts, respectively,
    and bold indicates the lowest overall error.}
    \label{tab:smart_global_generalization}
\end{table}

\begin{table}[t]
    \centering
    \small 
    \addtolength{\tabcolsep}{-0.5em}
    \begin{tabular}{llccccccc}
        \toprule
          Model      &    Normalization      & DrivAerNet & DrivAerML & Emmi-Wing & WindsorML & SuperWing & Double-Delta & Mean \\
        \midrule
        Expert  & Dataset-specific  &      0.142 &     0.047 &     0.030 &     0.067 &     0.048 &        0.057 & 0.065 \\
                & Global &      0.142 &     0.046 &     0.030 &     0.067 &     0.047 &        0.058 & 0.065 \\
        \midrule
        Joint   & Dataset-specific   &      0.173 &     0.134 &     0.043 &     0.074 &     0.082 &        0.090 & 0.099 \\
                & Global &      0.172 &     0.132 &     0.041 &     0.074 &     0.079 &        0.091 & 0.098 \\
        \bottomrule
        \end{tabular}
    \caption{Comparison of dataset-specific and global normalization during pretraining. Losses are computed in the de-normalized space (raw \(C_p, C_f, u^*\)). Note that each cell in the Expert block is a different model, while each row in the Joint block is a single model evaluated across all datasets.}
    \label{tab:pretraining_norm}
\end{table}

Normalization for simulation parameters \(\theta\), geometries \(G\), and queries \(x\) are left unchanged. Simulation parameters are already min--max scaled using fixed bounds shared across all datasets, \(\theta_{\min} = [0, -20^\circ]\) and \(\theta_{\max} = [1, 20^\circ]\), corresponding to \(\mathrm{Ma} \in [0, 1]\) and \(\alpha \in [-20^\circ, 20^\circ]\). This was found to be important such that the same normalized value corresponds to the same operating condition across all datasets. We choose dataset-specific statistics for the geometry because the centroid \(\bar{c}_G\) and characteristic bounding box or length scale are design parameters and therefore are known at deployment. Each input geometry is centered at the origin and scaled by a dataset-specific factor \(s_d\). Table~\ref{tab:norm_comparison} provides a summary for how normalization is typically done for output fields (means/variances computed over a dataset) and a global normalization.

\paragraph{Results.} We repeat the main experiments of Section~\ref{sec:generalization} for expert and joint models trained with global normalization. Zero- and few-shot generalization results are reported in Table~\ref{tab:smart_global_generalization}, and pretraining results are in Table~\ref{tab:pretraining_norm}. The overall trend still holds, with the joint model matching or outperforming all baselines
in 15 of the 16 settings. Few-shot errors of the pretrained models change little compared
to dataset-specific normalization (Table~\ref{tab:generalization}), and although the joint
model's absolute zero-shot error rises on three of the four held-out datasets, it now
achieves the lowest zero-shot error on all four.

During pretraining the two normalization schemes are also very similar. For expert models, the choice of normalization has no measurable effect, which is expected. Each expert sees only a single dataset, so any constant shift or scale is readily absorbed during training. For the joint model, the effect is likewise negligible during pretraining. Global normalization may even be advantageous for the output fields, since it preserves the physical meaning of each variable across datasets. For example, \(C_p = 1\) consistently corresponds to a stagnation point, and \(C_f = 0\) (vanishing wall-shear stress) consistently indicates flow separation, regardless of the dataset. Overall, these results show that global normalization can perform well, but the choice of normalization is still important and deserves more study. Given the large number of mesh points per simulation, it may still be possible to get sufficient estimates of means/variances from a few finetuning samples. 

\subsection{Discretization}
\label{app:discretization}

\paragraph{Overview.} In our main experiments, we sample query points approximately uniformly in space rather than drawing them at random from the simulation mesh. This is a design choice we made to keep the discretization more consistent across domains. Furthermore, uniform spatial sampling also better reflects deployment conditions: meshing is expensive, predominantly CPU-based, and still relies on human expertise to refine, so a simulation mesh is generally unavailable at inference time. However, this is generally not an issue, as prior studies have already shown that sampling the queries differently during inference does not significantly affect errors due to the discretization invariance of AB-UPT and SMART. For example, training on the simulation mesh, then querying the model on the CAD mesh has already been shown to work well \citep{alkin2025abuptscalingneuralcfd,alkin2025abuptautomotiveaerospaceapplications,hagnberger2026smartscalablemeshfreeaerodynamic}. Nevertheless, we also perform a similar study by querying models trained on the more spatially uniform mesh on the simulation mesh. In some cases, querying on the simulation mesh may better resolve steep gradients, for example in the boundary layer or in regions of adaptive mesh refinement. However, as noted above, this mesh is usually not available before the simulation has been run.

\begin{table}[t]
    \centering
    \small
    \addtolength{\tabcolsep}{-0.375em}
    \begin{tabular}{@{}l?cccc?cccc?cccc?cccc@{}}
        \toprule
        %& \multicolumn{16}{c}{Held-out target dataset} \\
        %\cmidrule(lr){2-17}
        \rlap{Held-out Datasets:} & \multicolumn{4}{c}{AhmedML}
        & \multicolumn{4}{c}{SHIFT-Sub}
        & \multicolumn{4}{c}{SHIFT-CCA}
        & \multicolumn{4}{c}{HiLiftAeroML} \\
        \cmidrule(lr){2-5}
        \cmidrule(lr){6-9}
        \cmidrule(lr){10-13}
        \cmidrule(lr){14-17}
        \rlap{\(\#\) FT Samples:} 
        & 0 & 2 & 4 & 8
        & 0 & 2 & 4 & 8
        & 0 & 2 & 4 & 8
        & 0 & 2 & 4 & 8 \\
        \midrule

        \multicolumn{17}{@{}l}{\textit{No Pretraining}} \\[-0.15em]

       Random Init.  & 0.77 & 0.57 & 0.54 & 0.40 & 0.83 & 0.53 & 0.49 & 0.47 & 0.87 & 0.53 & 0.45 & 0.46 & 0.87 & 0.76 & 0.75 & 0.76 \\

        \addlinespace[0.25em]
        \multicolumn{17}{@{}l}{
            \textit{Domain Experts}
        } \\[-0.15em]
        DrivAerNet++    & 0.57 & \best{0.17} & \best{0.14} & \best{0.12} & \best{\textbf{0.64}} & 0.31 & 0.23 & 0.19 & 0.88 & 0.34 & 0.26 & 0.24 & 0.97 & 0.65 & 0.57 & 0.49 \\
        Emmi-Wing     & 0.81 & 0.22 & 0.21 & 0.19 & 0.77 & \worst{0.38} & 0.31 & \worst{0.27} & \best{0.77} & 0.28 & 0.26 & 0.26 & 0.98 & 0.62 & 0.54 & \worst{0.57} \\
        SuperWing     & \worst{0.86} & \worst{0.24} & \worst{0.22} & \worst{0.19} & 0.80 & 0.38 & \worst{0.34} & 0.27 & \worst{0.93} & 0.26 & 0.25 & 0.24 & \worst{1.11} & 0.60 & 0.49 & 0.47 \\
        DrivAerML     & 0.55 & 0.17 & 0.18 & 0.14 & 0.70 & \best{0.29} & \best{0.22} & \best{\textbf{0.18}} & 0.84 & 0.32 & 0.24 & 0.23 & 0.93 & 0.70 & 0.56 & 0.46 \\
        WindsorML     & \best{0.45} & 0.20 & 0.14 & 0.12 & 0.69 & 0.35 & 0.29 & 0.24 & 0.87 & \worst{0.37} & \worst{0.29} & \worst{0.28} & 0.92 & \worst{0.71} & \worst{0.60} & 0.54 \\
        Double-Delta  & 0.79 & 0.19 & 0.17 & 0.17 & \worst{0.82} & 0.36 & 0.31 & 0.23 & 0.85 & \best{0.24} & \best{0.22} & \best{0.22} & \best{\textbf{0.89}} & \best{0.55} & \best{\textbf{0.40}} & \best{0.37} \\

        \addlinespace[0.25em]
        \multicolumn{17}{@{}l}{\textit{Cross-Domain}} \\[-0.15em]

        Joint         & \textbf{0.34} & \textbf{0.12} & \textbf{0.10} & \textbf{0.09} & 0.66 & \textbf{0.25} & \textbf{0.21} & \textbf{0.18} & \textbf{0.70} & \textbf{0.22} & \textbf{0.20} & \textbf{0.20} & \textbf{0.89} & \textbf{0.52} & \textbf{0.40} & \textbf{0.36} \\
        \bottomrule
    \end{tabular}

    \caption{Validation error for \textit{SMART (evaluated on simulation mesh)} on four held-out target datasets.
    Rows indicate the model initialization; single-domain experts are
    identified by their pretraining dataset. Columns give the number of
    fine-tuning samples, with 0 denoting zero-shot evaluation. Green and
    red indicate the best and worst single-domain experts, respectively,
    and bold indicates the lowest overall error.}
    \label{tab:smart_mesh_generalization}
\end{table}

\paragraph{Results.} 
We replicate the main results in Section \ref{sec:generalization}, but instead query the finetuned models on the simulation mesh. Zero- and few-shot errors for the experts and joint models are given in Table \ref{tab:smart_mesh_generalization}. We can see that the error for all models tends to rise; this is because spatially uniform samples tend to resolve more of the freestream, which is easier to model. However, we can see that the overall trend is still consistent, where the joint model has better zero- and few-shot performance, which validates that cross-domain pretraining is still helpful, regardless of the discretization. It also serves to corroborate prior studies that show discretization invariance for SMART/AB-UPT. 

\subsection{Scaling Model Size or Compute}
\label{app:scaling}
We present additional results on the effects of scaling model size and pretraining compute (number of gradient steps), both on pretraining (Table~\ref{tab:pretraining_scale}) and on generalization to unseen datasets (Table~\ref{tab:scaling_gen}). As expected, the pretraining loss decreases consistently with both larger models and longer training. However, these gains do not generally carry over to unseen datasets: beyond a certain point, further increasing either model size or the number of gradient steps can raise the zero-shot error, suggesting that the model begins to overfit to the pretraining distribution. Few-shot performance likewise does not consistently improve with scale, although larger models tend to be slightly more sample-efficient. These results suggest that generating more CFD data is the most promising way to improve generalization, followed by increasing model size and then training compute.

\begin{table}[t]
    \centering
    \small 
    \addtolength{\tabcolsep}{-0.15em}
        \begin{tabular}{p{2.25cm}cccccc}
        \toprule
        \textit{Model} & DrivAerNet & Emmi-Wing & WindsorML & Double-Delta & DrivAerML & SuperWing \\
        \midrule
        Joint (Base) & 0.174 & 0.043 &  0.075 &  0.091 &  0.134 & 0.082 \\
         Joint (121M) &  \textbf{0.150} &  0.039 & 0.072 &  0.070 & 0.074 &  0.074 \\
         Joint (6xSteps) &  \textbf{0.150} & \textbf{0.029} &  \textbf{0.069} & \textbf{0.063} & \textbf{0.073} & \textbf{0.041} \\
        \bottomrule
        \end{tabular}
    \caption{Validation error of the joint model across the pretraining datasets, either with an increased model size (121M) or increased gradient budget (6xSteps).
    }
    \label{tab:pretraining_scale}
\end{table}

\begin{table}[t]
    \centering
    \small
    \addtolength{\tabcolsep}{-0.375em}
    \begin{tabular}{@{}l?cccc?cccc?cccc?cccc@{}}
        \toprule
        \rlap{Held-out Datasets:} & \multicolumn{4}{c}{AhmedML}
        & \multicolumn{4}{c}{SHIFT-Sub}
        & \multicolumn{4}{c}{SHIFT-CCA}
        & \multicolumn{4}{c}{HiLiftAeroML} \\
        \cmidrule(lr){2-5}
        \cmidrule(lr){6-9}
        \cmidrule(lr){10-13}
        \cmidrule(lr){14-17}
        \rlap{\(\#\) FT Samples:} 
        & 0 & 2 & 4 & 8
        & 0 & 2 & 4 & 8
        & 0 & 2 & 4 & 8
        & 0 & 2 & 4 & 8 \\
        \midrule

        Joint (Base)
        & \textbf{0.31} & \textbf{0.12} & 0.10 & 0.09
        & \textbf{0.52} & \textbf{0.14} & \textbf{0.10} & 0.09
        & \textbf{0.52} & 0.11 & 0.10 & 0.10
        & \textbf{0.90} & 0.46 & 0.34 & 0.33 \\
        Joint (121M)
        & \textbf{0.31} & \textbf{0.12} & \textbf{0.09} & \textbf{0.08}
        & 0.53 & 0.16 & 0.11 & \textbf{0.08}
        & 0.55 & \textbf{0.09} & \textbf{0.08} & \textbf{0.08}
        & 0.92 & 0.45 & \textbf{0.32} & \textbf{0.28} \\
        Joint (6xSteps)
        & 0.32 & \textbf{0.12} & 0.10 & \textbf{0.08}
        & 0.61 & 0.16 & \textbf{0.10} & 0.09
        & 0.55 & 0.10 & 0.09 & 0.09
        & 1.11 & \textbf{0.44} & \textbf{0.32} & 0.30 \\
        \bottomrule
    \end{tabular}

    \caption{Validation error on four held-out target datasets. Each row is a different joint model, either with increased model size (121M) or increased gradient budget (6xSteps). Columns give the number of
    fine-tuning samples, with 0 denoting zero-shot evaluation. }
    \label{tab:scaling_gen}
\end{table}

\begin{table}[t]
    \centering
    \small 
    \addtolength{\tabcolsep}{-0.15em}
        \begin{tabular}{p{2.25cm}cccccc}
        \toprule
        \textit{Model} & DrivAerNet & Emmi-Wing & WindsorML & Double-Delta & DrivAerML & SuperWing \\
        \midrule
        Joint (Base) & 0.174 & 0.043 &  0.075 &  0.091 &  0.134 & 0.082 \\
         Joint (IC)  & 0.176 & 0.075 &  0.076 &  0.098 &  0.144 & 0.098 \\
        \bottomrule
        \end{tabular}
    \caption{Pretraining validation error of the joint base model (SMART) and the in-context model (SMART-IC) on each dataset. SMART-IC additionally receives a solved context simulation from the same dataset as the query. Providing context does not reduce the error on any dataset.}
    \label{tab:pretraining_IC}
\end{table}

\subsection{In-Context Pretraining}
\label{app:cross}
In time-dependent PDE settings, the predominant approach for cross-domain models is to infer the underlying dynamics from a set of paired context frames. To test whether a similar mechanism is needed for steady-state problems, we train a joint in-context model across domains (SMART-IC). Alongside the standard inputs (the query geometry $G_q$, query points $x_q$, and query parameters $\theta_q$), SMART-IC receives a solved context simulation drawn from the same dataset as the query. This context consists of the field values $f_c(x_c)$ at a set of context points $x_c$ and the corresponding simulation parameters $\theta_c$, so the model's prediction is $f_\phi(x_q; G_q, \theta_q, f_c(x_c), \theta_c)$. We do not pass the context geometry $G_c$ explicitly, since it is implicitly encoded by the surface points on which the context field $f_c(x_c)$ is defined.

Table \ref{tab:pretraining_IC} compares the pretraining performance of SMART-IC with that of the base model, SMART. Providing a context simulation yields no benefit, and in most cases the error is slightly higher. This is expected for steady-state problems: the forward problem is well-posed given only the query geometry $G_q$ and parameters $\theta_q$, so the output $f_q(x_q)$ is fully determined by these inputs. Additional context would help only if unobserved factors, such as differences in closure model or solver, varied across similar model inputs and left the problem ill-defined. This does not seem to occur in current, steady-state CFD datasets. The extra input therefore carries no useful information and, at worst, adds variance that increases the error. Consistent with this, we found that SMART-IC ignores the context entirely: zeroing out the context at inference time leaves its predictions unchanged.

As a whole, these results suggest that cross-domain CFD surrogates may require different strategies from those developed for time-dependent PDEs. In temporal problems, context is useful because domains differ mainly in their dynamics (for example, in the PDE terms or coefficients), which the model can infer from observed context frames. In steady-state CFD, by contrast, every domain is governed by the same Navier–Stokes equations, and the variation across domains lies instead in the geometry and boundary conditions. Future cross-domain methods may therefore benefit more from better representations of geometric and boundary-condition variability than from in-context mechanisms.

\subsection{Variable-Specific Errors}
\label{sec:var_specifics}

For the main finetuning experiment in Section \ref{sec:generalization}, we also report per-variable errors in place of the averaged error. Tables \ref{tab:generalization_surface_cp}, \ref{tab:generalization_surface_cf}, \ref{tab:generalization_volume_p}, and \ref{tab:generalization_volume_vel} give the NRMSE for surface $C_p$, surface $C_f$, volume $C_p$, and volume $\mathbf{u}^*$, respectively. Surface fields are generally harder to predict than volume fields, since much of the volume lies in the freestream, which is generally smooth. Nonetheless, the main trends hold across all variables: the joint model generally outperforms both the domain experts and the models trained from scratch.

\begin{table}[t]
    \centering
    \small
    \addtolength{\tabcolsep}{-0.375em}
    \begin{tabular}{@{}l?cccc?cccc?cccc?cccc@{}}
        \toprule
        \rlap{Held-out Datasets:} & \multicolumn{4}{c}{AhmedML}
        & \multicolumn{4}{c}{SHIFT-Sub}
        & \multicolumn{4}{c}{SHIFT-CCA}
        & \multicolumn{4}{c}{HiLiftAeroML} \\
        \cmidrule(lr){2-5}
        \cmidrule(lr){6-9}
        \cmidrule(lr){10-13}
        \cmidrule(lr){14-17}
        \rlap{\(\#\) FT Samples:}
        & 0 & 2 & 4 & 8
        & 0 & 2 & 4 & 8
        & 0 & 2 & 4 & 8
        & 0 & 2 & 4 & 8 \\
        \midrule

        \multicolumn{17}{@{}l}{\textit{No Pretraining}} \\[-0.15em]
        Random init.
        & 0.82 & 0.74 & 0.65 & 0.25
        & 0.88 & 0.53 & 0.39 & 0.37
        & 0.97 & 0.18 & 0.16 & 0.16
        & 0.96 & 0.68 & 0.69 & 0.68 \\

        \addlinespace[0.25em]
        \multicolumn{17}{@{}l}{\textit{Domain Experts}} \\[-0.15em]

        DrivAerNet++
        & 0.65 & 0.21 & 0.17 & 0.14
        & 0.87 & 0.36 & 0.20 & 0.13
        & 0.96 & 0.13 & 0.08 & 0.07
        & 1.08 & 0.59 & 0.53 & 0.52 \\

        Emmi-Wing
        & 0.85 & 0.24 & 0.22 & \worst{0.20}
        & 0.90 & \worst{0.40} & 0.30 & \worst{0.22}
        & \best{0.89} & 0.12 & \worst{0.10} & \worst{0.10}
        & 1.12 & 0.58 & 0.51 & 0.52 \\

        SuperWing
        & 0.82 & \worst{0.28} & \worst{0.25} & 0.20
        & 0.89 & 0.37 & 0.24 & 0.19
        & 0.97 & 0.11 & 0.10 & 0.09
        & \worst{1.12} & 0.55 & 0.48 & 0.47 \\

        DrivAerML
        & 0.61 & 0.21 & 0.19 & 0.17
        & 0.87 & 0.33 & \best{0.18} & \best{0.12}
        & 0.92 & 0.11 & \best{\textbf{0.07}} & \best{\textbf{0.06}}
        & 1.02 & 0.61 & 0.52 & 0.52 \\

        WindsorML
        & \best{0.49} & \best{0.21} & \best{0.15} & \best{0.12}
        & \best{\textbf{0.73}} & \best{0.32} & 0.22 & 0.15
        & \worst{1.02} & \worst{0.15} & 0.09 & 0.09
        & 1.04 & \worst{0.62} & \worst{0.57} & \worst{0.56} \\

        Double-Delta
        & \worst{0.86} & 0.22 & 0.21 & 0.18
        & \worst{0.92} & 0.39 & \worst{0.32} & 0.18
        & 1.02 & \best{\textbf{0.08}} & \textbf{0.07} & 0.07
        & \best{\textbf{0.87}} & \best{0.52} & \best{0.40} & \best{0.37} \\

        \addlinespace[0.25em]
        \multicolumn{17}{@{}l}{\textit{Cross-Domain}} \\[-0.15em]

        Joint Model
        & \textbf{0.37} & \textbf{0.14} & \textbf{0.11} & \textbf{0.09}
        & 0.78 & \textbf{0.22} & \textbf{0.14} & \textbf{0.11}
        & \textbf{0.76} & \textbf{0.08} & \textbf{0.07} & 0.07
        & \textbf{0.87} & \textbf{0.50} & \textbf{0.37} & \textbf{0.33} \\
        \bottomrule
    \end{tabular}

    \caption{Validation error in \textit{surface pressure coefficient} \(C_p\) on four held-out target datasets.
    Rows indicate the model initialization; single-domain experts are
    identified by their pretraining dataset. Columns give the number of
    fine-tuning samples, with 0 denoting zero-shot evaluation. Green and
    red indicate the best and worst single-domain experts, respectively,
    and bold indicates the lowest overall error.}
    \label{tab:generalization_surface_cp}
\end{table}

\begin{table}[t]
    \centering
    \small
    \addtolength{\tabcolsep}{-0.375em}
    \begin{tabular}{@{}l?cccc?cccc?cccc?cccc@{}}
        \toprule
        \rlap{Held-out Datasets:} & \multicolumn{4}{c}{AhmedML}
        & \multicolumn{4}{c}{SHIFT-Sub}
        & \multicolumn{4}{c}{SHIFT-CCA}
        & \multicolumn{4}{c}{HiLiftAeroML} \\
        \cmidrule(lr){2-5}
        \cmidrule(lr){6-9}
        \cmidrule(lr){10-13}
        \cmidrule(lr){14-17}
        \rlap{\(\#\) FT Samples:}
        & 0 & 2 & 4 & 8
        & 0 & 2 & 4 & 8
        & 0 & 2 & 4 & 8
        & 0 & 2 & 4 & 8 \\
        \midrule

        \multicolumn{17}{@{}l}{\textit{No Pretraining}} \\[-0.15em]
        Random init.
        & 0.65 & 0.61 & 0.60 & 0.49
        & 0.41 & 0.35 & 0.34 & 0.32
        & 0.41 & 0.30 & 0.31 & 0.31
        & 0.77 & 0.73 & 0.74 & 0.73 \\

        \addlinespace[0.25em]
        \multicolumn{17}{@{}l}{\textit{Domain Experts}} \\[-0.15em]

        DrivAerNet++
        & 0.64 & \best{0.17} & 0.14 & 0.12
        & 0.45 & 0.19 & 0.13 & 0.12
        & 0.44 & 0.20 & 0.18 & 0.18
        & 0.86 & 0.62 & 0.56 & 0.55 \\

        Emmi-Wing
        & \worst{0.75} & 0.19 & 0.19 & 0.18
        & \worst{0.49} & \worst{0.19} & 0.16 & \worst{0.15}
        & 0.39 & 0.19 & 0.18 & 0.18
        & \worst{1.39} & 0.63 & 0.59 & \worst{0.61} \\

        SuperWing
        & 0.72 & \worst{0.23} & \worst{0.21} & \worst{0.18}
        & 0.46 & 0.19 & 0.15 & 0.15
        & \best{0.39} & 0.19 & 0.18 & 0.18
        & 1.13 & 0.60 & 0.58 & 0.58 \\

        DrivAerML
        & 0.56 & 0.17 & 0.15 & 0.14
        & 0.42 & \best{0.17} & \best{\textbf{0.12}} & \best{\textbf{0.11}}
        & \worst{0.46} & 0.19 & \best{\textbf{0.17}} & \best{\textbf{0.17}}
        & 0.86 & 0.64 & 0.55 & 0.55 \\

        WindsorML
        & \best{0.41} & 0.17 & \best{0.13} & \best{0.11}
        & \best{\textbf{0.37}} & 0.17 & 0.14 & 0.12
        & 0.39 & \worst{0.21} & \worst{0.19} & \worst{0.19}
        & \best{\textbf{0.81}} & \worst{0.66} & \worst{0.60} & 0.60 \\

        Double-Delta
        & 0.68 & 0.18 & 0.17 & 0.16
        & 0.41 & 0.19 & \worst{0.17} & 0.14
        & 0.45 & \best{0.18} & 0.18 & 0.18
        & 0.82 & \best{0.58} & \best{0.47} & \best{0.46} \\

        \addlinespace[0.25em]
        \multicolumn{17}{@{}l}{\textit{Cross-Domain}} \\[-0.15em]

        Joint Model
        & \textbf{0.34} & \textbf{0.11} & \textbf{0.10} & \textbf{0.08}
        & 0.38 & \textbf{0.14} & \textbf{0.12} & \textbf{0.11}
        & \textbf{0.30} & \textbf{0.17} & \textbf{0.17} & \textbf{0.17}
        & 0.84 & \textbf{0.56} & \textbf{0.44} & \textbf{0.42} \\
        \bottomrule
    \end{tabular}

    \caption{Validation error in \textit{surface skin-friction coefficient} \(C_f\) on four held-out target datasets.
    Rows indicate the model initialization; single-domain experts are
    identified by their pretraining dataset. Columns give the number of
    fine-tuning samples, with 0 denoting zero-shot evaluation. Green and
    red indicate the best and worst single-domain experts, respectively,
    and bold indicates the lowest overall error.}
    \label{tab:generalization_surface_cf}
\end{table}

\begin{table}[t]
    \centering
    \small
    \addtolength{\tabcolsep}{-0.375em}
    \begin{tabular}{@{}l?cccc?cccc?cccc?cccc@{}}
        \toprule
        \rlap{Held-out Datasets:} & \multicolumn{4}{c}{AhmedML}
        & \multicolumn{4}{c}{SHIFT-Sub}
        & \multicolumn{4}{c}{SHIFT-CCA}
        & \multicolumn{4}{c}{HiLiftAeroML} \\
        \cmidrule(lr){2-5}
        \cmidrule(lr){6-9}
        \cmidrule(lr){10-13}
        \cmidrule(lr){14-17}
        \rlap{\(\#\) FT Samples:}
        & 0 & 2 & 4 & 8
        & 0 & 2 & 4 & 8
        & 0 & 2 & 4 & 8
        & 0 & 2 & 4 & 8 \\
        \midrule

        \multicolumn{17}{@{}l}{\textit{No Pretraining}} \\[-0.15em]
        Random init.
        & 0.98 & 0.57 & 0.47 & 0.36
        & 1.00 & 0.42 & 0.33 & 0.36
        & 0.99 & 0.21 & 0.21 & 0.21
        & 0.98 & 0.62 & 0.63 & 0.63 \\

        \addlinespace[0.25em]
        \multicolumn{17}{@{}l}{\textit{Domain Experts}} \\[-0.15em]

        DrivAerNet++
        & 0.63 & 0.24 & 0.21 & 0.18
        & \best{0.85} & 0.24 & 0.13 & 0.09
        & 1.14 & 0.14 & 0.09 & 0.08
        & 1.48 & 0.53 & 0.46 & 0.46 \\

        Emmi-Wing
        & 1.01 & 0.28 & \worst{0.30} & \worst{0.27}
        & 1.01 & 0.29 & 0.21 & \worst{0.18}
        & \best{0.91} & 0.13 & 0.10 & 0.10
        & 1.13 & 0.51 & 0.46 & 0.47 \\

        SuperWing
        & \worst{1.04} & \worst{0.30} & 0.29 & 0.25
        & \worst{1.08} & 0.26 & 0.17 & 0.15
        & 0.96 & 0.11 & 0.10 & 0.10
        & 1.05 & 0.49 & 0.43 & 0.41 \\

        DrivAerML
        & 0.67 & 0.25 & 0.27 & 0.23
        & 0.85 & \best{0.21} & \best{0.12} & \best{0.08}
        & 1.03 & 0.14 & \best{0.08} & \best{\textbf{0.07}}
        & 1.13 & 0.53 & 0.45 & 0.45 \\

        WindsorML
        & \best{0.45} & \best{0.21} & \best{0.17} & \best{0.14}
        & 0.93 & 0.22 & 0.14 & 0.10
        & \worst{1.19} & \worst{0.17} & \worst{0.12} & \worst{0.11}
        & \worst{1.50} & \worst{0.54} & \worst{0.52} & \worst{0.49} \\

        Double-Delta
        & 1.00 & 0.27 & 0.27 & 0.23
        & 1.01 & \worst{0.29} & \worst{0.23} & 0.14
        & 1.00 & \best{0.09} & 0.08 & 0.08
        & \best{\textbf{1.02}} & \best{0.46} & \best{0.35} & \best{0.33} \\

        \addlinespace[0.25em]
        \multicolumn{17}{@{}l}{\textit{Cross-Domain}} \\[-0.15em]

        Joint Model
        & \textbf{0.36} & \textbf{0.18} & \textbf{0.14} & \textbf{0.13}
        & \textbf{0.73} & \textbf{0.15} & \textbf{0.09} & \textbf{0.07}
        & \textbf{0.84} & \textbf{0.08} & \textbf{0.07} & \textbf{0.07}
        & 1.09 & \textbf{0.43} & \textbf{0.32} & \textbf{0.29} \\
        \bottomrule
    \end{tabular}

    \caption{Validation error in \textit{volume pressure coefficient} \(C_p\) on four held-out target datasets.
    Rows indicate the model initialization; single-domain experts are
    identified by their pretraining dataset. Columns give the number of
    fine-tuning samples, with 0 denoting zero-shot evaluation. Green and
    red indicate the best and worst single-domain experts, respectively,
    and bold indicates the lowest overall error.}
    \label{tab:generalization_volume_p}
\end{table}

\begin{table}[t]
    \centering
    \small
    \addtolength{\tabcolsep}{-0.375em}
    \begin{tabular}{@{}l?cccc?cccc?cccc?cccc@{}}
        \toprule
        \rlap{Held-out Datasets:} & \multicolumn{4}{c}{AhmedML}
        & \multicolumn{4}{c}{SHIFT-Sub}
        & \multicolumn{4}{c}{SHIFT-CCA}
        & \multicolumn{4}{c}{HiLiftAeroML} \\
        \cmidrule(lr){2-5}
        \cmidrule(lr){6-9}
        \cmidrule(lr){10-13}
        \cmidrule(lr){14-17}
        \rlap{\(\#\) FT Samples:}
        & 0 & 2 & 4 & 8
        & 0 & 2 & 4 & 8
        & 0 & 2 & 4 & 8
        & 0 & 2 & 4 & 8 \\
        \midrule

        \multicolumn{17}{@{}l}{\textit{No Pretraining}} \\[-0.15em]
        Random init.
        & 0.26 & 0.26 & 0.25 & 0.25
        & 0.26 & 0.25 & 0.25 & 0.25
        & 0.35 & 0.24 & 0.25 & 0.25
        & 0.53 & 0.49 & 0.50 & 0.49 \\

        \addlinespace[0.25em]
        \multicolumn{17}{@{}l}{\textit{Domain Experts}} \\[-0.15em]

        DrivAerNet++
        & 0.19 & 0.09 & 0.07 & 0.06
        & 0.27 & \best{0.08} & \textbf{0.07} & \textbf{0.06}
        & 0.39 & 0.12 & 0.11 & 0.11
        & 0.97 & 0.38 & 0.33 & 0.33 \\

        Emmi-Wing
        & 0.29 & 0.12 & 0.13 & 0.12
        & \best{0.24} & 0.09 & 0.09 & 0.08
        & \best{0.25} & 0.12 & 0.11 & 0.12
        & 0.77 & 0.37 & 0.35 & \worst{0.36} \\

        SuperWing
        & \worst{0.70} & \worst{0.14} & \worst{0.13} & \worst{0.13}
        & \worst{0.69} & 0.09 & 0.09 & 0.09
        & \worst{0.65} & 0.11 & 0.11 & 0.11
        & \worst{1.74} & \best{0.37} & 0.34 & 0.33 \\

        DrivAerML
        & 0.21 & 0.09 & 0.10 & 0.09
        & 0.29 & 0.08 & \best{\textbf{0.07}} & \best{\textbf{0.06}}
        & 0.44 & 0.12 & 0.11 & 0.11
        & 0.95 & 0.40 & 0.33 & 0.32 \\

        WindsorML
        & \best{0.18} & \best{0.08} & \best{0.06} & \best{0.06}
        & 0.32 & \worst{0.13} & \worst{0.11} & \worst{0.09}
        & 0.41 & \worst{0.16} & \worst{0.15} & \worst{0.15}
        & 0.80 & \worst{0.44} & \worst{0.35} & 0.35 \\

        Double-Delta
        & 0.29 & 0.11 & 0.11 & 0.10
        & 0.26 & 0.08 & 0.08 & 0.07
        & 0.30 & \best{\textbf{0.09}} & \best{\textbf{0.09}} & \best{\textbf{0.09}}
        & \best{\textbf{0.73}} & 0.37 & \best{0.28} & \best{0.27} \\

        \addlinespace[0.25em]
        \multicolumn{17}{@{}l}{\textit{Cross-Domain}} \\[-0.15em]

        Joint Model
        & \textbf{0.15} & \textbf{0.07} & \textbf{0.05} & \textbf{0.05}
        & \textbf{0.18} & \textbf{0.07} & \textbf{0.07} & \textbf{0.06}
        & \textbf{0.20} & 0.10 & \textbf{0.09} & 0.10
        & 0.79 & \textbf{0.36} & \textbf{0.25} & \textbf{0.25} \\
        \bottomrule
    \end{tabular}

    \caption{Validation error in \textit{volume velocity} \(\mathbf{u}^*\) on four held-out target datasets.
    Rows indicate the model initialization; single-domain experts are
    identified by their pretraining dataset. Columns give the number of
    fine-tuning samples, with 0 denoting zero-shot evaluation. Green and
    red indicate the best and worst single-domain experts, respectively,
    and bold indicates the lowest overall error.}
    \label{tab:generalization_volume_vel}
\end{table}

\section{Additional Visualizations}
\label{app:viz}

For the main set of experiments in Section \ref{sec:generalization}, we plot model predictions after zero-shot and few-shot evaluation (\(S=2\)). These are 16 figures in total; for each finetuning dataset the surface $C_p$, surface $C_{f_x}$, volume $C_p$, and volume magnitude $|\mathbf{u}^*|$ is plotted. 

\begin{figure}[h!]
    \centering
    \includegraphics[width=\linewidth]{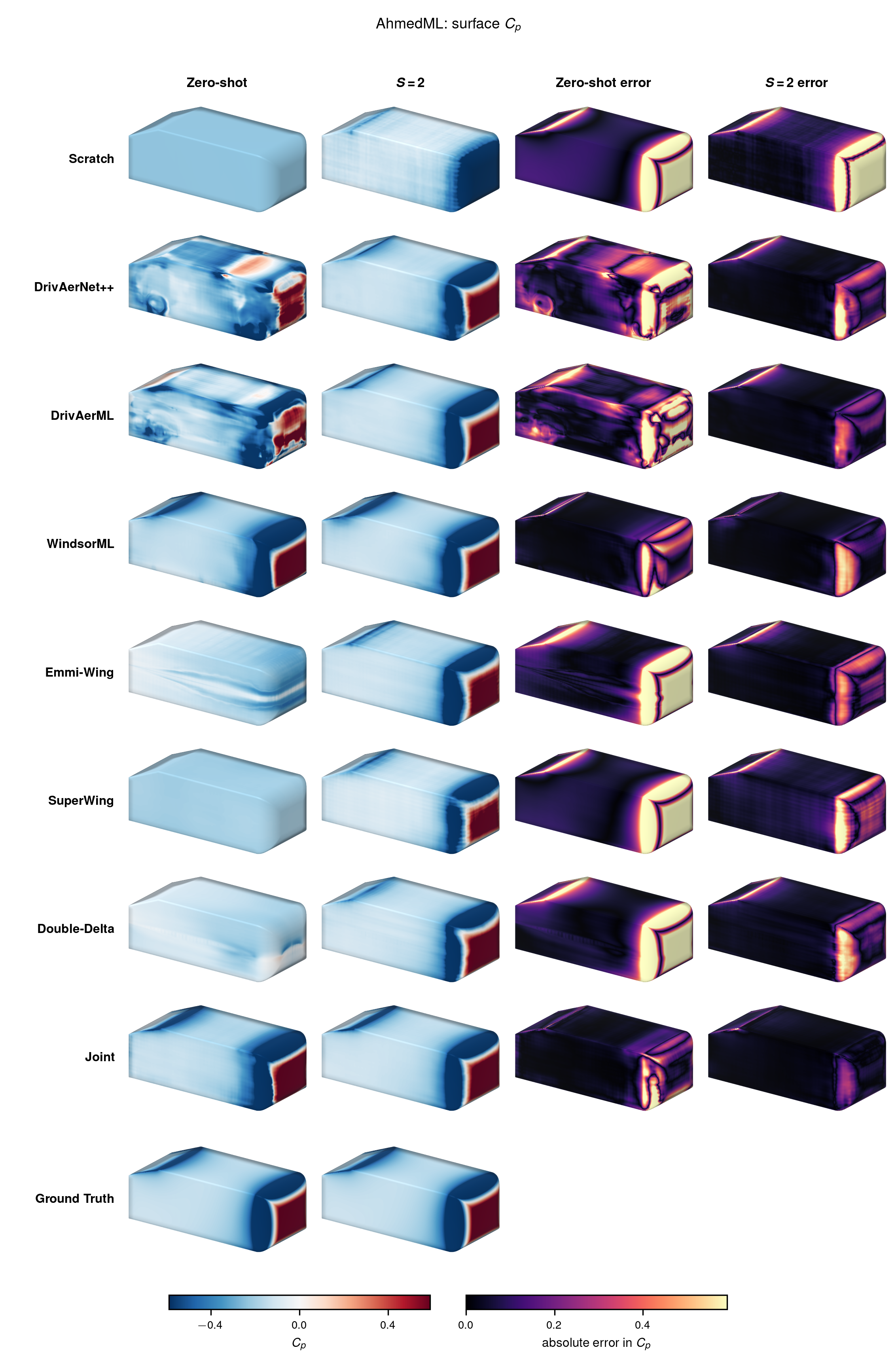}
    \caption{Model predictions of \textit{surface \(C_p\)} on AhmedML. }
    \label{fig:aml_surface_cp}
\end{figure}

\begin{figure}[h!]
    \centering
    \includegraphics[width=\linewidth]{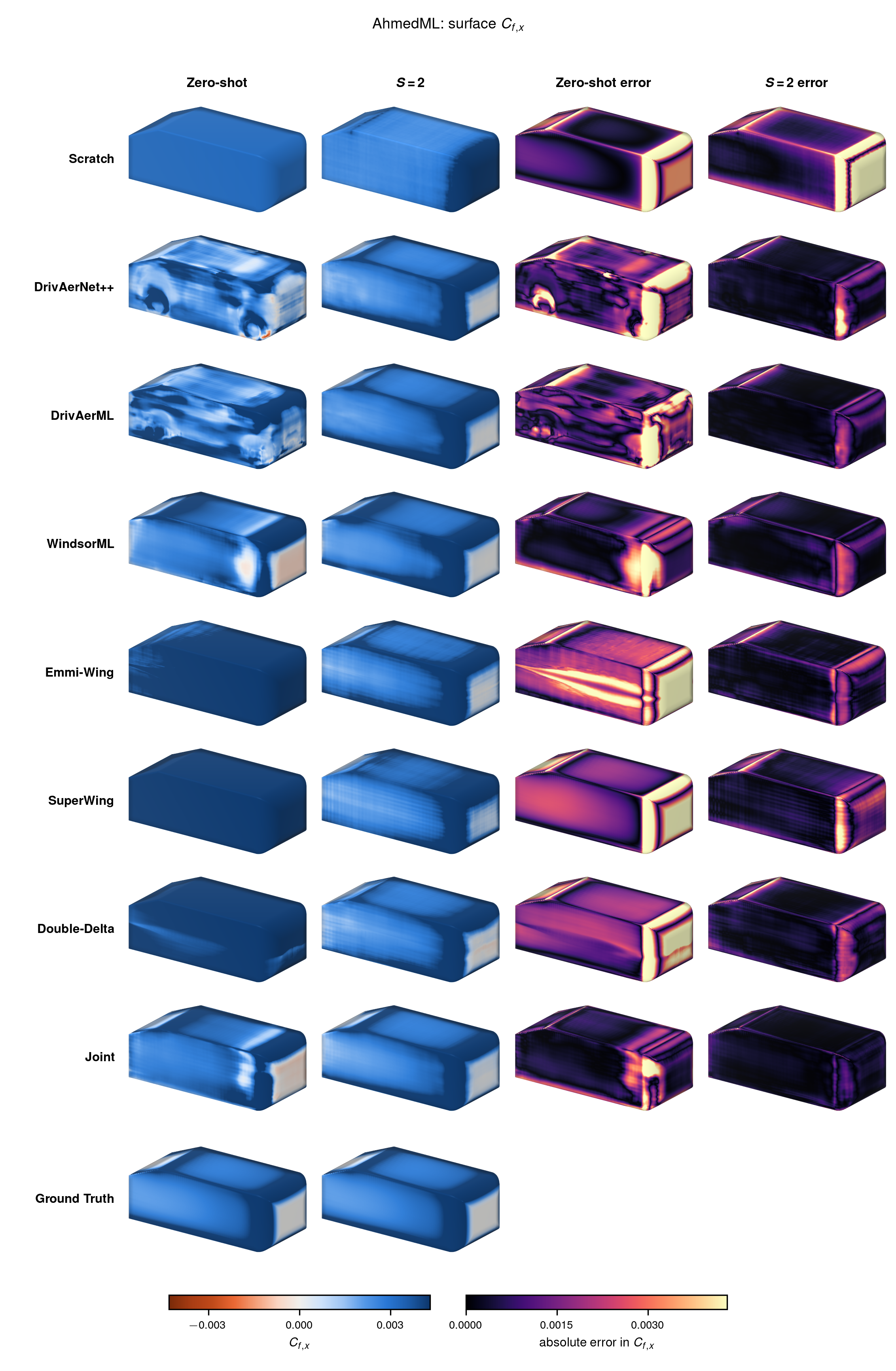}
    \caption{Model predictions of \textit{surface \(C_{f_x}\)} on AhmedML. }
    \label{fig:aml_surface_cf}
\end{figure}

\begin{figure}[h!]
    \centering
    \includegraphics[width=0.95\linewidth]{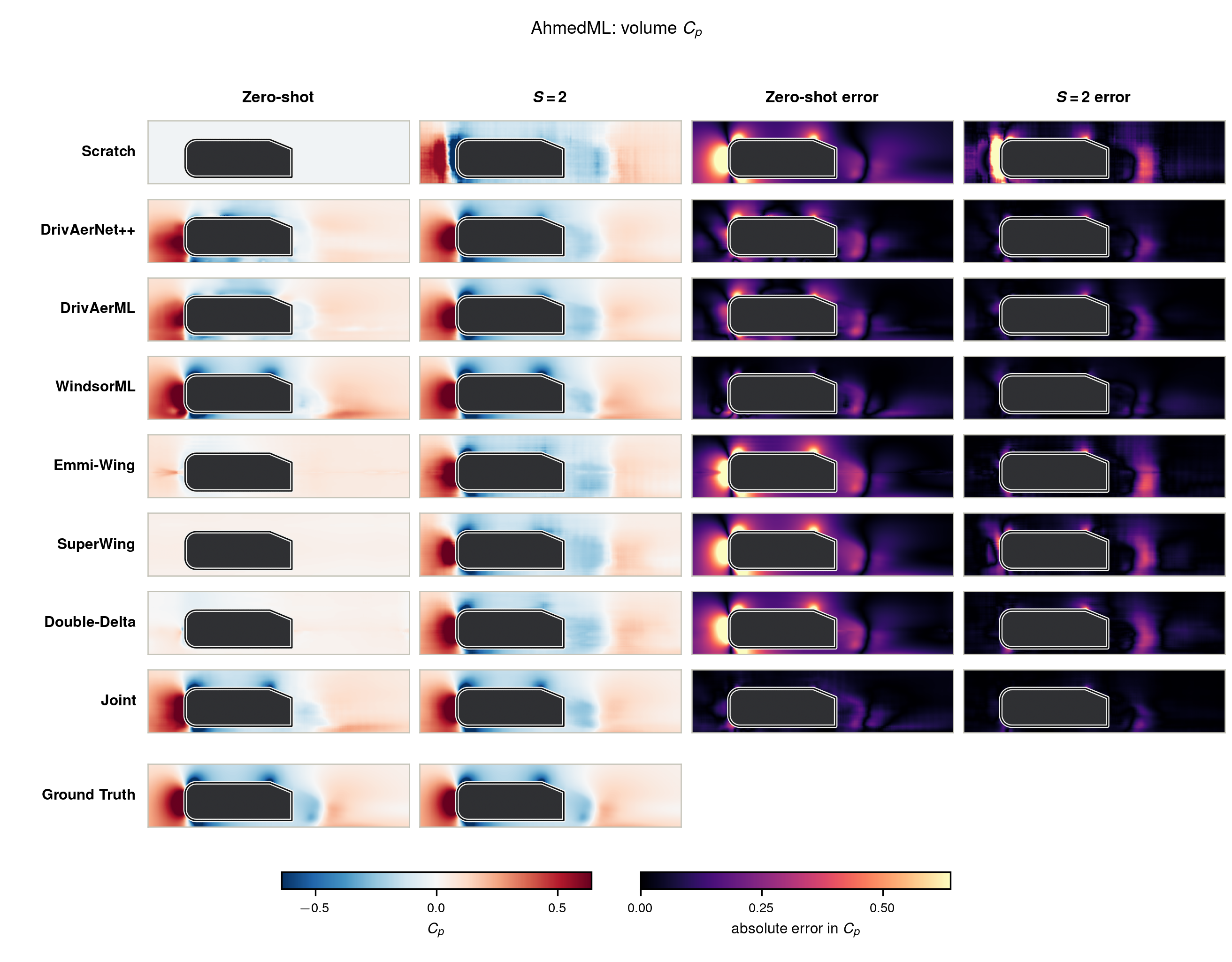}
    \caption{Model predictions of \textit{volume \(C_{p}\)} on AhmedML. }
\end{figure}

\begin{figure}[h!]
    \centering
    \includegraphics[width=0.95\linewidth]{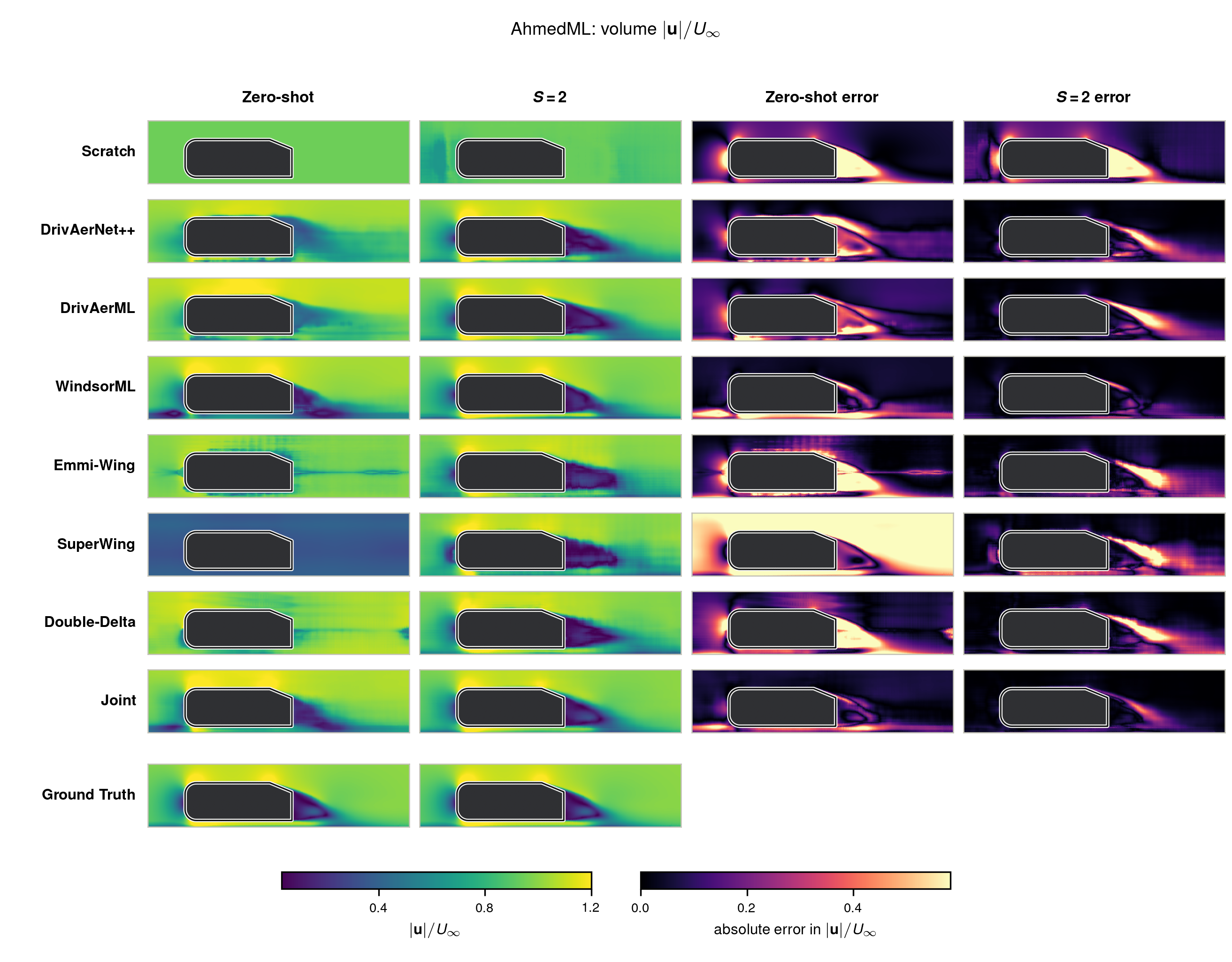}
    \caption{Model predictions of \textit{volume \(|\mathbf{u}^*|\)} on AhmedML. }
\end{figure}

%%%%%%%%%%%%%%%%

\begin{figure}[h!]
    \centering
    \includegraphics[width=\linewidth]{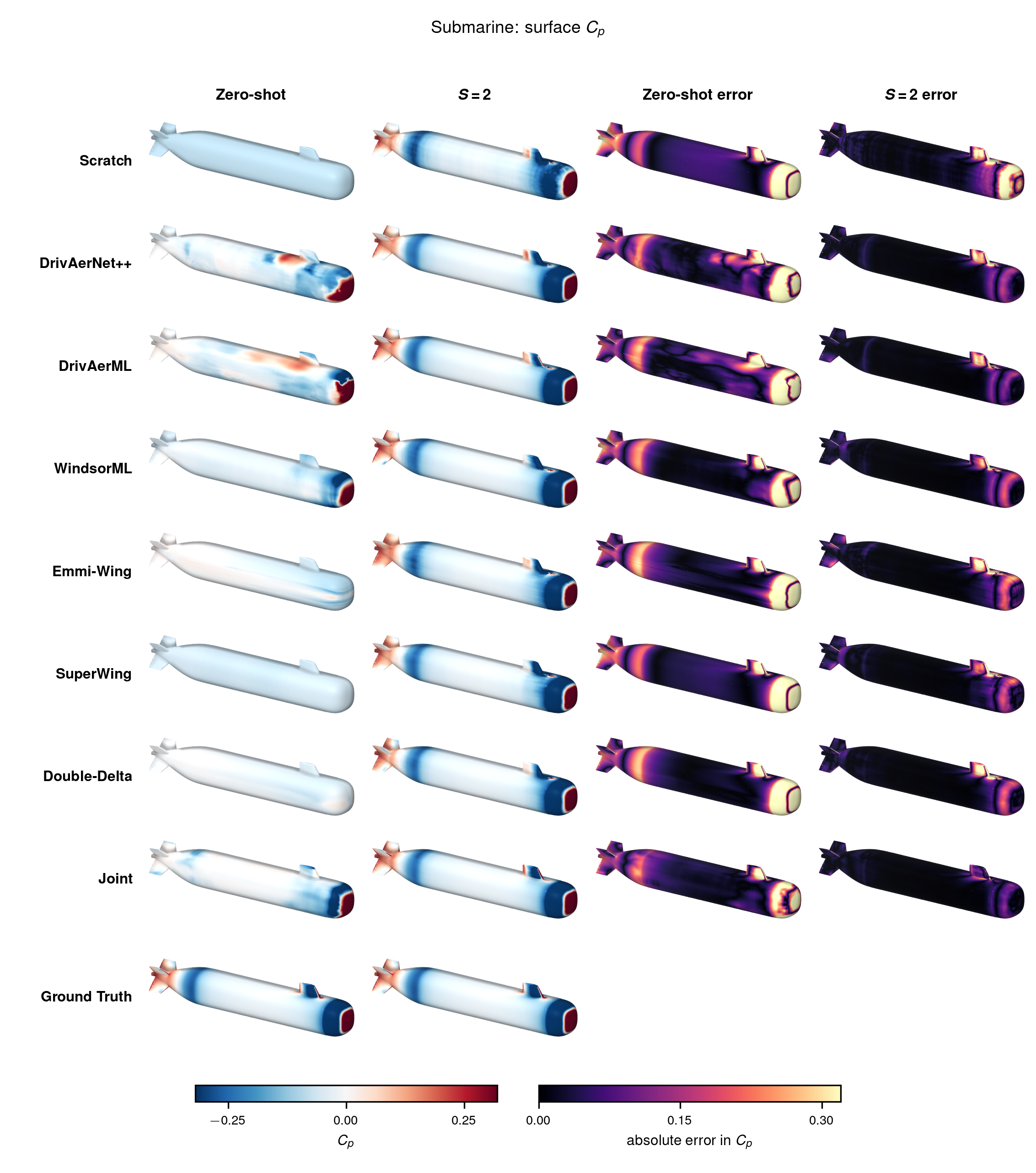}
    \caption{Model predictions of \textit{surface \(C_p\)} on SHIFT-Submarine. }
\end{figure}

\begin{figure}[h!]
    \centering
    \includegraphics[width=\linewidth]{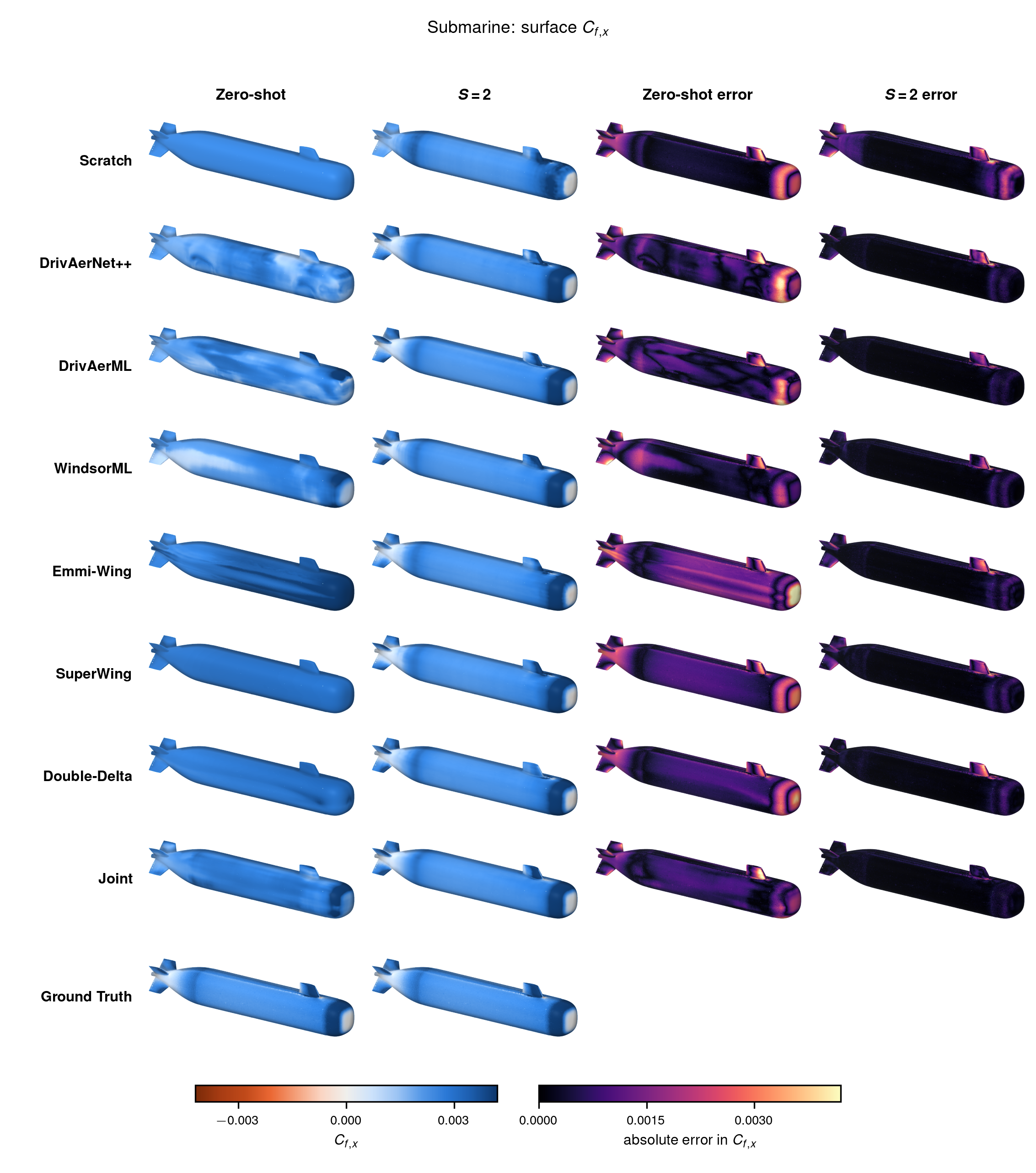}
    \caption{Model predictions of \textit{surface \(C_{f_x}\)} on SHIFT-Submarine. }
    \label{fig:aml_surface_cf}
\end{figure}

\begin{figure}[h!]
    \centering
    \includegraphics[width=\linewidth]{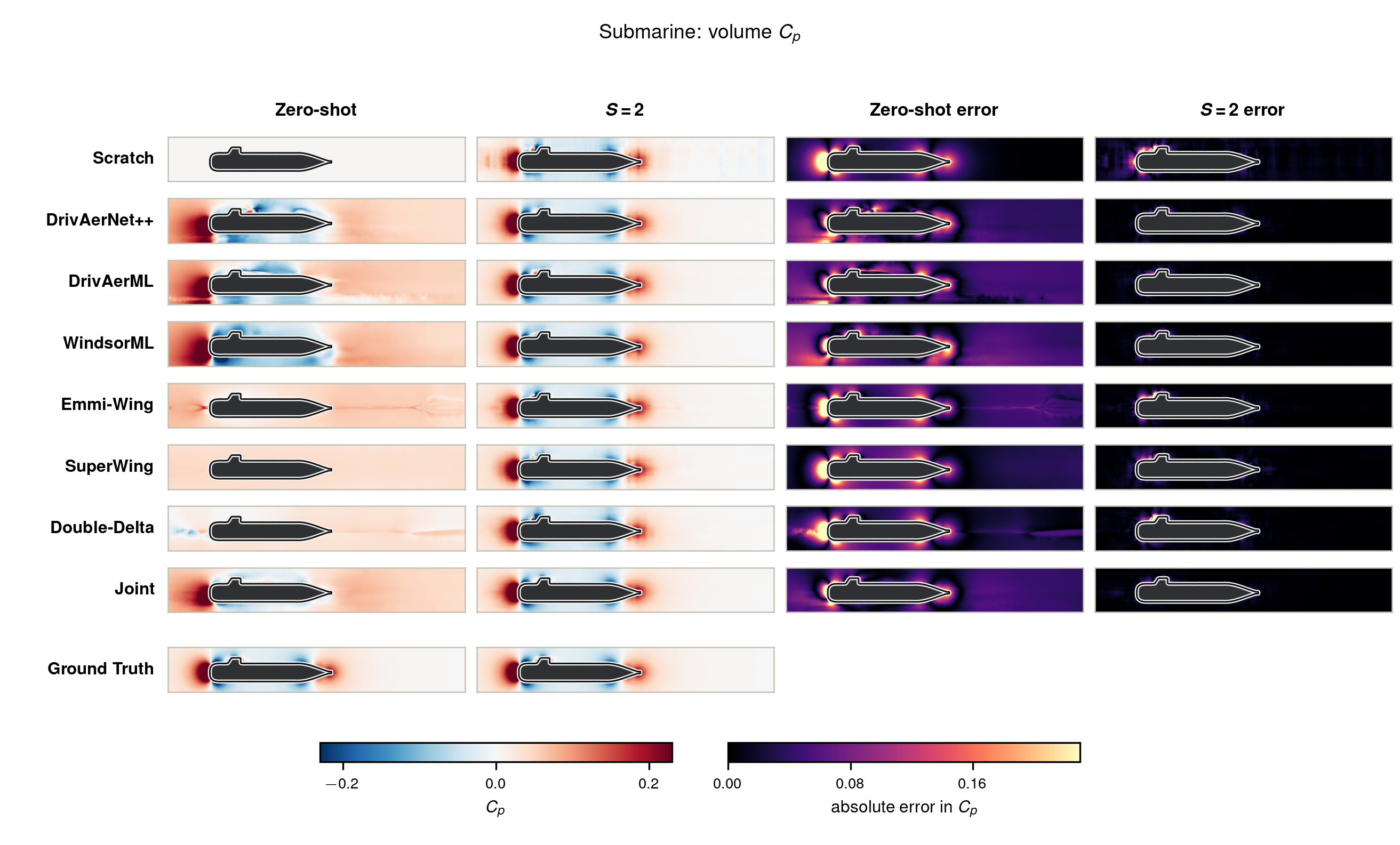}
    \caption{Model predictions of \textit{volume \(C_{p}\)} on SHIFT-Submarine. }
\end{figure}

\begin{figure}[h!]
    \centering
    \includegraphics[width=\linewidth]{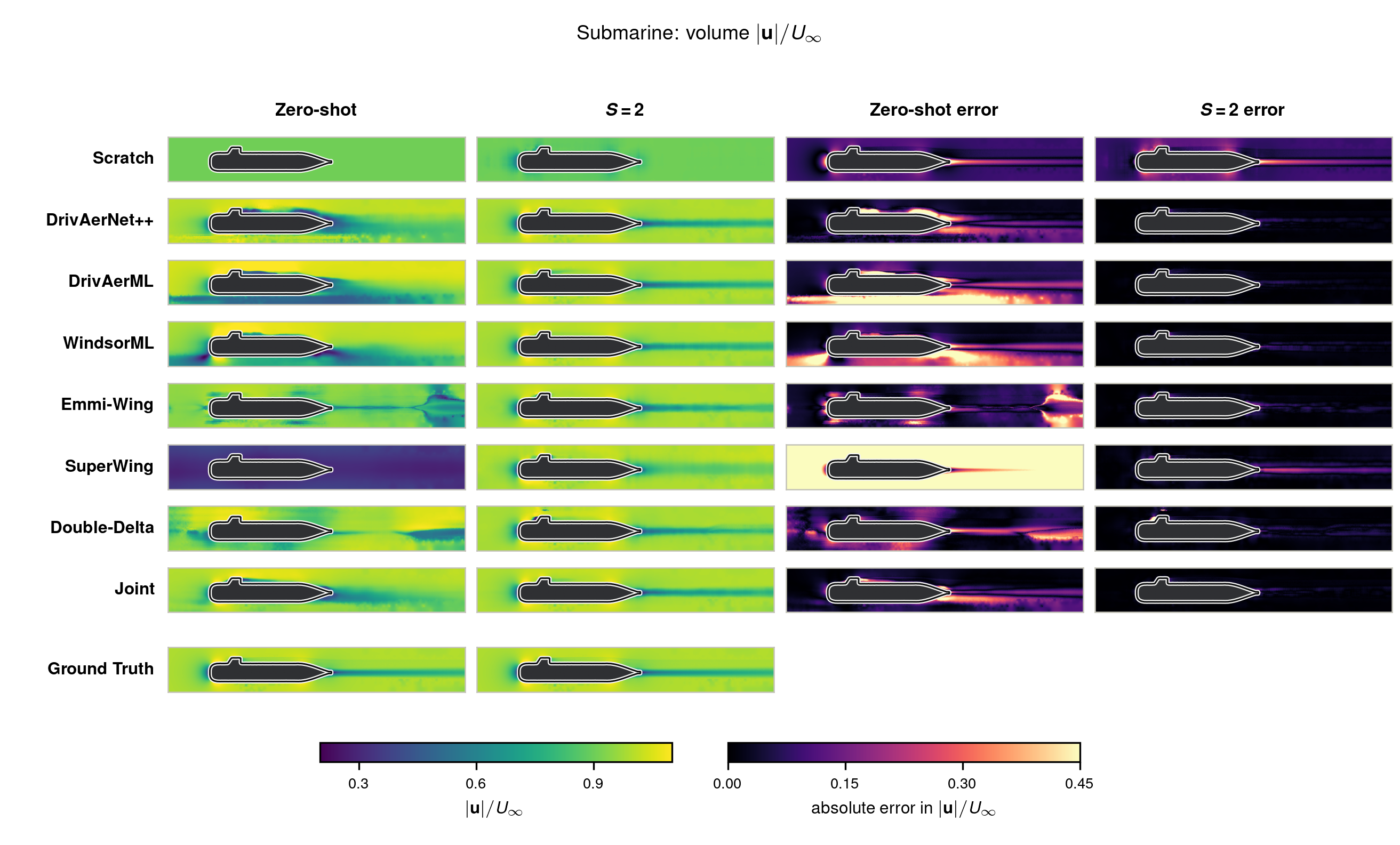}
    \caption{Model predictions of \textit{volume \(|\mathbf{u}^*|\)} on SHIFT-Submarine. }
\end{figure}

%%%%%%%%%%%%%%%%

\begin{figure}[h!]
    \centering
    \includegraphics[width=\linewidth]{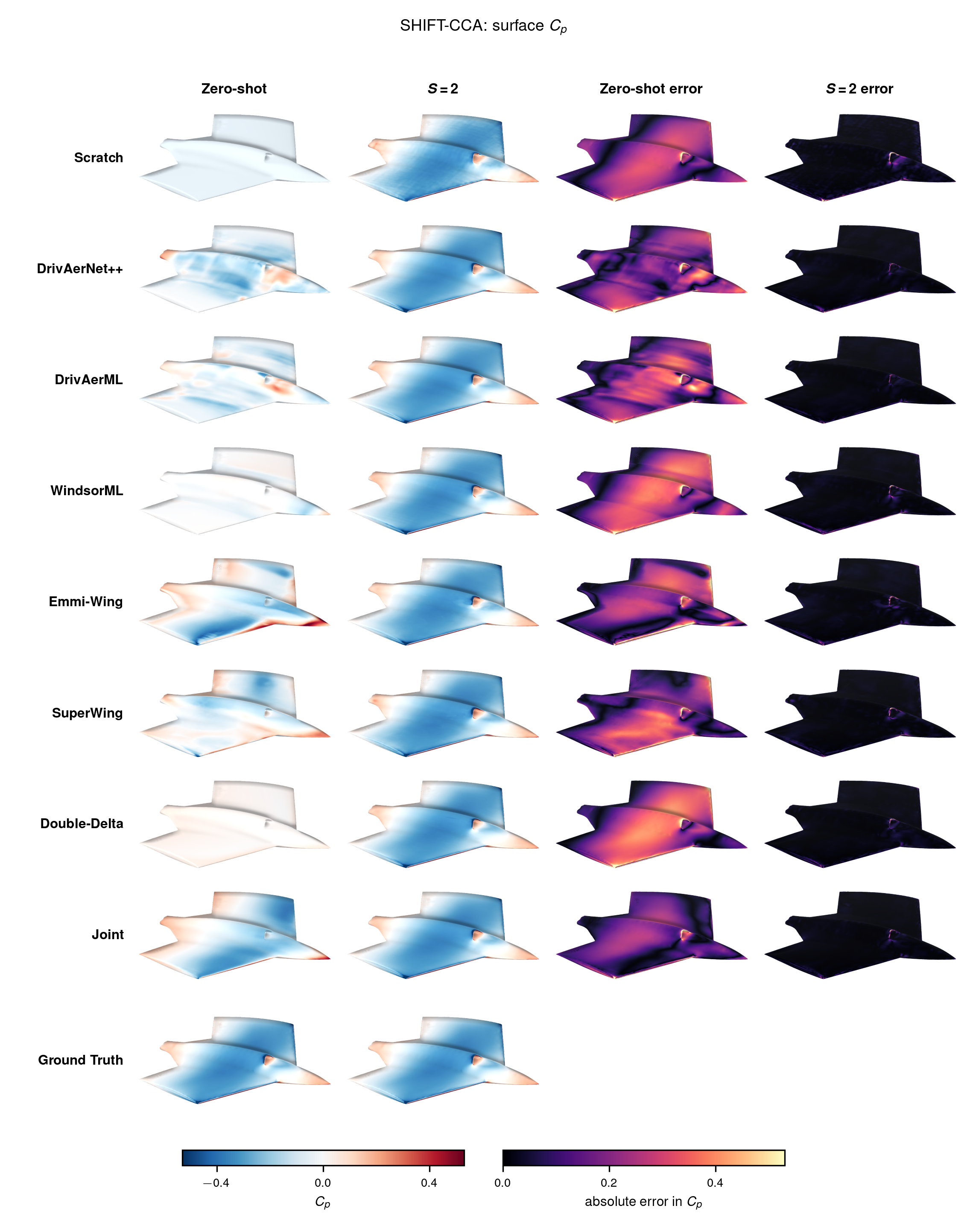}
    \caption{Model predictions of \textit{surface \(C_p\)} on SHIFT-CCA. }
\end{figure}

\begin{figure}[h!]
    \centering
    \includegraphics[width=\linewidth]{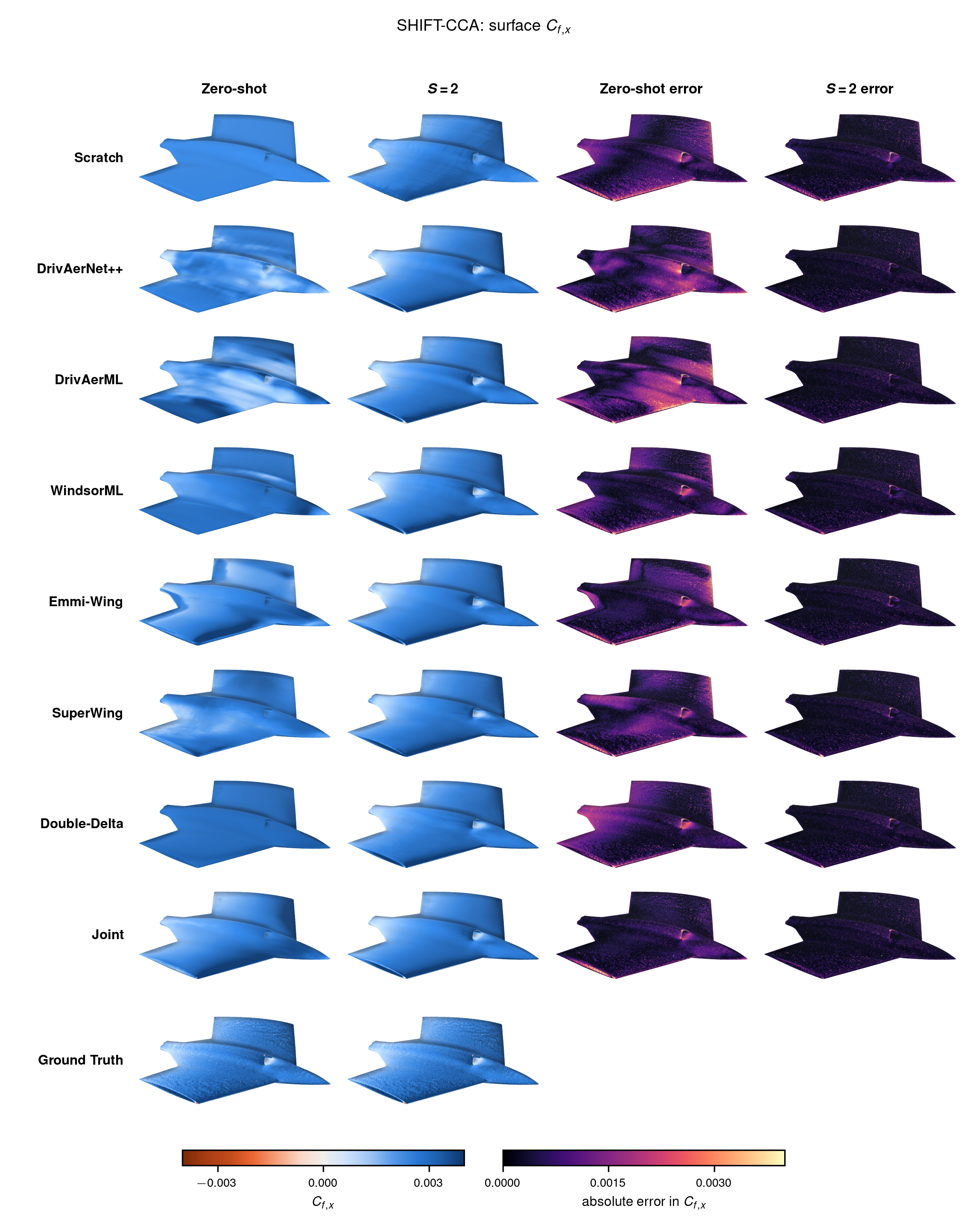}
    \caption{Model predictions of \textit{surface \(C_{f_x}\)} on SHIFT-CCA. }
    \label{fig:aml_surface_cf}
\end{figure}

\begin{figure}[h!]
    \centering
    \includegraphics[width=\linewidth]{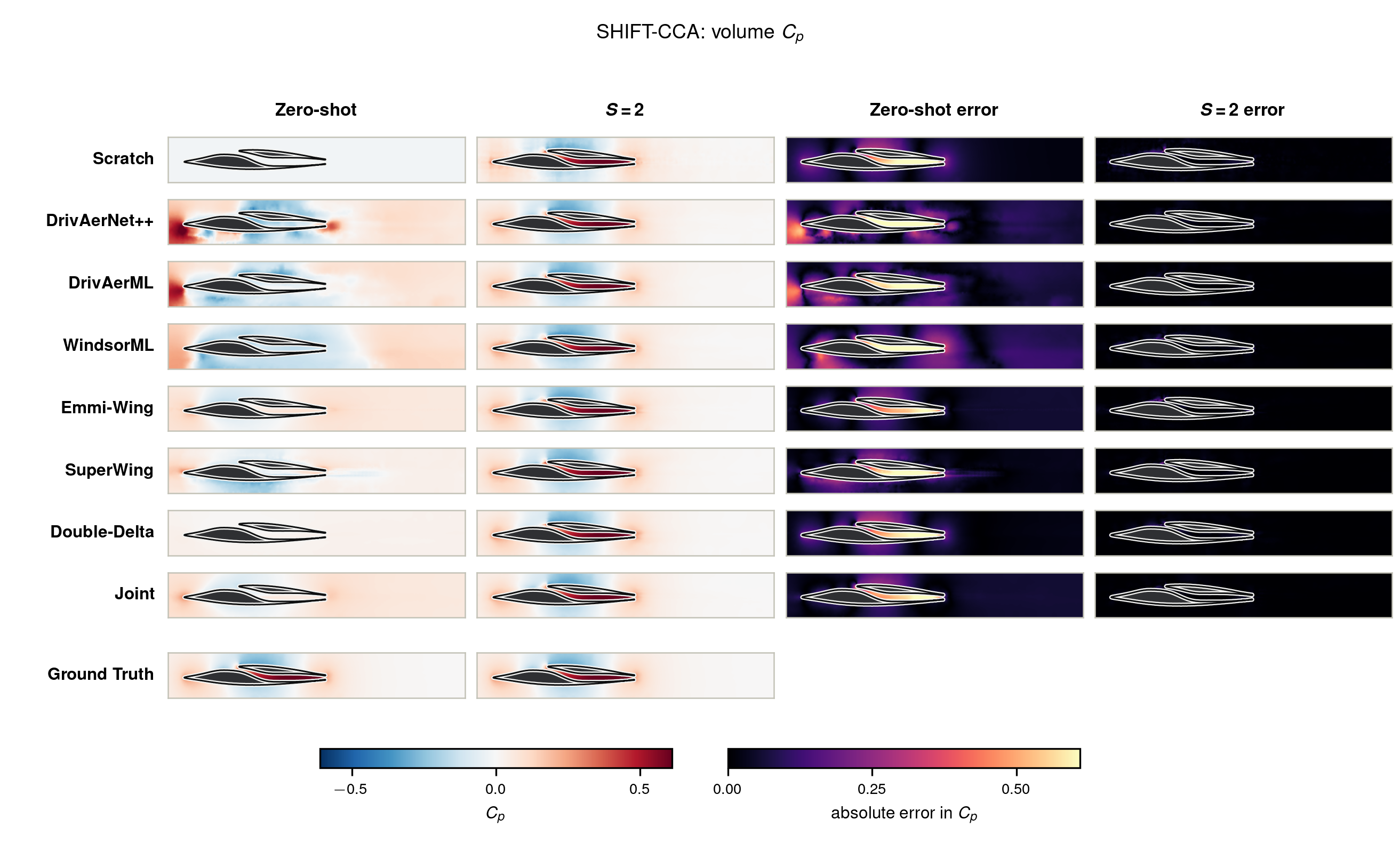}
    \caption{Model predictions of \textit{volume \(C_{p}\)} on SHIFT-CCA. }
\end{figure}

\begin{figure}[h!]
    \centering
    \includegraphics[width=\linewidth]{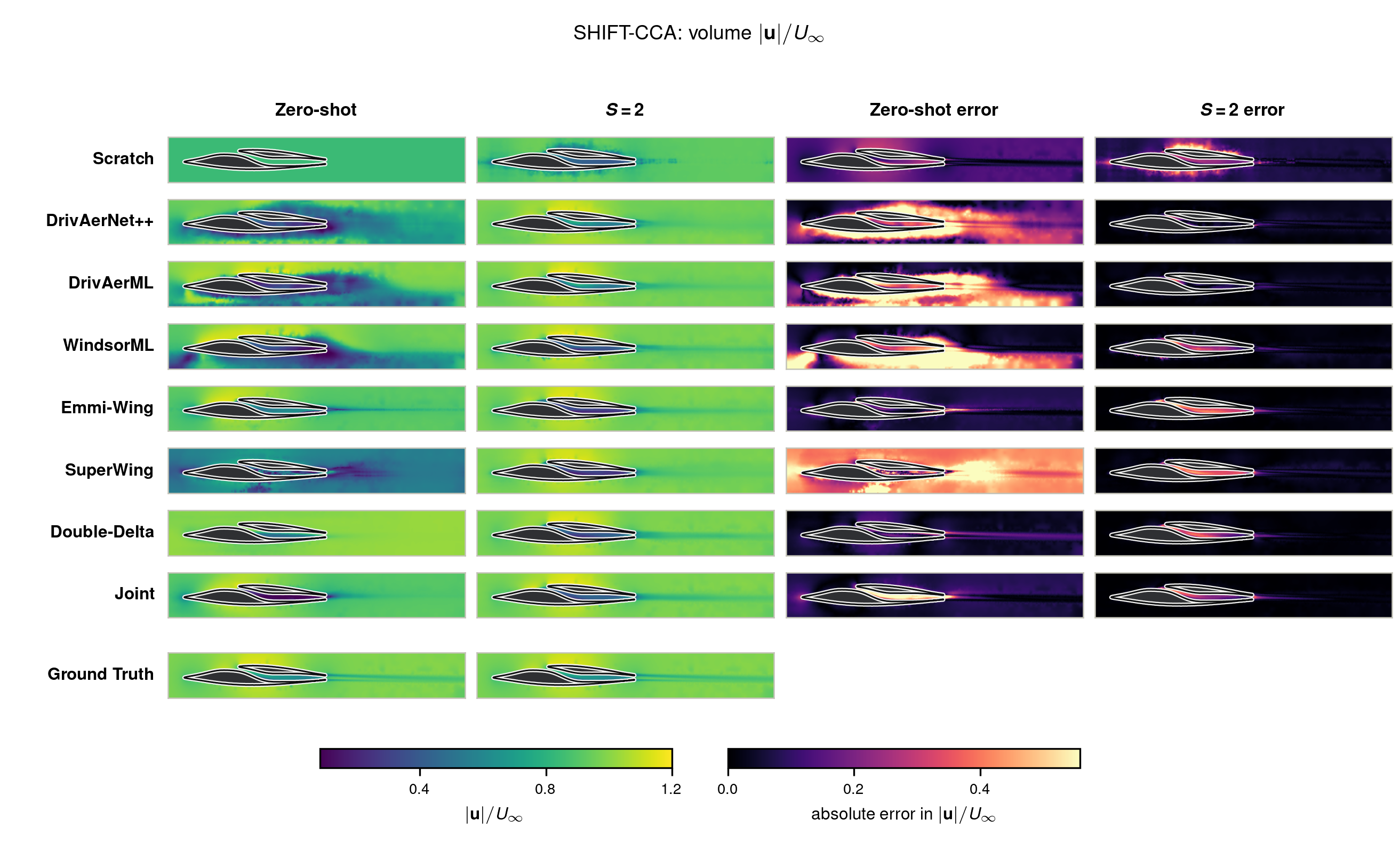}
    \caption{Model predictions of \textit{volume \(|\mathbf{u}^*|\)} on SHIFT-CCA }
\end{figure}

%%%%%%%%%%%%%%%%

\begin{figure}[h!]
    \centering
    \includegraphics[width=\linewidth]{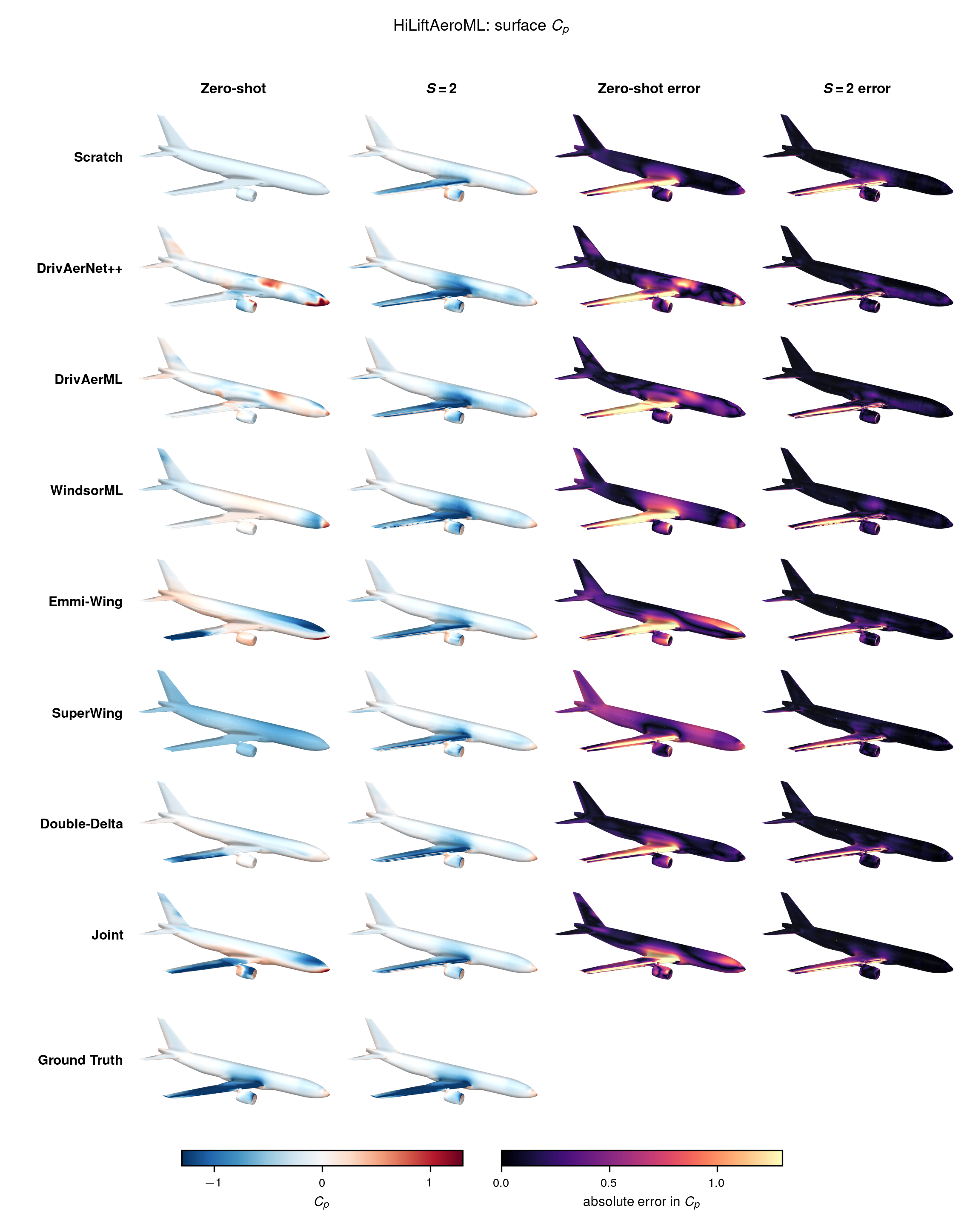}
    \caption{Model predictions of \textit{surface \(C_p\)} on HiLiftAeroML. }
\end{figure}

\begin{figure}[h!]
    \centering
    \includegraphics[width=\linewidth]{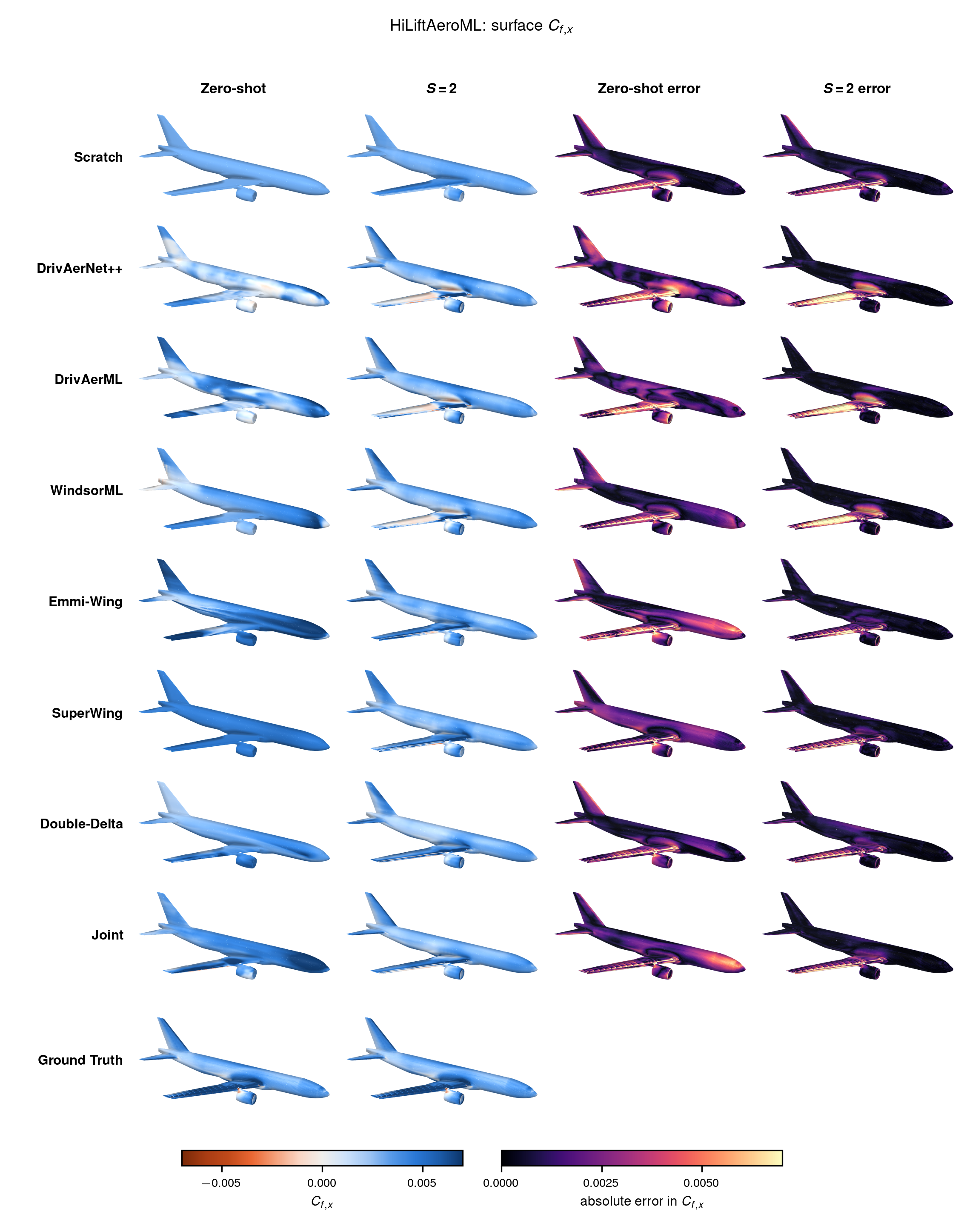}
    \caption{Model predictions of \textit{surface \(C_{f_x}\)} on HiLiftAeroML. }
\end{figure}

\begin{figure}[h!]
    \centering
    \includegraphics[width=\linewidth]{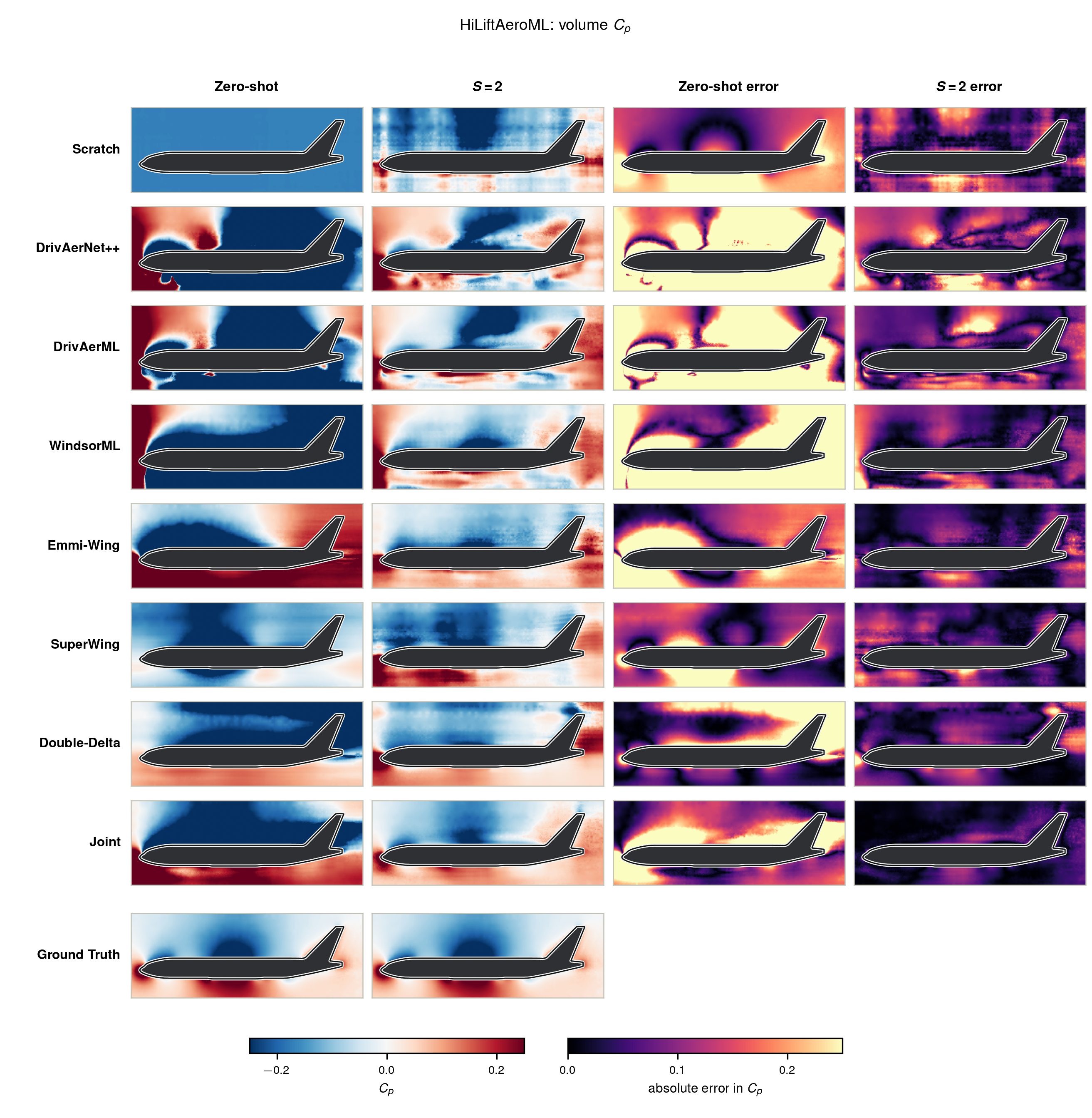}
    \caption{Model predictions of \textit{volume \(C_{p}\)} on HiLiftAeroML. }
\end{figure}

\begin{figure}[h!]
    \centering
    \includegraphics[width=\linewidth]{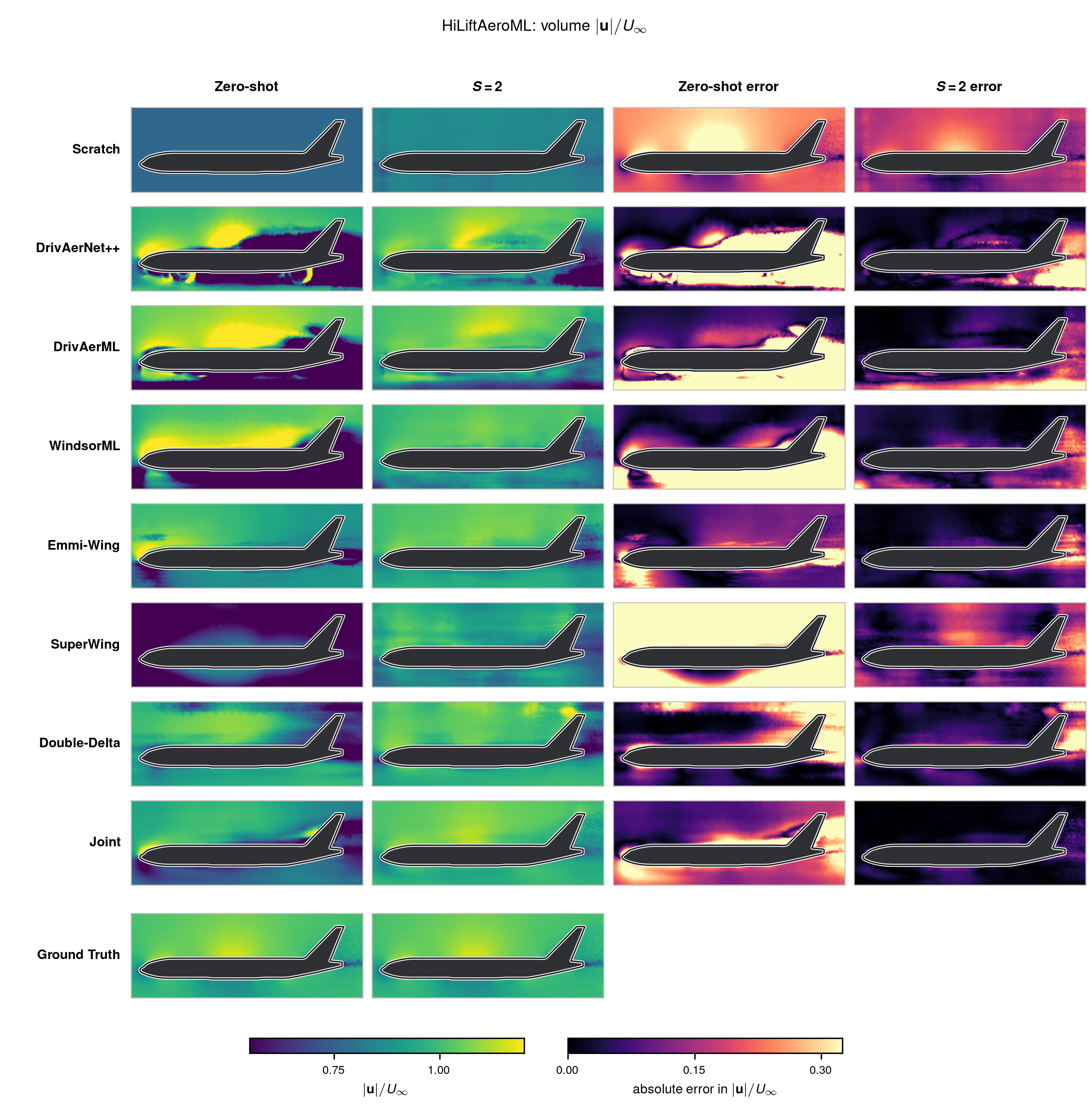}
    \caption{Model predictions of \textit{volume \(|\mathbf{u}^*|\)} on HiLiftAeroML }
\end{figure}

\end{document}

%% file: math_commands.tex
\usepackage{amsmath,amsfonts,bm}

\def\eqref#1{equation~\ref{#1}}
\def\1{\bm{1}}

\DeclareMathAlphabet{\mathsfit}{\encodingdefault}{\sfdefault}{m}{sl}
\SetMathAlphabet{\mathsfit}{bold}{\encodingdefault}{\sfdefault}{bx}{n}